\documentclass[preprint,12pt]{elsarticle}

\makeatletter
\def\ps@pprintTitle{%
	\let\@oddhead\@empty
	\let\@evenhead\@empty
	\let\@oddfoot\@empty
	\let\@evenfoot\@oddfoot
}
\makeatother
\usepackage[12pt]{extsizes}
\usepackage{amsmath,amssymb,amsbsy,amsfonts,latexsym,amsopn,amstext,amsxtra,euscript,amscd,hyperref}
\usepackage{mathptmx}      
\usepackage{graphicx}
\usepackage{hyperref}
\usepackage{algorithm}
\usepackage{algorithmic}
\usepackage{natbib}
\usepackage{notoccite}
\usepackage{graphics,graphicx}
\usepackage{epsfig,epstopdf}
\usepackage{array,tikz}
\usepackage{caption}
\usepackage{subcaption}
\usepackage{slashbox}
\usepackage{wrapfig}
\usepackage{xcolor}
\usepackage{setspace}
\usepackage{titlesec}
\newcolumntype{C}{@{}c@{}}

\usepackage{float}
\usepackage[top=1.5cm, left=1.5cm,right=1.6cm,bottom=1.28cm]{geometry}
\newcounter{myeqno}
\usepackage{chngcntr}

\usepackage{tikz}
\usetikzlibrary{shapes.geometric,arrows}
\usetikzlibrary{shapes.geometric,arrows,fit,matrix,positioning,shapes.multipart}
\tikzstyle{startstop}=[rectangle, rounded corners, minimum width=3cm, minimum height=1cm, draw=black]
\tikzstyle{startstop1}=[rectangle, rounded corners, minimum width=8cm, minimum height=4cm, draw=black]
\tikzstyle{startstop2}=[square, rounded corners, minimum width=0.5cm, minimum height=1cm, draw=black]
\tikzstyle{startstop3}=[square, rounded corners, minimum width=5cm, minimum height=1cm, draw=black]
\tikzstyle{startstop4}=[square, rounded corners, minimum width=1cm, minimum height=5cm, draw=black]
\tikzstyle{round} = [ellipse, minimum width=3cm, minimum height=1cm, draw= black]
\tikzstyle{startstop2}=[rectangle, rounded corners, minimum width=2cm, minimum height=1cm, draw=black]
\tikzstyle{round} = [ellipse, minimum width=3cm, minimum height=1cm, draw= black]
\tikzstyle{arrow} =[draw, -latex']

\usepackage{longtable,lscape}
\usepackage{tabularx}
\usepackage{ltablex}
\usepackage{booktabs,adjustbox}
\usepackage{rotating}
\usepackage{outlines}
\usepackage{multirow,multicol}
\usepackage{makecell}

\usepackage{stfloats}

\usepackage{rotating}
\usepackage{pdflscape}
\usepackage{xcoffins}
\NewCoffin\tablecoffin

\usepackage{colortbl}
\begin{document}
		\begin{frontmatter}
  \title{An Explainable Reconfiguration-Based Optimization Algorithm for Industrial and Reliability-Redundancy Allocation Problems}
            \author{Dikshit Chauhan}
		\ead{dikshitchauhan608@gmail.com}
        \author{Nitin Gupta}
        \ead{nitin291997@gmail.com}
		\author{Anupam Yadav\corref{cor1}}
		\ead{anupam@nitj.ac.in}
		\cortext[cor1]{Corresponding author}

		\address{Department of Mathematics and Computing\\
			Dr. B. R. Ambedkar National Institute of Technology,
			Jalandhar - 144008, INDIA}
\begin{abstract}

Industrial and reliability optimization problems often involve complex constraints and require efficient, interpretable solutions. This paper presents AI-AEFA, an advanced parameter reconfiguration-based metaheuristic algorithm designed to address large-scale industrial and reliability-redundancy allocation problems. AI-AEFA enhances search space exploration and convergence efficiency through a novel log-sigmoid-based parameter adaptation and chaotic mapping mechanism. The algorithm is validated across twenty-eight IEEE CEC 2017 constrained benchmark problems, fifteen large-scale industrial optimization problems, and seven reliability-redundancy allocation problems, consistently outperforming state-of-the-art optimization techniques in terms of feasibility, computational efficiency, and convergence speed. The additional key contribution of this work is the integration of SHAP (Shapley Additive Explanations) to enhance the interpretability of AI-AEFA, providing insights into the impact of key parameters such as Coulomb’s constant, charge, acceleration, and electrostatic force. This explainability feature enables a deeper understanding of decision-making within AI-AEFA framework during the optimization processes. The findings confirm AI-AEFA as a robust, scalable, and interpretable optimization tool with significant real-world applications.
\end{abstract}

\begin{keyword}
	Intelligent algorithms, Real-parameter, Real-world optimization, Reliability-redundancy allocation problems, Parameter reconfiguration, Explainability
\end{keyword}
\end{frontmatter}
\section{Introduction}
The advent of 5G networks has transformed the networking industry through the softwarization of services. Technologies like software-defined networking (SDN) and network function virtualization (NFV) have introduced dynamic features such as programmability, flexibility, and scalability. However, these advancements also present new challenges in system reliability, including software/hardware failures and misconfiguration of Virtual Network Functions (VNFs). Today's communication services consist of multiple sessions, each supported by various network functions across different network nodes. Evaluating system reliability by assessing source-to-sink sub-networks in isolation is not feasible due to these complexities. Reliability is a critical metric in advanced engineering, indicating a system or component's ability to perform without failure under defined conditions for a specific period. In addressing reliability issues, the aim is to minimize the likelihood of failure or downtime, requiring analysis and optimization of individual components and overall system reliability. Mathematical models, statistical methods, and optimization algorithms are commonly used for this purpose. Engineers actively enhance system and product reliability during the design phase, with redundancy emerging as a promising technique. This involves integrating redundant components to boost system reliability. Redundancy is widely applied in engineering disciplines where exceptionally high reliability is crucial, including spacecraft, telecommunications, and nuclear power plants~\cite{navarro2020survey,trivedi2017reliability}.

Redundancy is a strategy employed to bolster system reliability by introducing additional components or resources capable of taking over in the event of a failure. Various types of redundancy exist, including standby redundancy (activated backup components upon failure), active redundancy (multiple components operating simultaneously), and hybrid redundancy (a combination of standby and active). The challenge in redundancy problems lies in determining the optimal configuration of redundant components to achieve the desired reliability level, considering factors such as cost, weight, and other constraints. The redundancy allocation problem (RAP) involves selecting a set of available components and determining their optimal redundancy levels at different stages. The objective is to attain maximum or desired system reliability or availability while adhering to constraints like volume, budget, weight, etc., or minimizing expenditures~\cite{garg2015efficient,ardakan2018multi}. Due to its widespread significance, the RAP has been extensively explored from various perspectives over the past few decades~\cite{yeh2017optimal}.

The traditional RAP focuses on determining redundancy levels for components within subsystems, categorizing available components by functions, volumes, costs, weights, etc. Redundancy levels are treated as discrete decision variables, rendering RAP an integer programming optimization problem subject to multiple resource constraints ~\cite{peiravi2019reliability}. A more challenging problem, the reliability-redundancy allocation (RRA) problem, involves unknown reliability and redundancy variables, in contrast to RAP, where redundancy variables are known and reliability variables are unknown. RRA problem requires simultaneous optimization of both components' reliabilities and redundancy levels to enhance system reliability~\cite{li2022methods}. Formulated as a computationally challenging non-linear mixed-integer programming problem, the RRA problem takes precedence in this paper over the traditional RAP.

Most studies on RRA problems or RAPs emphasize the active redundancy strategy, where all subsystem components operate simultaneously in parallel upon system activation~\cite{nath2022evolutionary}. However, this approach introduces vulnerability, as the subsystem becomes inoperative when the last active redundant component fails. In contrast, the cold standby redundancy strategy, summing the lifetimes of all subsystem components, is explored to enhance subsystem reliability, with applications in steam and power systems~\cite{sun2015reliable} and integrated monitoring and control systems for offshore drilling~\cite{yeh2019solving,hsieh2021component}. Recent research introduces a mixed redundancy strategy in RRA problems~\cite{gholinezhad2017new}, allowing designers to select active redundant and cold standby components from a pool. Extended to RRA problems, this strategy enhances system reliability compared to active redundancy or cold standby alone. Previous assumptions of homogeneity in component reliability within subsystems may not align with practical engineering scenarios, where redundant systems often comprise components with distinct reliability measures. For example, airplanes use primary electronic gyroscopes alongside secondary mechanical gyroscopes, differing significantly in structure and reliability measures. Various types of RRA problems have emerged, such as a fuzzy RRA problem addressing parameter uncertainty~\cite{ashraf2015pso}, a heterogeneous RRA problem incorporating different component types~\cite{ouyang2019improved}, and a multi-state $k$-out-of-$n$ RRA problem with repairable components~\cite{baladeh2022reliability}. Regardless of the standby strategy, redundancy components are crucial to replace failed main components and prevent system crashes. The active RRA problem with an active redundancy strategy is the most widely employed, making it the selected method for the present study.	

The NP-hard nature of the RRA problem results in increased computational complexity as the system configuration scales up. These problems are highly non-linear, involving mixed discrete and continuous variables. The intricate nature of RRA problems has prompted extensive research over the past decades, leading to the exploration of various intelligent techniques for diverse systems~\cite{yi2019trade}. In addition to exact and heuristic methods like dynamic programming and branch-and-bound~\cite{meng2021comparative,ha2006reliability}, meta-heuristic algorithms have been employed due to their robust global search capabilities, efficiently finding optimal solutions within a limited time frame.

Novel operators and variations, such as a penalty-based Artificial bee colony algorithm for RRA problem~\cite{monalisa2023multi}, a hybrid Harmony search incorporating Particle swarm optimization (PSO)~\cite{zou2011effective}, an advanced Imperialist competitive algorithm for RRA problem~\cite{afonso2013modified}, and a particle-based simplified swarm optimization~\cite{yeh2021simplified}, contribute to improved global search ability. Additionally, innovative algorithms like the Artificial fish swarm algorithm~\cite{he2015novel}, which mimics fish swarm behavior, demonstrate high computational accuracy and efficiency for large-scale RRA problems.

Inspired by these studies, we propose a restructured metaheuristic algorithm to efficiently solve large-scale RRA problems. The proposed algorithm is based on the physics-based Artificial electric field algorithm (AEFA) \cite{yadav2019AEFA}. Initially designed for unconstrained problems, AEFA stands out for its promising performance on non-linear unconstrained problems, featuring faster convergence, fewer parameters, and reduced computational time. This article introduces a new version of AEFA, termed AI-AEFA, by redefining its parameters tailored for industrial and RRA problems. The revised parameters facilitate the exploration of the entire search space, as supported by experimental results discussed in the experimental section.  

{The key contributions of this article are as follows:}
\begin{enumerate}[(i)]
\item Proposal of a parameter reconfiguration-based algorithm designed to address real-world industrial and RRA problems.
\item Introduction of an intelligent parameter adaptation mechanism within the algorithm.
\item Performance evaluation of the algorithm on twenty-eight constrained CEC 2017 benchmark problems.
\item Demonstrate the algorithm's effectiveness by successfully resolving fifteen real-world industrial problems and seven RRA problems.
\item The explainable analysis of AI-AEFA has been done using SHAP. 
\end{enumerate}

This article is organized as Section~\ref{aefa} explains the proposed optimization algorithm along with the artificial electric field algorithm and bounds mechanism. Section~\ref{constrainedhandling} explains the constraint handling technique. An experimental study between AEFA, AEFA-C, and AI-AEFA is given in Section~\ref{AEFAHKvsAEFA}. In Section~\ref{experimental}, the experimental results are compared with other optimization algorithms, and the effectiveness of the AI-AEFA is evaluated. Industrial real-world optimization problems are solved in Section~\ref{reallifeapplication}. Reliability-redundancy allocation problems are discussed in Section~\ref{reliability}. The explainable analysis of AI-AEFA is conducted in Section \ref{explainable}. Section~\ref{conclusion} presents the findings of this article along with future directions.
\section{Proposed intelligent parameter reconfiguration algorithm}\label{aefa}
	This section explains a detailed methodology of the proposed algorithm along with the original AEFA and adaptive bounds.
	\subsection{AEFA}
	AEFA~\cite{yadav2019AEFA} stands out among physics-based optimization algorithms, leveraging principles from physical laws like Coulomb's and motion laws. Initially designed for unconstrained optimization problems, AEFA evaluates agent performance by treating them as charged entities. In this approach, agents employ Coulomb's law to exert electrostatic forces and follow motion laws to adjust their positions. Essentially, each agent's trajectory is influenced by the force it experiences, with higher-charged agents attracting those with lower charges. This mechanism alters the movement direction of each agent, reducing distances between them. The charge, determined by fitness, inter-agent distances, and Coulomb's constant, dictates the electrostatic force's impact on all agents. To understand the framework of AEFA, let us assume $N$ agents (population)  in the search space whose positions and velocities are represented as $\mathbf{X}_i^d=\{\mathbf{x}_i^1,\mathbf{x}_i^2,\cdots,\mathbf{x}_i^D\}$, $\mathbf{V}_i^d=\{\mathbf{v}_i^1,\mathbf{v}_i^2,\cdots,\mathbf{v}_i^D\}$ for all $i=1,2,\cdots, N$. The working procedure of AEFA starts by randomly initializing the position of each agent within their search bounds, and the velocity is set to zero initially. After initializing the position of each agent in the search space, the fitness value of each agent in the population is evaluated using the objective function $O(\mathbf{X}_i^d)$, represented as $\mathbf{fit}_i$.
	\begin{figure}
		\centering
		\begin{tikzpicture}[node distance=1.5cm,auto]
		\tikzset{->-/.style={decoration={
					markings,
					mark=at position #1 with {\arrow{>}}},postaction={decorate}}}
		\draw[line width=1.5pt,dashed](0,-0.7)--(0,4)--(8,4)--(8,-0.7)--(0,-0.7);
		\filldraw[gray!50] (3,3.5) circle (12pt) node[anchor=center](2)at (3,3.5){\textbf{\textcolor{black}{$Q_2$}}};
		\filldraw[gray!50] (0.85,1.8) circle (9pt) node[anchor=center](1)at (0.95,1.8){\textbf{\textcolor{black}{$Q_1$}}};
		\filldraw[gray!50] (7,0.3) circle (18pt) node[anchor=center](3)at (7,0.3){\textbf{\textcolor{black}{$Q_3$}}};
		\filldraw[gray!50] (6,3) circle (15pt) node[anchor=center](5)at (6,3){\textbf{\textcolor{black}{$Q_5$}}};
		\filldraw[gray!50] (2,0) circle (11pt) node[anchor=center](4)at (2,0){\textbf{\textcolor{black}{$Q_4$}}};
		\draw[thick,dashed](4,2.5)--(5.5,1.5);
		\draw[thick,dashed](4,0.8)--(5.5,1.5);
		\draw[arrow,line width=1.5pt](1.1,2.2)--(2,2.7);
		\draw[arrow,line width=1.5pt](1)--(4,2.5);
		\draw[arrow,line width=1.5pt](1)--(4,0.8);
		\draw[arrow,line width=1.5pt](1)--(5.5,1.5);
		\node[right]at(1,2.8){\textbf{$F_{12}$}};
		\node[above]at(3.5,2.4){\textbf{$F_{15}$}};
		\node[left]at(3.5,1.4){\textbf{$F_{13}$}};
		\node[right]at(3.8,1.8){\textbf{$F_{total}$}};
		\node[below]at(4,-1){\textbf{$(a)$}};
		
		\draw[line width=1.5pt,dashed](9,-0.7)--(9,4)--(17,4)--(17,-0.7)--(9,-0.7);
		\filldraw[gray!50] (12,3.5) circle (12pt) node[anchor=center](2)at (12,3.5){\textbf{\textcolor{black}{$Q_2$}}};
		\filldraw[gray!50] (9.85,1.8) circle (9pt) node[anchor=center](1)at (9.95,1.8){\textbf{\textcolor{black}{$Q_1$}}};
		\filldraw[gray!50] (16,0.3) circle (18pt) node[anchor=center](3)at (16,0.3){\textbf{\textcolor{black}{$Q_3$}}};
		\filldraw[gray!50] (15,3) circle (15pt) node[anchor=center](5)at (15,3){\textbf{\textcolor{black}{$Q_5$}}};
		\filldraw[gray!50] (11,0) circle (11pt) node[anchor=center](4)at (11,0){\textbf{\textcolor{black}{$Q_4$}}};
		\draw[thick,dashed](14,3)--(15.5,1.5);
		\draw[thick,dashed](14.8,1.2)--(15.5,1.5);
		\draw[arrow,line width=1.5pt](2)--(14,3);
		\draw[arrow,line width=1.5pt](2)--(14.8,1.2);
		\draw[arrow,line width=1.5pt](2)--(15.5,1.5);
		\node[above]at(13.8,3.2){\textbf{$F_{25}$}};
		\node[below]at(14,1.7){\textbf{$F_{23}$}};
		\node[right]at(15.5,1.8){\textbf{$F_{total}$}};
		\node[below]at(13,-1){\textbf{$(b)$}};
		\end{tikzpicture}
		\caption{ {Interaction of agents in AEFA.}}\label{fig: aefa representation}
	\end{figure}
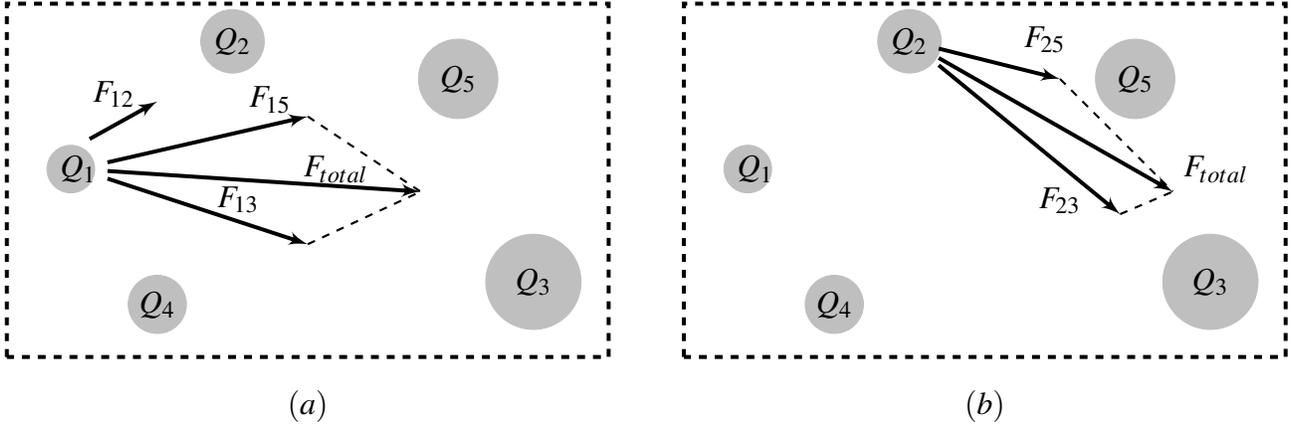
	
	 In each iteration $\l$, the set of best agents is selected from the population and denoted as $K_{best}$. Each agent in the set $K_{best}$ inflicts the electrostatic force on the rest of the agents in the population. It is noted that the agents outside the set $K_{best}$ in the population do not impose any electrostatic force on other agents. The size of the set $K_{best}$ decreases linearly from $N$ to 2, meaning that initially, all agents are considered, and in later iterations, only the best agents are used for attraction. The concept of electrostatic force in AEFA is expressed as follows:
	 \begin{equation}\label{electrosticforce}
	F_{ij}^{d}(\l) = K(\l)\frac{Q_{i}(\l) *Q_{j}(\l)(\mathbf{x}_{j}^{d}(\l) - \mathbf{x}_{i}^{d}(\l))}{R_{ij}(\l) + \epsilon},
	\end{equation} where $Q_{i}(\l)$ and $Q_{j}(\l)$ represent the charges on the $i^{th}$ and $j^{th}$ agents. The charge, which is a function of fitness, is calculated as follows: \begin{equation}
	q_{i}(\l) = exp\left(\frac{\mathbf{fit}_{i}(\l) - max(\mathbf{fit}_{i}(\l))}{min(\mathbf{fit}_{i}(\l)) - max(\mathbf{fit}_{i}(\l))}\right).
	\end{equation} 
	
	 Due to the direct relationship between charge and fitness, a normalization method is employed to normalize the value of the charge and obtain the final charge value on the $i^{th}$ agent, given as:\begin{equation}\label{charge}
	Q_{i}(\l) = \frac{q_{i}(\l)}{\sum_{i=1}^{N}q_{i}(\l)}.
	\end{equation}  
	
	In Eq.~\eqref{electrosticforce}, Coulomb's constant, $K(\l)$, a crucial parameter controlling the algorithm's searching behavior, is defined by Eq.~(\ref{coulomb}), \begin{equation}\label{coulomb}
		K(\l)=K_{0}e^{-\alpha \l/\l_{max}}.
	\end{equation} 
	
	In this context, $K_0$ and $\alpha$ represent user-defined parameters, while $\l$ and $\l_{max}$ denote the current and maximum number of iterations, respectively. Coulomb's constant governs both the overall applied force and the balancing steps in the motion of candidate solutions. As $K(\l)$ is an exponentially decreasing function with iterations, it compels more substantial movements in the early stages and minor movements in the later phases of the search process. This characteristic allows AEFA to exhibit both explorative and exploitative behavior during the search. The Euclidean distance ($R_{ij}(\l)$)) and the total applied force ($F_i(\l)$) on agent $i$ are determined as follows: \begin{align}\label{totalforce}
	&R_{ij}(\l)=\| \mathbf{x}_{i}(\l), \mathbf{x}_{j}(\l)\|,\nonumber\\&
	F_{i}^{d}(\l)=\sum_{j=1,j\neq i}^{N} rand_{j}F_{ij}^{d}(\l),
	\end{align} 
	Here, $rand_{j}$ is a uniformly distributed random variable ranging from 0 to 1. This variable is employed to introduce both the agent's random movement and the electrostatic force, enabling a more diverse exploration of the search space. 
	
	In Fig.~\ref{fig: aefa representation}, a hypothetical scenario illustrates the movement of AEFA search agents in a simplified setting. Five agents (1$-$5) with different charges ($Q_1$ to $Q_5$) aim to find a global optimum. Assuming $K_{best}$ is set to 3, comprising agents 2, 3, and 5, the electrostatic force interactions between agents inside and outside $K_{best}$ are depicted in $(a)$ and $(b)$ scenarios. The forces are exerted based on the charges, where higher charges, like $Q_3$ and $Q_5$, have a stronger influence. The resulting force, $F_{total}$, is presented in both scenarios.
	
	 Following the computation of the overall force, two physical laws are applied to calculate the acceleration of each agent. The acceleration for each agent is defined as follows:\begin{equation}\label{acceleration}
	\mathbf{ac}_{i}^{d}(\l) = \frac{F_{i}^{d}(\l)}{Ms(\l)}.
	\end{equation}
	
	Here, the mass $Ms(\l)$ is considered as the unit mass. Moreover, to guide the agents from one location to another, the velocity and position of each agent are iteratively updated in response to the electrostatic force, as defined in Eqs. (\ref{AEFAvelocity}) and(\ref{AEFAposition}) for the $i^{th}$ agent.\begin{align}
	&\mathbf{v}_{i}^{d}(\l+1)=rand()*\mathbf{v}_{i}^{d}(\l)+ \mathbf{ac}_{i}^{d}(\l),\label{AEFAvelocity}\\&
	\mathbf{x}_{i}^{d}(\l+1)=\mathbf{x}_{i}(\l)+ \mathbf{v}_{i}^{d}(\l+1),\label{AEFAposition}
	\end{align}  
	where, $\mathbf{v}_{i}^{d}(\l)$, $\mathbf{x}_{i}^{d}(\l)$, and $\mathbf{ac}_{i}^{d}(\l)$ represent the velocity, position, and acceleration of the $i^{th}$ agent at the $\l^{th}$ iteration in the $d^{th}$ dimension, respectively. The operational steps of the original AEFA are outlined in Algorithm~\ref{AEFApseudocode}. The subsequent section provides an elaborate explanation of the proposed chaotic hierarchical Coulomb's constant, also known as the restructured Coulomb's constant, and introduces modifications to the bounds of velocity and position.
	\begin{algorithm}
		\caption{: {Pseudo code of AEFA}}
		\label{AEFApseudocode}
		\begin{algorithmic}[1]
			\STATE Generate initial position and velocity,
			\WHILE{Stopping criterion is not satisfied,}
			\STATE	Calculate the fitness values for all agents,
			\STATE	\textbf{ Calculate} Coulomb's constant $K(\l)$ using Eq.~(\ref{coulomb}),
			\STATE \textbf{Calculate} the charge on each agent according to Eq.~(\ref{charge}),
			\STATE \textbf{Update} the electrostatic force and acceleration of all agents by using Eqs.~(\ref{electrosticforce}) and~(\ref{acceleration}),
			\STATE \textbf{Update} the velocities and positions for all agents as Eqs.~(\ref{AEFAvelocity}) and~(\ref{AEFAposition}), respectively, 
			
			\ENDWHILE
			\RETURN the best search agent and the best solution.
		\end{algorithmic}
	\end{algorithm}
	\subsection{Parameter reconfiguration mechanism}\label{chk}
	This subsection proposes a new variant of AEFA called AI-AEFA for the constrained optimization problems. To propose this algorithm, a new chaotic sigmoid Coulomb's constant is proposed.	The next section explains in detail the sigmoid Coulomb's constant.
	\subsubsection{Chaotic sigmoid Coulomb's constant}
\begin{figure}
		\begin{minipage}[b]{0.5\linewidth}
			\centering
			\includegraphics[scale = 0.45]{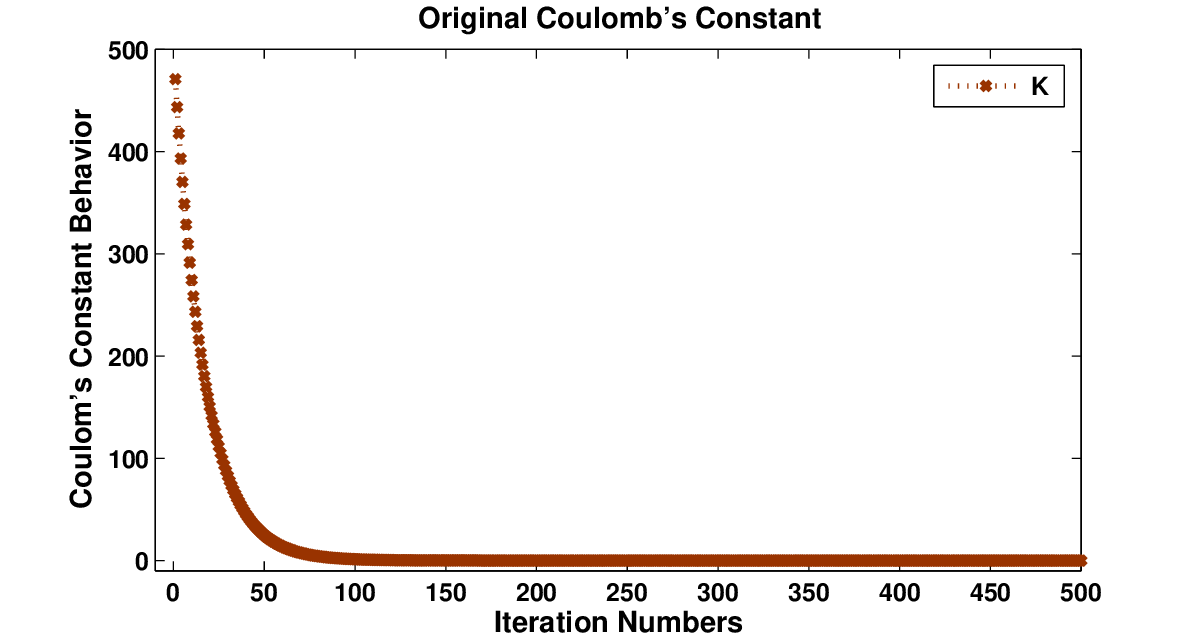}
		\end{minipage}
		\begin{minipage}[b]{0.5\linewidth}
			\centering
			\includegraphics[scale = 0.45]{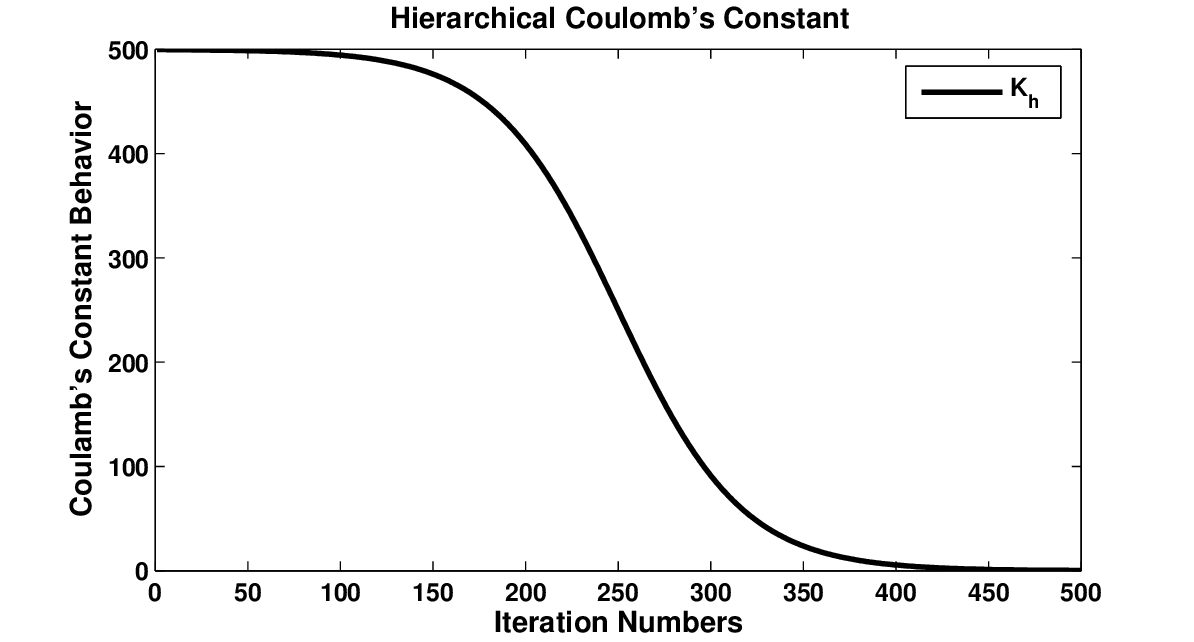}
		\end{minipage}
		\caption{{The plot of the original and proposed Coulomb's constant}}\label{fig: sigmoidgraph}
	\end{figure}

	This section introduces a restructured definition of Coulomb's constant, incorporating the log-sigmoid function and chaotic maps. The algorithm's exploration and exploitation capabilities are pivotal for effectively navigating the search space in both the initial and later stages. In the original AEFA, Coulomb's constant is a decreasing function over iterations, limiting substantial movements in the initial stages while permitting minor adjustments in the later phases (refer to Fig.~\ref{fig: sigmoidgraph} (left)). Additionally, $K(\l)$ influences population diversity maintenance. However, the original AEFA's use of exponentially declining $K(\l)$ for electrostatic force calculations may restrict exploration, leading to weaker exploration abilities~\cite{chauhan2023archive}. This limitation hampers AEFA from adequately exploring the entire search space, and during later stages, exploitation may target an inferior optimum. In essence, there is room for enhancing AEFA's exploration abilities, especially in the early iterations.
	
	Given the critical role of an algorithm's exploration ability, this article proposes an improved definition for Coulomb's constant to boost AEFA's exploration capabilities. The log-sigmoid function is employed in this new definition, outlined in Eq.~(\ref{sigmoidcoulomb}): \begin{equation}\label{sigmoidcoulomb}
	K_{h}(l)=K_{0} \left(1+exp\left(\frac{\beta(l-\frac{l_{max}}{2})}{\delta}\right)\right)^{-1},
	\end{equation} where $\beta$ and $\delta$ are initially defined parameters. After applying the hierarchy definition, a chaotic Sine map~\cite{li2011novel} (Eq.~(\ref{sinemap})) formulates the chaotic hierarchical Coulomb's constant as follows:
	
	\begin{equation}\label{sinemap}
	K_{1}(\l+1)=\frac{r}{4}\sin(\pi K_{1}(\l)).
	\end{equation}
	
	Here, $r$ represents the chaotic parameter. To incorporate the chaotic map into the modified Coulomb's constant, a normalization process is employed. The normalized value of $K_{1}(\l)$ is evaluated within the intervals $[r_{1},r_{2}]$ and $[0,g(\l)]$ (Eq.~(\ref{norm})):
	
	\begin{equation}\label{norm}
	K_{norm}(\l)= \frac{(K_{1}(\l)-r_{1})g(\l)}{r_{2}-r_{1}},
	\end{equation}
	
	Here, $K_{1}(\l)$, $r_{1}$, and $r_{2}$ are the chaotic value and the range of the chaotic map, respectively. The value of $g(\l)$ is defined in Eq.~(\ref{norm1}):
	
	\begin{equation}\label{norm1}
	g(\l)=a+\gamma(\l),
	\end{equation}
	
	Where $\gamma(\l)=\frac{-\l}{\l_{max}}(a-b)$, and $a$ and $b$ are user-defined values $(a>b)$. The suggested values of $a$ and $b$ in~\cite{mirjalili2017chaotic} are 20 and $1e^{-10}$, respectively. Combining these concepts into one equation, the chaotic hierarchical Coulomb's constant is introduced in Eq.~(\ref{final_K}):\begin{equation}\label{final_K}
	K_{final}(\l)=K_{norm}(\l) + K_{h}(\l).
	\end{equation}
	
The proposed Coulomb's constant is an amalgamation of a normalized chaotic map and the restructured hierarchical Coulomb's constant. This combination is designed to enhance the exploration rate, particularly in the initial stages of execution. 
\subsubsection{Difference between the original and the proposed  Coulomb's constants}
\begin{figure}[h]
		\begin{subfigure}[b]{0.49\linewidth}
			\centering
			\includegraphics[width=1\linewidth]{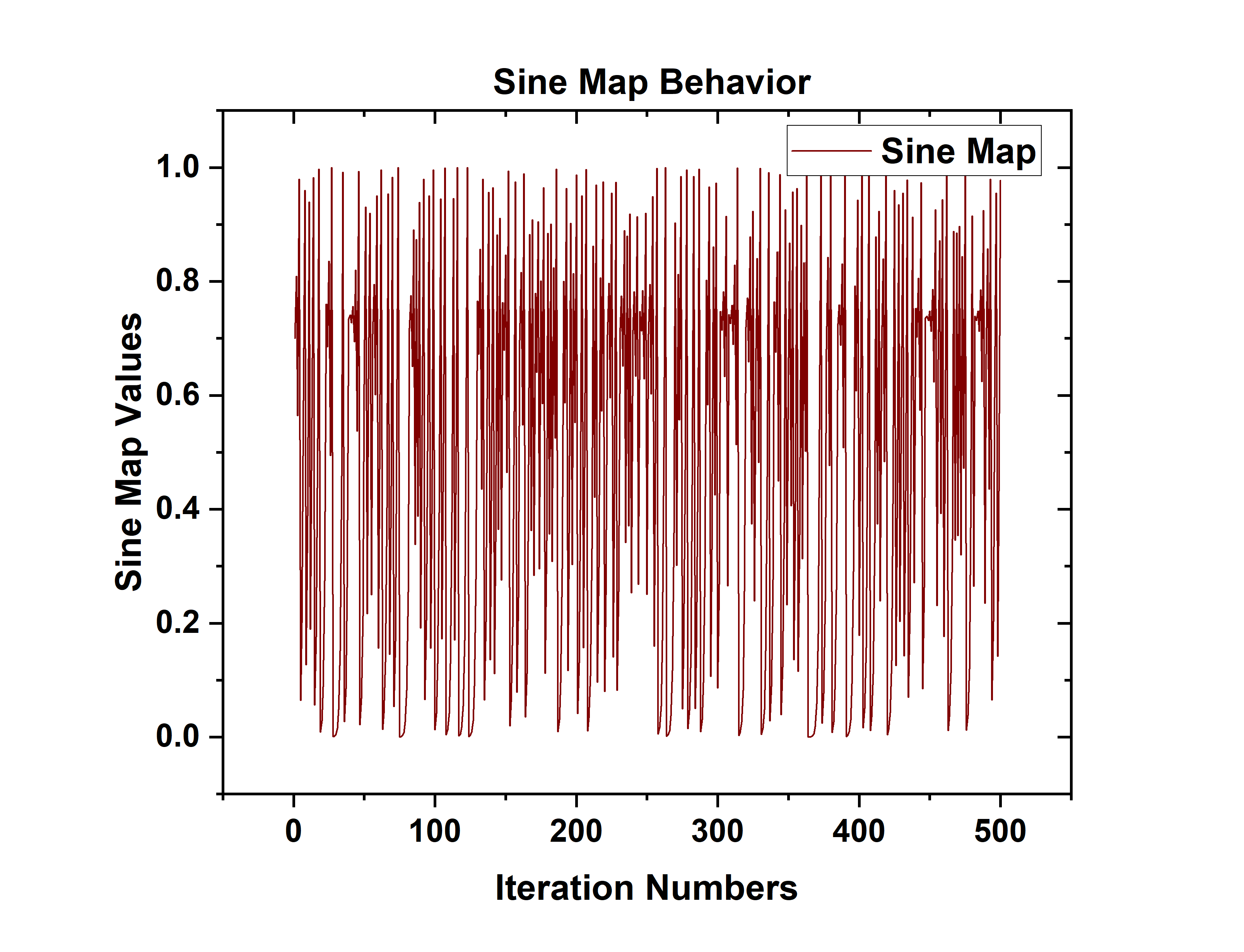}
			\caption{ {A}
				\label{sine}}
		\end{subfigure}
		\begin{subfigure}[b]{0.49\linewidth}
			\centering
			\includegraphics[width=1\linewidth]{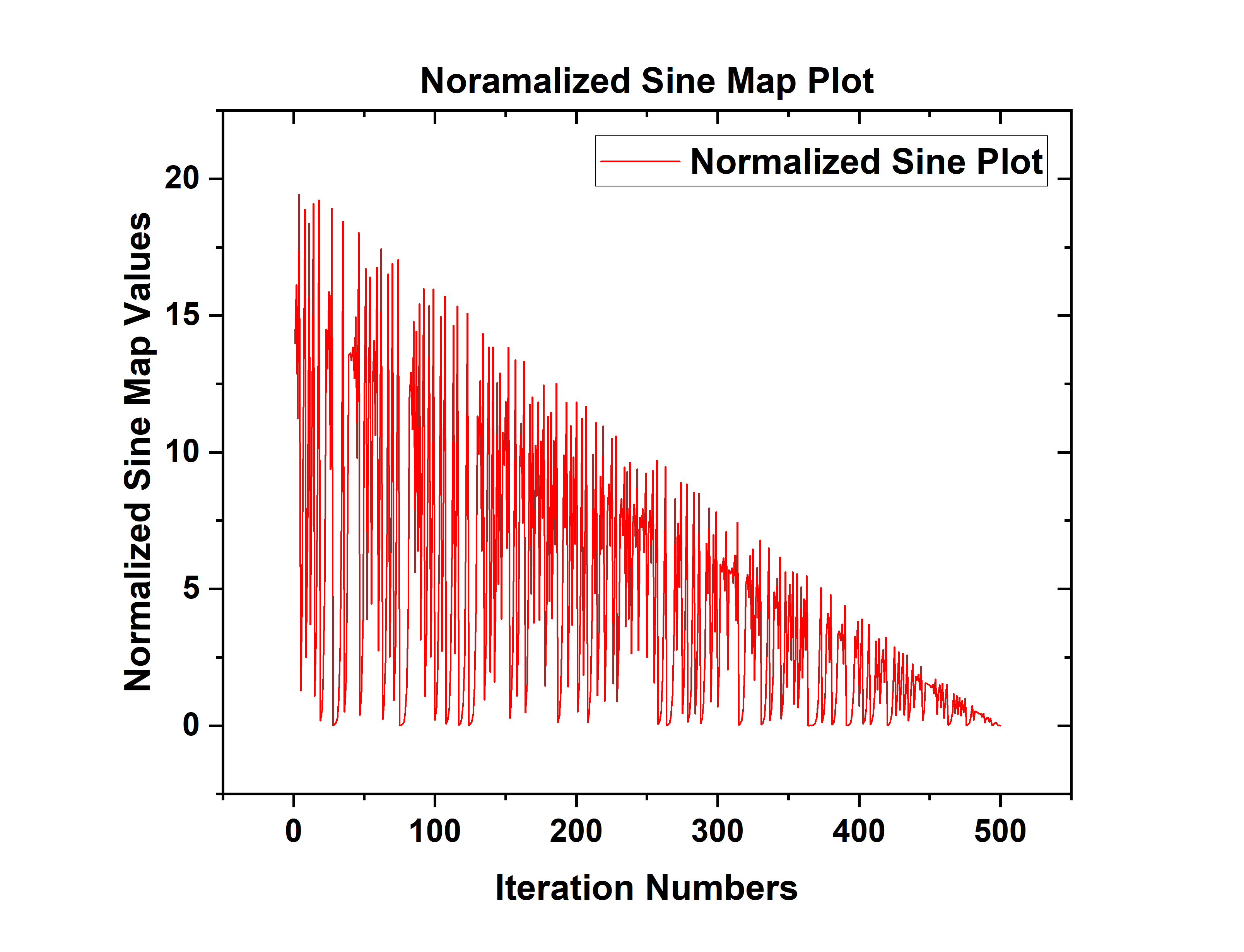}
			\caption{{B}
				\label{normalized}}
		\end{subfigure}
		\begin{subfigure}[b]{0.49\linewidth}
			\centering
			\includegraphics[width=1\linewidth]{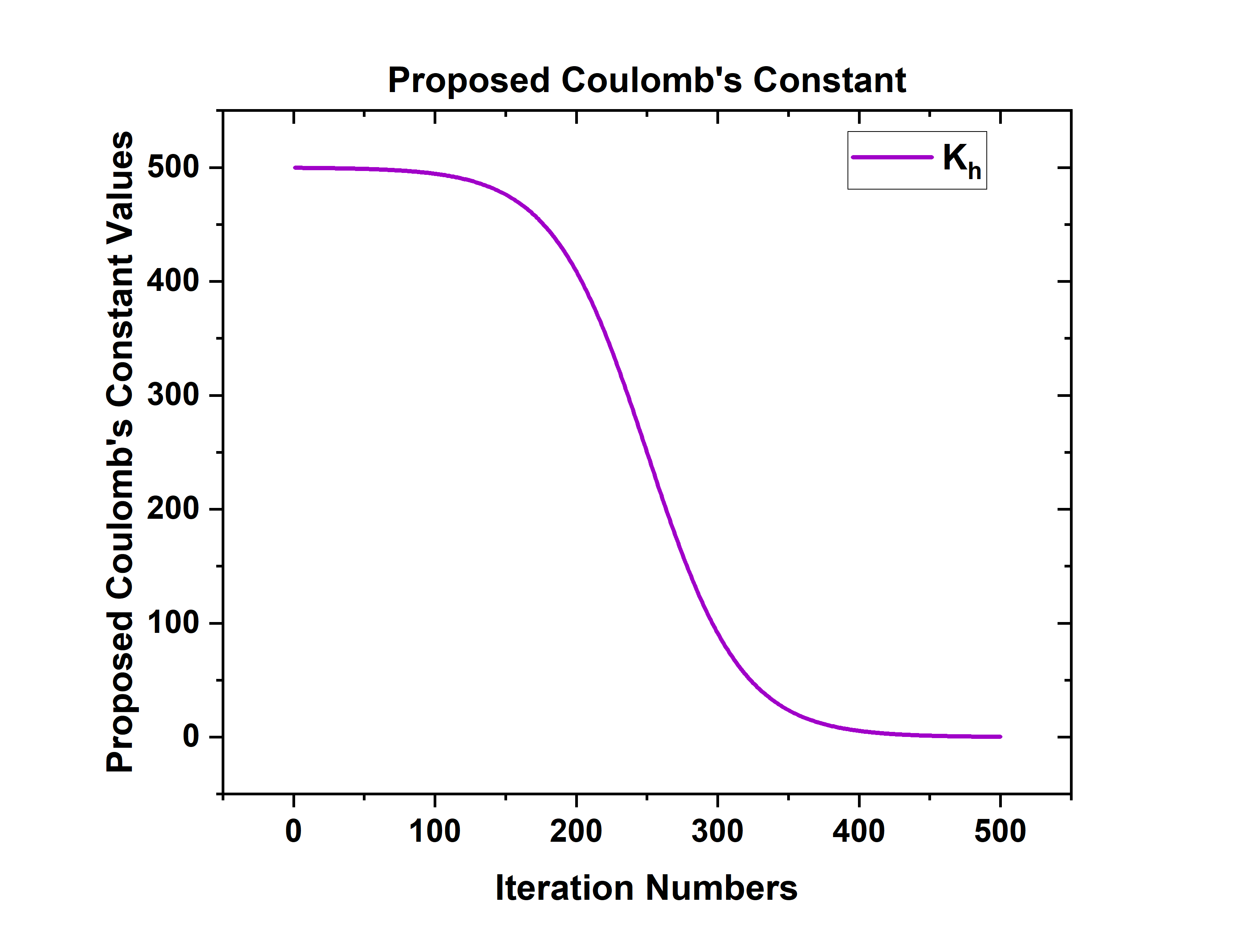}
			\caption{{C}
				\label{proposedcoulomb}}
		\end{subfigure}
		\begin{subfigure}[b]{0.49\linewidth}
			\centering
			\includegraphics[width=1\linewidth]{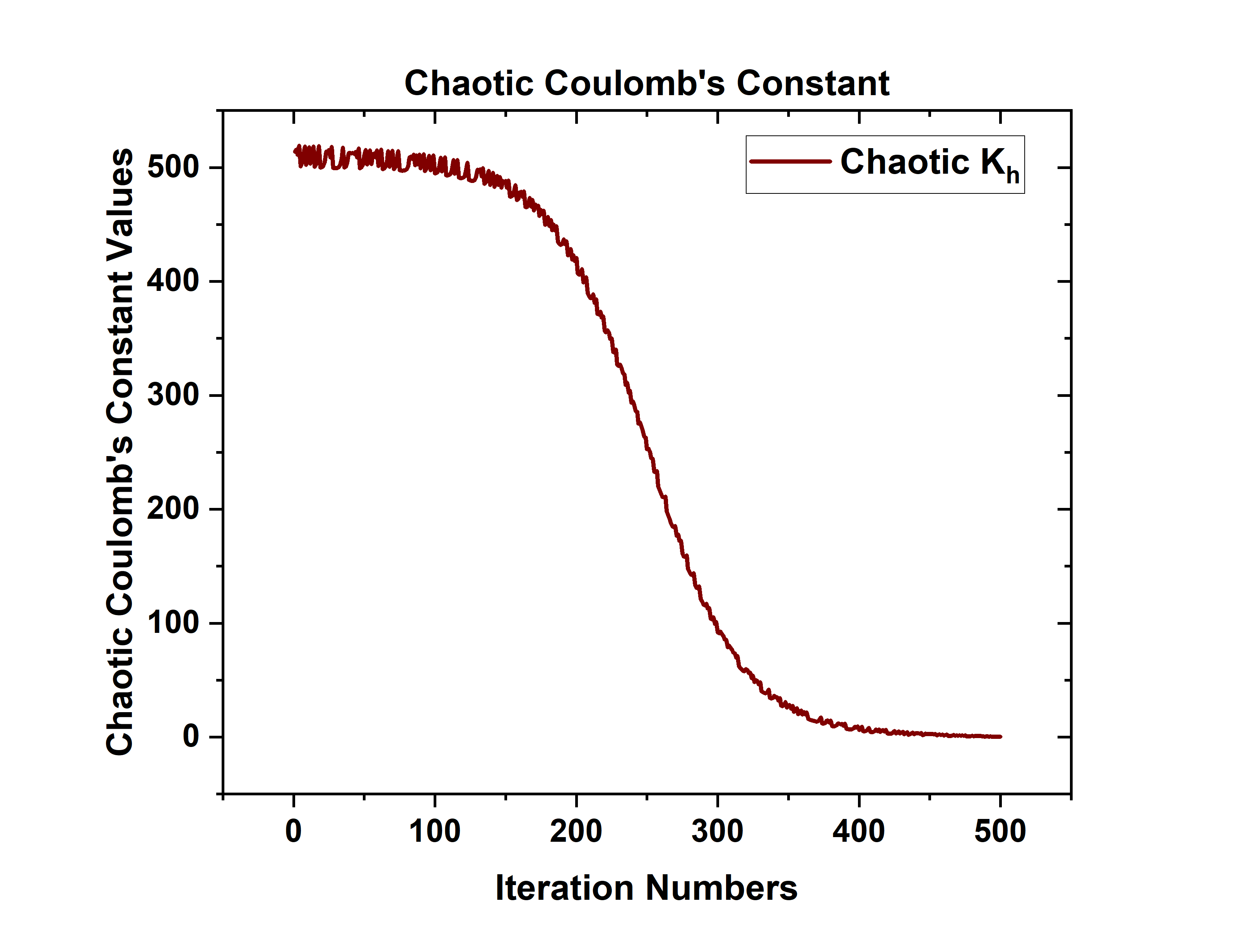}
			\caption{{D}
				\label{chaoticcoulomb}}
		\end{subfigure}
		\caption{{Procedure for embedding the chaotic map into the proposed Coulomb's constant as follows: Figs: A $\rightarrow$ B$\rightarrow$ C $\rightarrow$ D.}}\label{chaotic_K}
	\end{figure}

The suggested Coulomb's constant $K_{h}(\l)$ exhibits a distinct pattern from the original AEFA constant $K(\l)$, as illustrated in Fig. \ref{fig: sigmoidgraph} for $\l_{max}=500$, $\beta=3$, $\delta=100$, and $K_{0}=500$. In this figure, it is evident that the value of $K(\l)$ experiences a rapid decline before reaching 100 iterations, indicating the limited exploration capability of AEFA. In contrast, the proposed $K_{h}(\l)$ maintains a high value for half of the iterations before gradually decreasing and approaching zero. This modification ensures that AEFA possesses a robust exploration ability in the early stages, allowing sufficient time for searching for an approximate solution that can be further refined during the exploitation phase. Consequently, all adjustments to $K(\l)$ are centered on enhancing the value of $K_{h}(\l)$ in the initial iterations, implying that the suggested $K_{h}(\l)$ can improve AEFA's exploration capability. Moreover, the process of incorporating the chaotic map into $K_{h}(\l)$ is illustrated in Fig. \ref{chaotic_K}. In this figure, the chaotic map is initially added to the normalized curve (A $\rightarrow$ B). Subsequently, the resulting curve is combined with the proposed hierarchical Coulomb's constant curve (B $\rightarrow$ C), yielding the final chaotic curve (C $\rightarrow$ D).

\begin{figure}[htbp]
		\begin{center}
			\begin{tikzpicture}[node distance=1.5cm,auto]
			\tikzset{->-/.style={decoration={
						markings,
						mark=at position #1 with {\arrow{>}}},postaction={decorate}}}
			\node at (-0.8,19.5)[round](star){{Start algorithm}};
			\node [startstop, right of=star]at (2.5,19.5)(init){Initialization};
			\node[startstop, right of=init]at (7,19.5)(fitness){{Calculate fitness}};
			\node[startstop, right of=fitness]at (12,19.5)(global){{Calculate global best}};
			\node[startstop,below of=global]at (11.5,18.5)(constant){{Calculate Coulomb's constant Eq.~\ref{final_K}}};
			\node[startstop,left of=constant]at (6.5,17)(force){{Calculate force}};
			\node[startstop,left of=force]at(1,17)(acc){{Calculate acceleration}};
			\node[startstop,below of=acc](velocity)at(-0.5,16){Update velocity by Eq.~\ref{AEFAvelocity}};
			\node[startstop,right of=velocity ](velbounds)at (5,14.5) {{Bounds velocity}};
			\node[startstop,right of=velbounds](position)at (11,14.5) {{Update position, Eq.~(\ref{AEFAposition})}};
			\node[startstop,below of=position](posbounds)at (11,13.5) {{Drag back agents into their search region}};
			\node[startstop,left of=posbounds](stop)at (4.5,12){{Stopping criteria $?$}};
			\node[startstop,below of=stop](end)at (3,11){{End algorithm}};
			\draw[arrow,thick,line width=2pt](star)--(init);
			\draw[arrow,thick,line width=2pt](init)--(fitness);
			\draw[arrow,thick,line width=2pt](fitness)--(global);
			\draw[thick,line width=2pt](13.5,19)--(13.5,18.3);
			\draw[thick,line width=2pt](13.5,18.3)--(11.5,18.3);
			\draw[arrow,thick,line width=2pt](11.5,18.3)--(11.5,17.5);
			\draw[arrow,thick,line width=2pt](constant)--(force);
			\draw[arrow,thick,line width=2pt](force)--(acc);
			\draw[arrow,thick,line width=2pt](acc)--(velocity);
			\draw[arrow,thick,line width=2pt](velocity)--(velbounds);
			\draw[arrow,thick,line width=2pt](velbounds)--(position);
			\draw[thick,line width=2pt](12.5,14)--(12.5,13);
			\draw[thick,line width=2pt](12.5,13)--(11,13);
			\draw[arrow,thick,line width=2pt](11,13)--(11,12.5);
			\draw[arrow,thick,line width=2pt](posbounds)--(stop);
			\draw[arrow,thick,line width=2pt](stop)--(end);
			\draw[thick,line width=2pt](1.3,12)--(-3,12);
			\draw[thick,line width=2pt](-3,12)--(-3,18.3);
			\draw[thick,line width=2pt](-3,18.3)--(8.5,18.3);
			\draw[arrow,thick,line width=2pt](8.5,18.3)--(8.5,19);
			\node[arrow,left,thick,blue]at(4.3,11){Yes};
			\node[arrow,above,thick,blue]at(0,12.2){No};
			\node[arrow,rotate=90, right,blue]at (-3.5,14){Repeat process};
			\end{tikzpicture}
		\end{center}
		\caption{{Flowchart of the proposed algorithm.} }
		\label{flowchart-constrainedAI-AEFA}
	\end{figure}
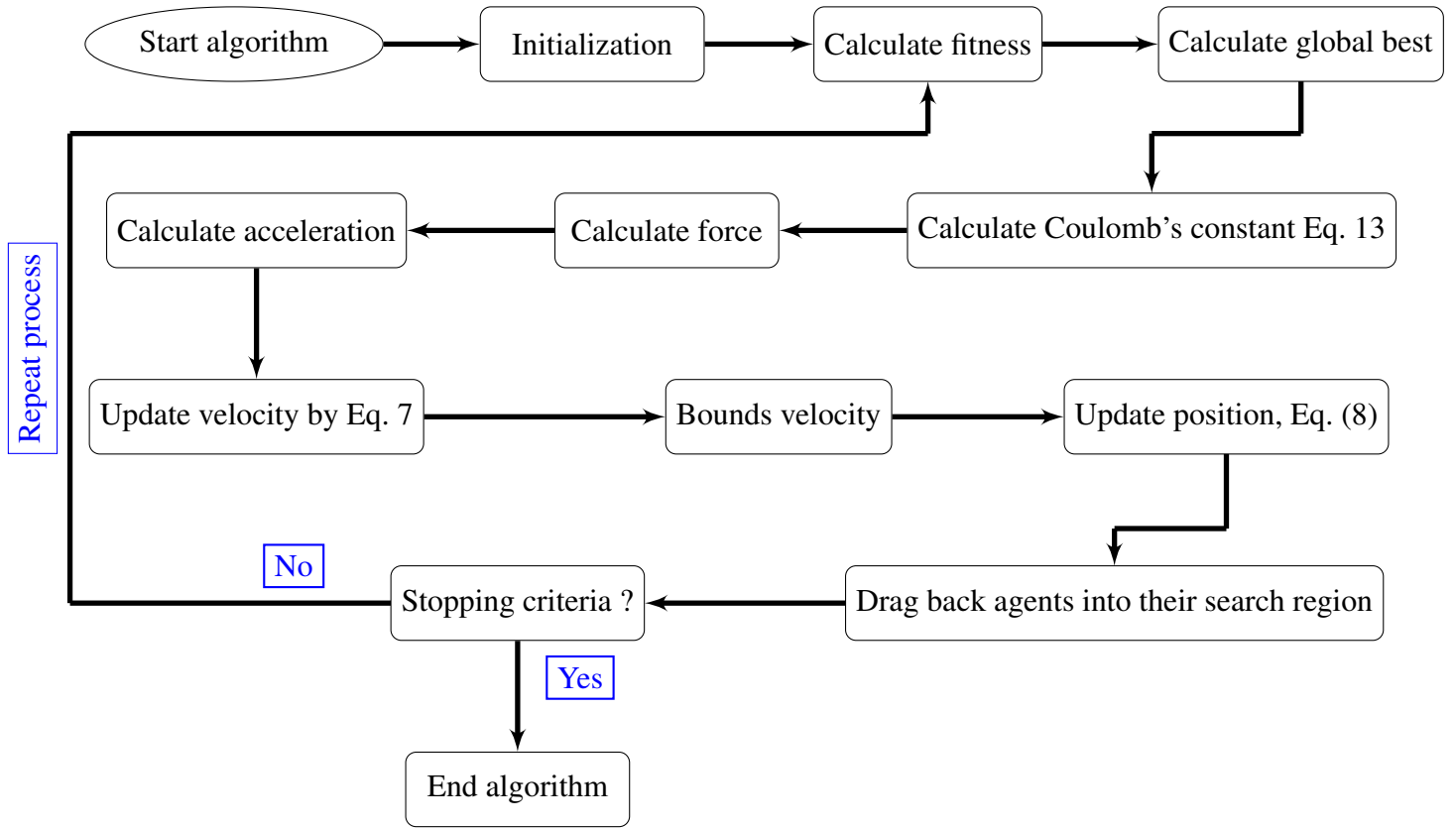

	The working procedure of the proposed algorithm is elucidated in Algorithm~\ref{pseudoAEFAHK}, and the flowchart is depicted in Fig.~\ref{flowchart-constrainedAI-AEFA}.
	\begin{algorithm}[h]
		\caption{: {Working procedure of the proposed algorithm}}\label{pseudoAEFAHK}
		\begin{algorithmic}[1]
			\STATE Initialize $K_{0}$, $a$, $b$, upper bound, lower bound, $\beta$, $N$, $D$, $\l_{max}$ and $\delta$,
			\STATE Generate the initial position and velocity for each agent,
			\FOR {$\l=1:\l_{max}$}
			\STATE	Calculate the fitness values for all agents,
			\STATE Sort the agents according to the rules which are given in section~\ref{constrainedhandling},
			\STATE Calculate the Coulomb's constant using Eq.~(\ref{sigmoidcoulomb}),
			\STATE Embed the chaotic map in the Coulomb's constant using Eqs.~(\ref{sinemap})$-$~(\ref{norm1}), and the final Coulomb's constant using Eq.~(\ref{final_K}), 
			\STATE \textbf{Calculate} the charge on the each agent using Eq.~\eqref{charge}, 
			\STATE Calculate the force and acceleration of each agent using Eqs.~\eqref{electrosticforce} and~\eqref{acceleration},
			\STATE Update the velocity for each agent using Eq.~(\ref{AEFAvelocity}),
			\STATE	{Apply velocity bounds}
			\STATE	If velocity of particle exceed the $v_{ub}$ is $\mathbf{v}_{i}(\l)=min(\mathbf{v}_{i}(\l), v_{ub})$,
			\STATE	If velocity of particle goes below the $v_{lb}$ is $\mathbf{v}_{i}(\l)=max(\mathbf{v}_{i}(\l), v_{lb})$,
			\STATE	Update position as $\mathbf{x}_{i}(\l+1)=\mathbf{x}_{i}(\l)+ \mathbf{v}_{i}(\l+1)$\\
			\STATE	{Apply position bounds},
			\STATE	If position of particle exceed the $ub$ is $\mathbf{x}_{i}(\l+1)=min(\mathbf{x}_{i}(\l+1), ub)$,
			\STATE	If position of particle goes below the $lb$ is $\mathbf{x}_{i}(\l+1)=max(\mathbf{x}_{i}(\l+1), lb)$,
			\IF {$\mathbf{fit}(\mathbf{x}_{i}(\l+1))<\mathbf{fit}(\mathbf{x}_{i}(\l))$} 
			\STATE	$\mathbf{fit}(\mathbf{x}_{i}(\l))=\mathbf{fit}(\mathbf{x}_{i}(\l+1))$,
			\ENDIF
			\STATE Apply feasibility rules which are given in section~\ref{constrainedhandling},
			\ENDFOR
			\RETURN the best search agent and the best solution.
		\end{algorithmic}
	\end{algorithm}
	\section{Constrained handling technique}\label{constrainedhandling}
This article employs a parameter-free constraint-handling technique to address COPs. In this technique, the degree of violation is quantified as the sum of equality and inequality constraints, divided by the total number of constraints.\begin{equation}\label{violation}
	vio=(\sum_{i}^{k}G_{i}(\mathbf{x})+\sum_{j}^{m}H_{j}(\mathbf{x}))/n,
	\end{equation}
	where $n=k+m$ is total number of constrained involving in the problem, $G_{i}(\mathbf{x})$ and $H_{j}(\mathbf{x})$ are inequality and equality constraints given as follow:
	\[G_{i}(\mathbf{x})=\begin{cases}
	g_{i}(\mathbf{x}),&\text{if } g_{i} (\mathbf{x})>0,\\
	0, & \text{if  } g_{i}(\mathbf{x})\leq 0,
	\end{cases}
	\quad
	\text{and}\hspace{0.5cm}
	H_{j}(\mathbf{x})=\begin{cases}
	|h_{j}(\mathbf{x})|,&\text{if } |h_j (\mathbf{x})|-\epsilon>0,\\
	\hspace{0.5cm}0, & \text{if  } |h_{j}(\mathbf{x})|-\epsilon\leq 0,
	\end{cases}
	\] 
	Here $\epsilon$ is a very small tolerance number that is used to convert equality constraints into inequality. In article~\cite{wu2017problem}, a sorting rule is given as:\begin{enumerate}[(i)]
		\item Sort feasible agents in comparison to infeasible.
		\item According to their fitness values, feasible agents are sorted in ascending order.
		\item As per their violation mean values, infeasible agents are sorted in ascending order. 
	\end{enumerate}
	\begin{figure}[h]
		\centering
		\includegraphics[scale=0.6]{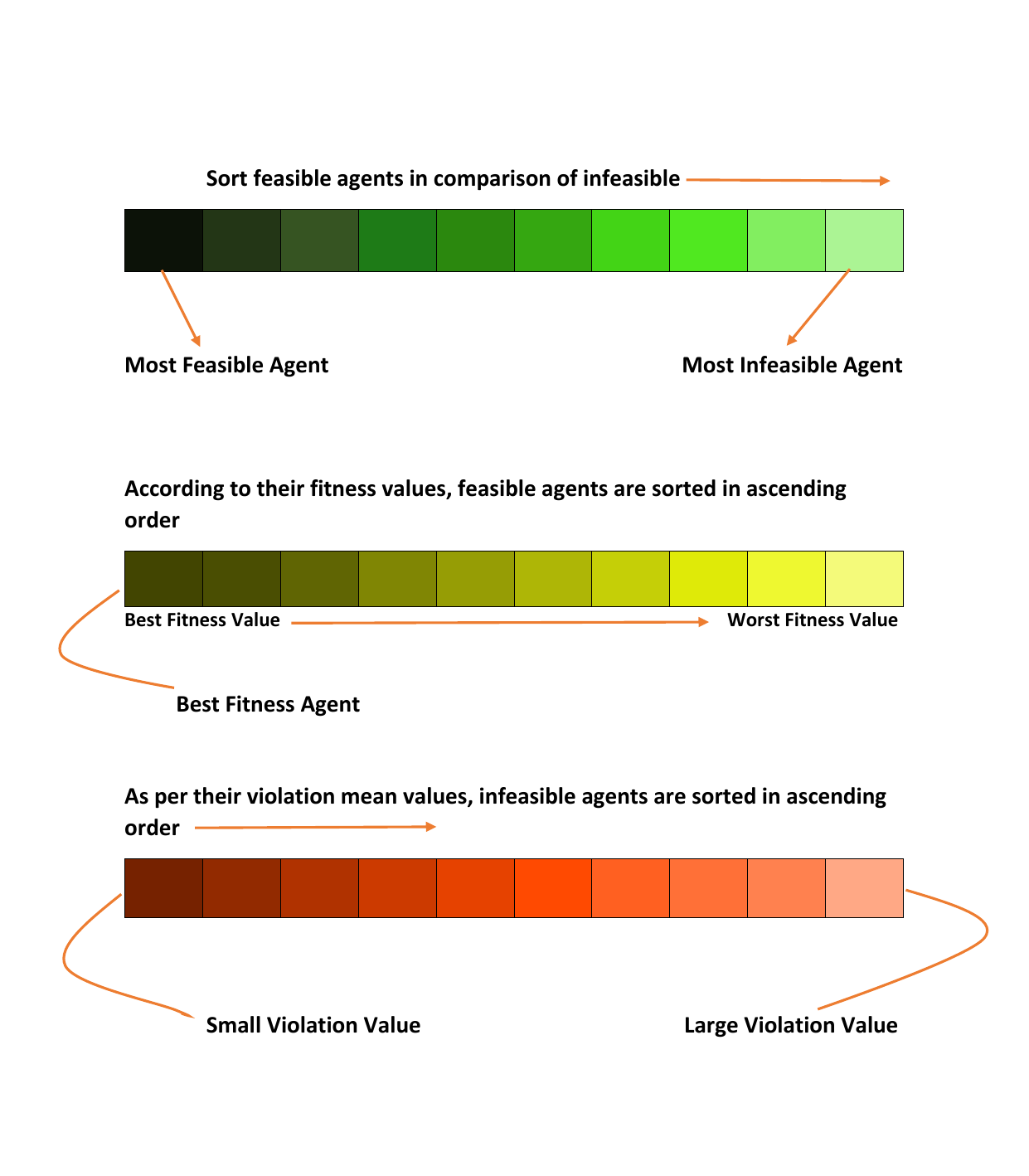}
		\caption{Schematic procedure diagram of sorting agent in the proposed algorithm.}\label{sorting}
	\end{figure}
	
	After sorting the agents in the population, two solutions are compared as: \begin{enumerate}[(a)]
		\item More priority is given to feasible agents in comparison to infeasible agents.
		\item In the population, if all agents are satisfying all constraints, then select all those agents that have better fitness values.
		\item If there is no feasible agent present in the population, more priority is given to all those agents that have smaller violation values.\\
		The schematic procedure of sorting agents is presented in Fig~\ref{sorting}.
	\end{enumerate}
\section{Experimental analysis of the proposed AI-AEFA $vs.$ AEFA-C and AEFA}\label{AEFAHKvsAEFA}
	
	The core principle of the proposed algorithm aligns with AEFA and AEFA-C, with the primary distinction lying in the definition of Coulomb's constant. This variation aims to enhance solution quality concerning optimality, exploration, and feasibility, incorporating the concept of bounds to prevent agents from exceeding their search range. These modifications are seamlessly integrated into the original AEFA framework, and their impact on AEFA and AEFA-C is thoroughly justified in this section.
	
\begin{figure}
		\centering
		\begin{subfigure}[b]{1\linewidth}
			
			\includegraphics[width=0.49\linewidth]{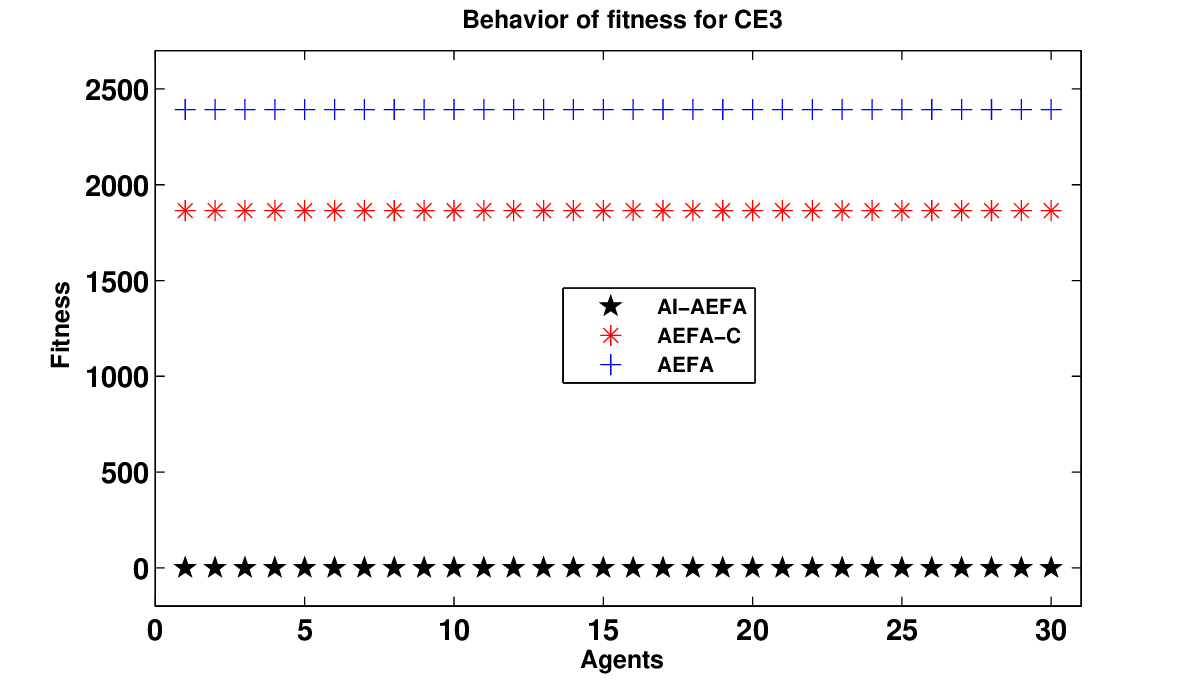}
			\includegraphics[width=0.49\linewidth]{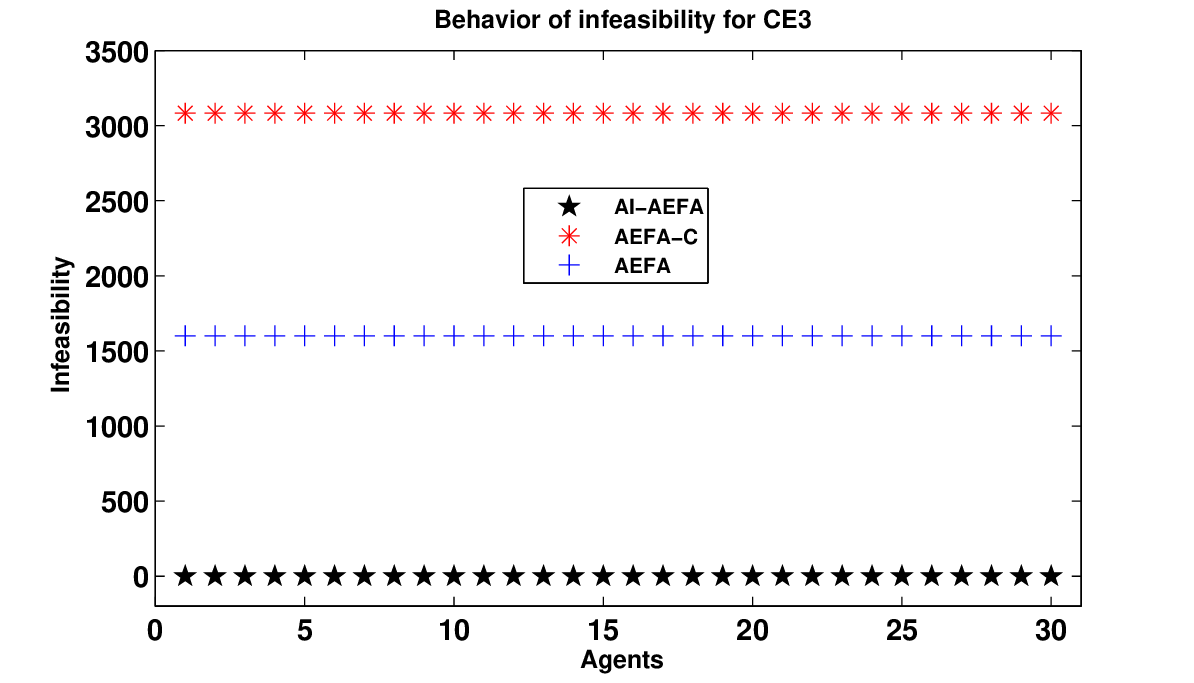}
			\includegraphics[width=0.49\linewidth]{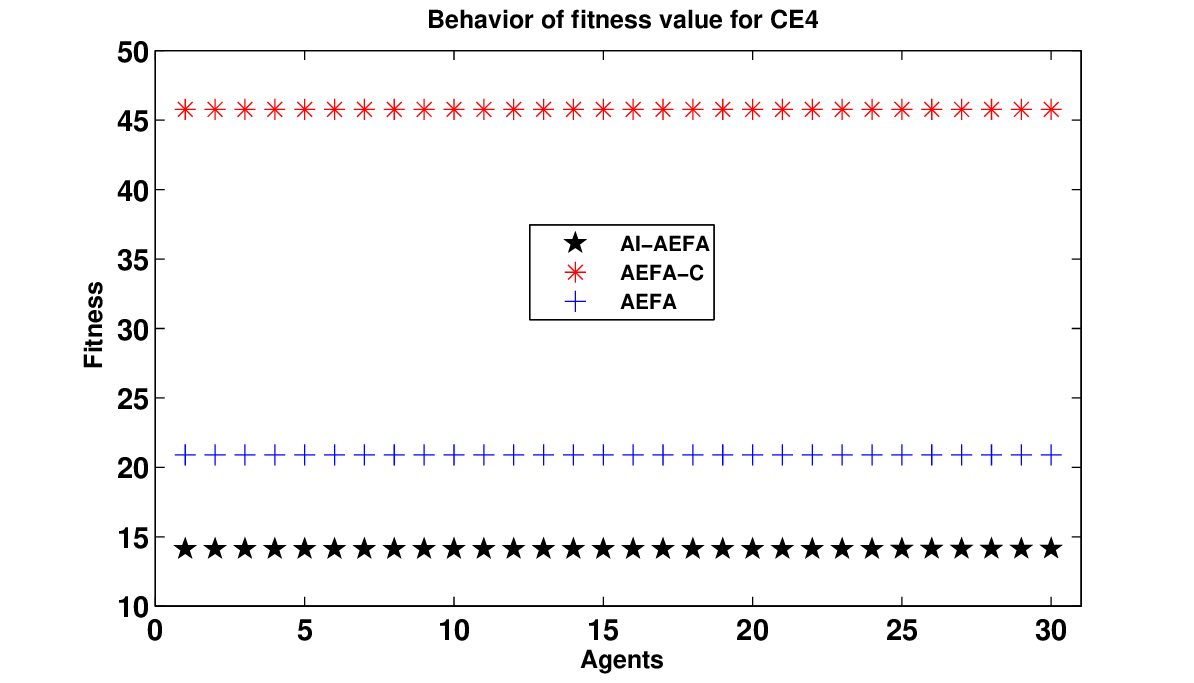}
			\includegraphics[width=0.49\linewidth]{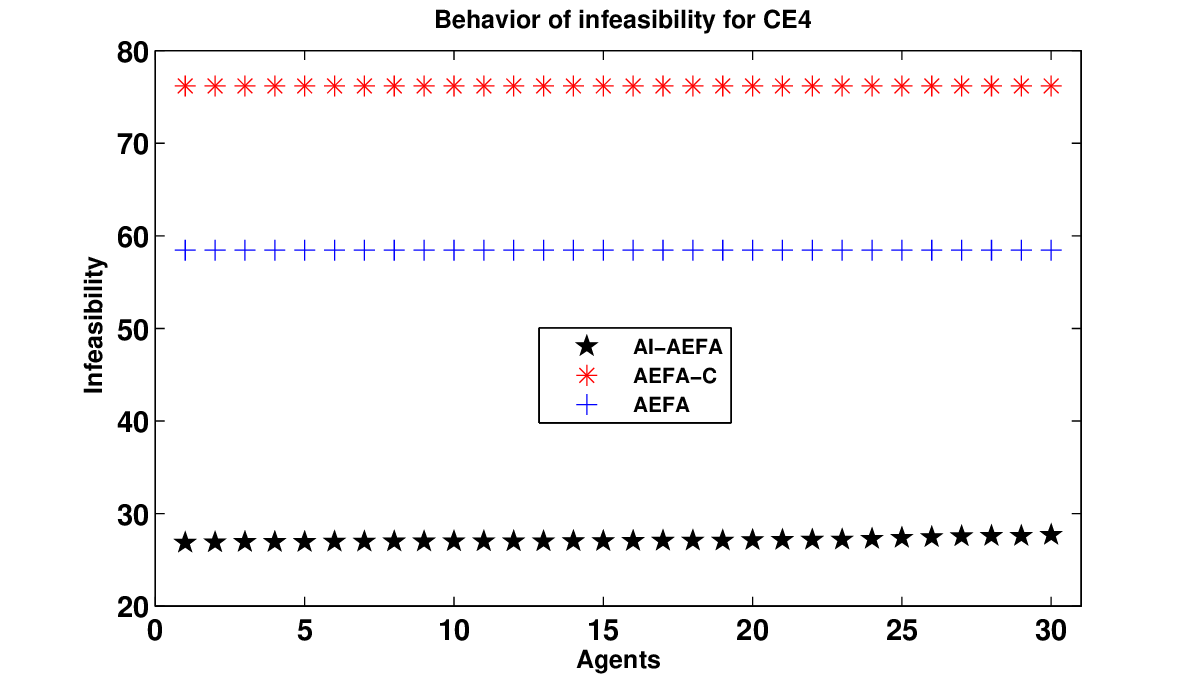}
			\includegraphics[width=0.49\linewidth]{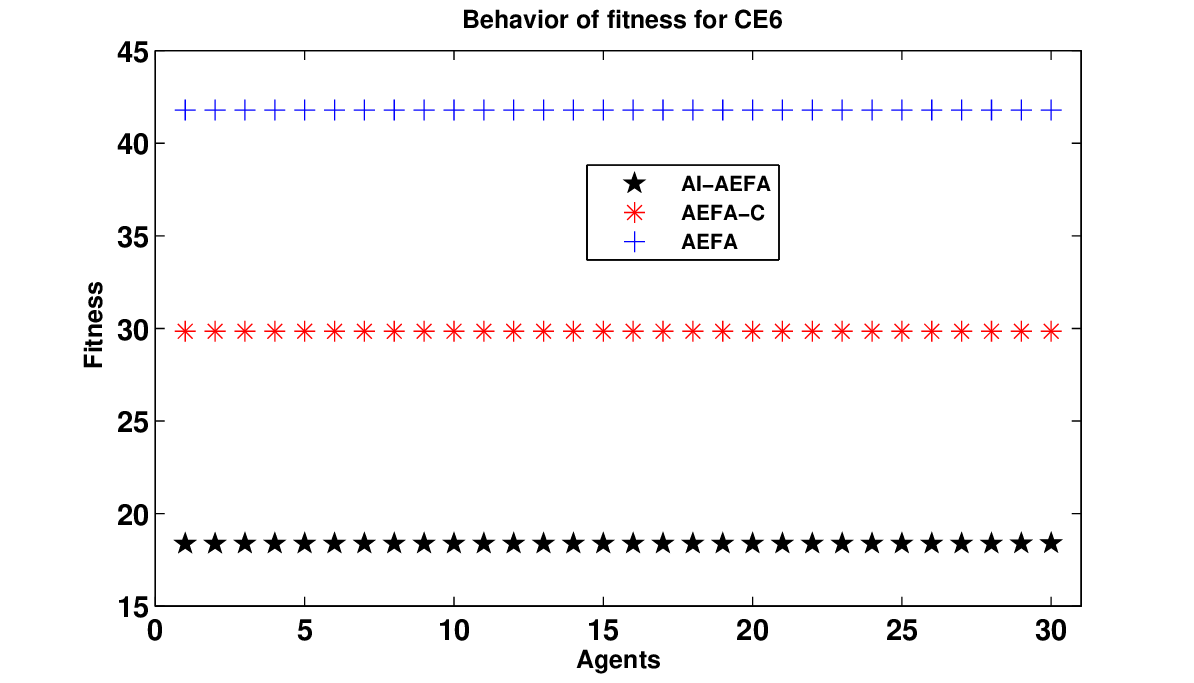}
			\includegraphics[width=0.49\linewidth]{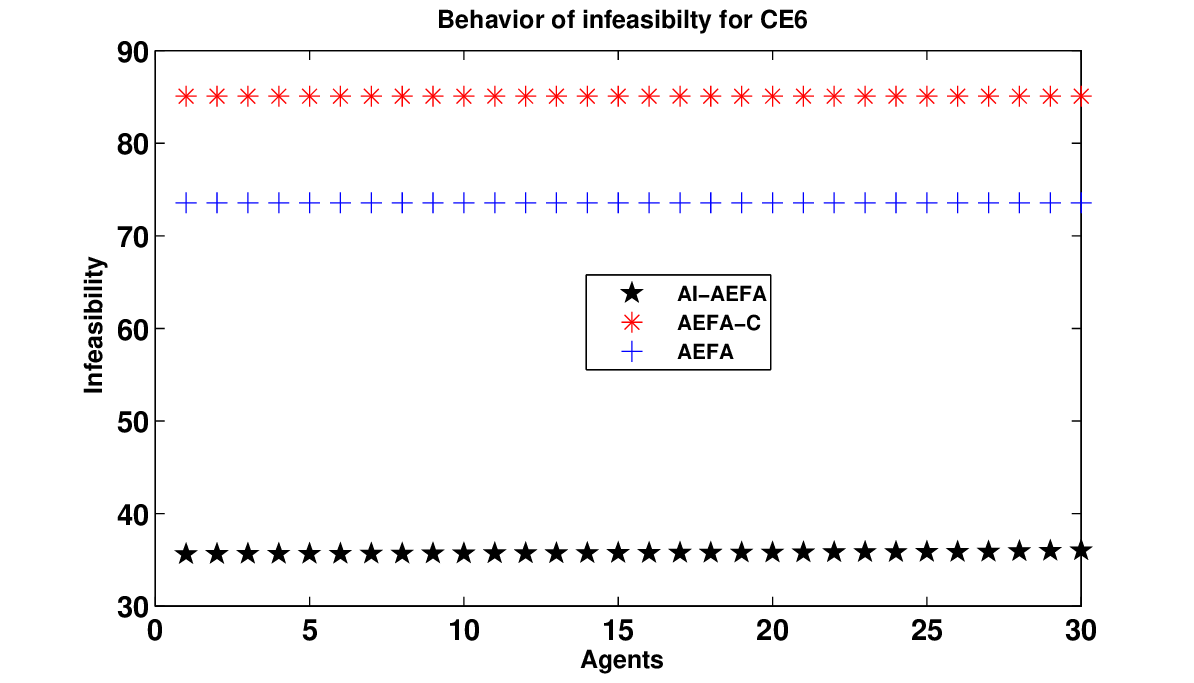}
			\includegraphics[width=0.49\linewidth]{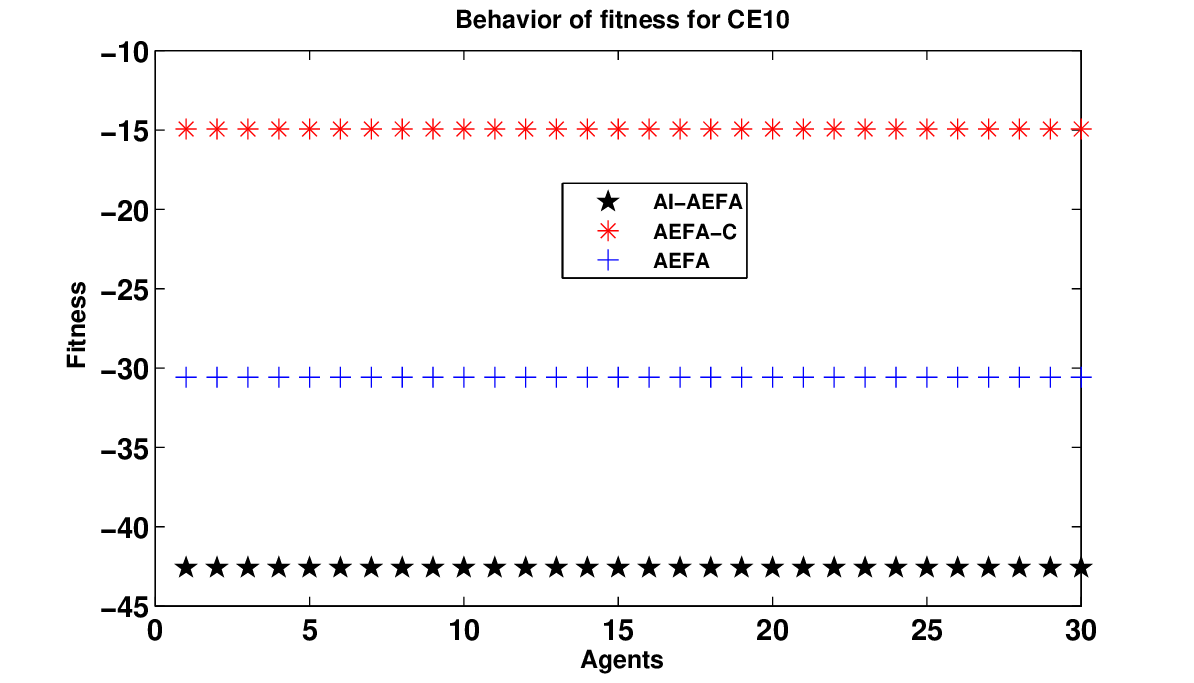}
			\includegraphics[width=0.49\linewidth]{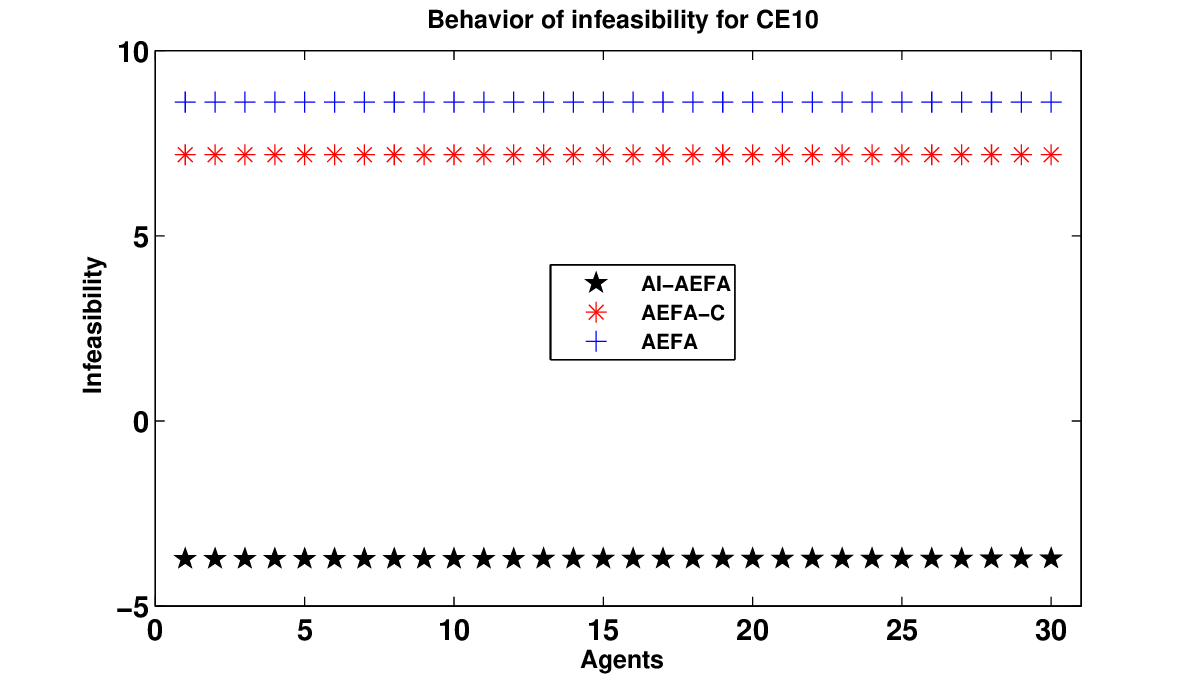}
		\end{subfigure}
		\caption{{Behavior of fitness value and infeasibility for last iteration}}\label{fig: fig1}
	\end{figure}

\begin{figure}
		\centering
		\begin{subfigure}[b]{1\linewidth}
			\includegraphics[width=0.49\linewidth]{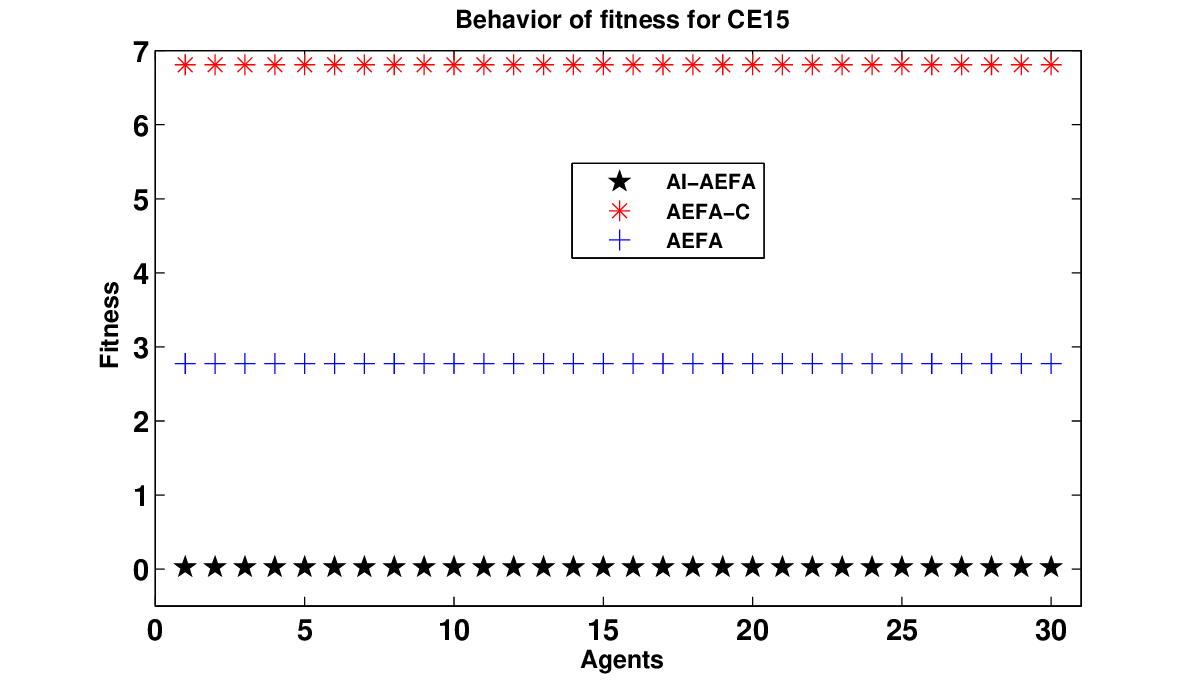}
			\includegraphics[width=0.49\linewidth]{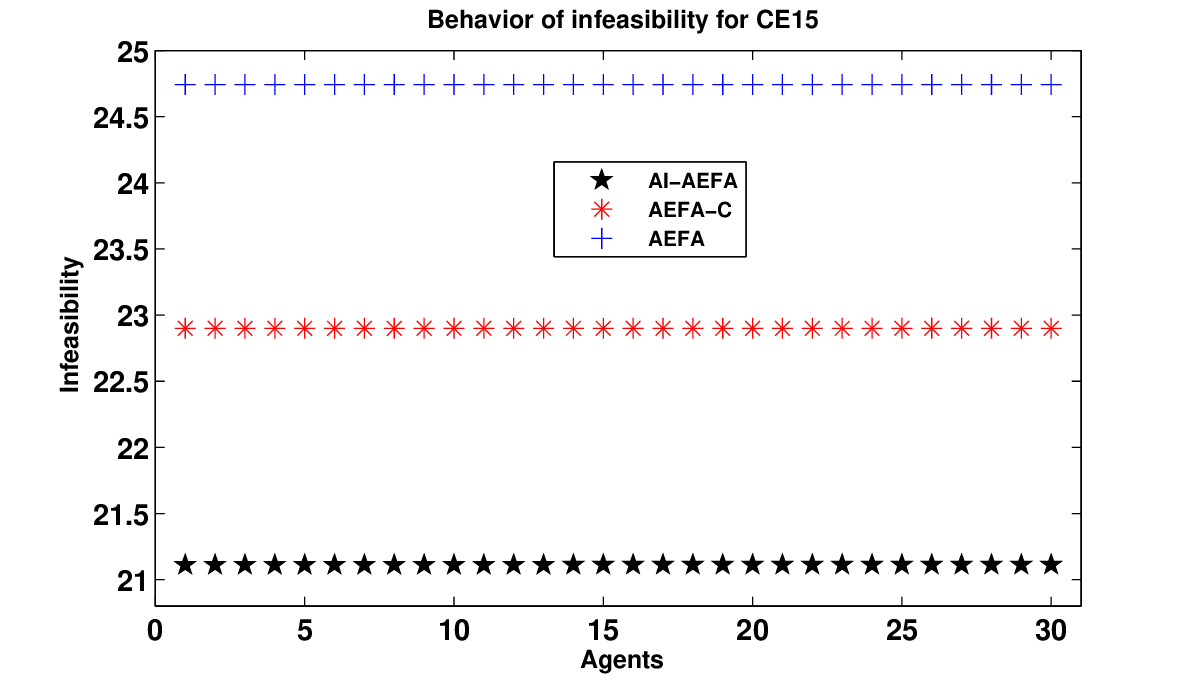}
			\includegraphics[width=0.49\linewidth]{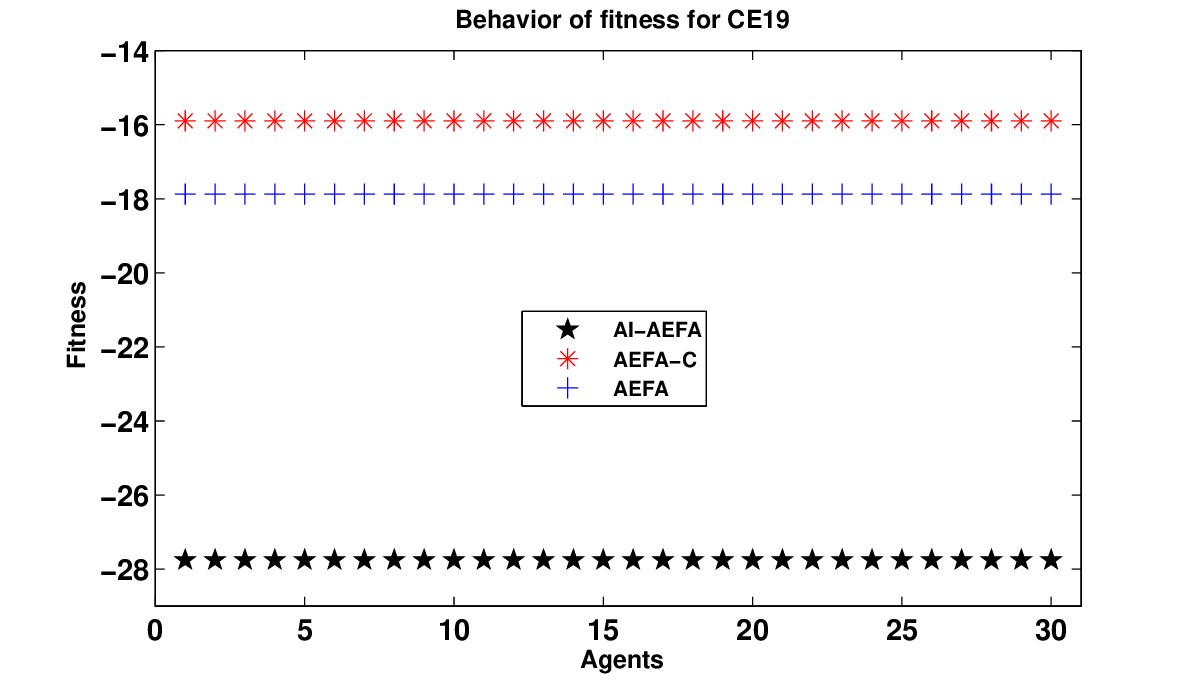}
			\includegraphics[width=0.49\linewidth]{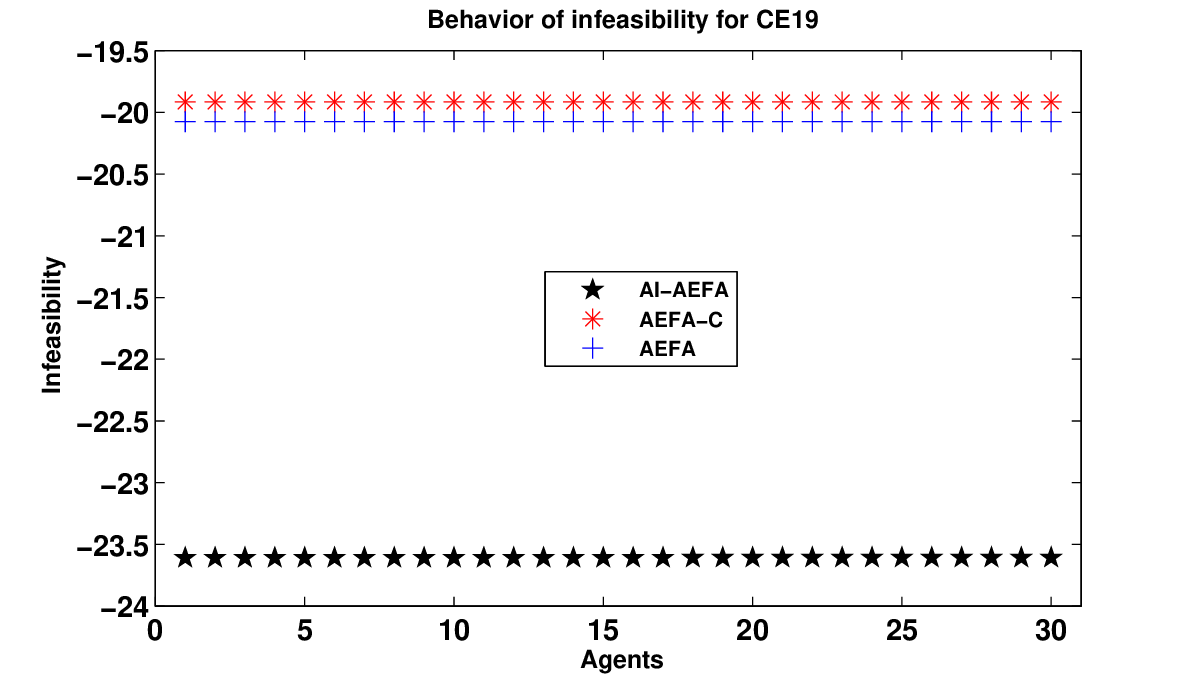}	\includegraphics[width=0.49\linewidth]{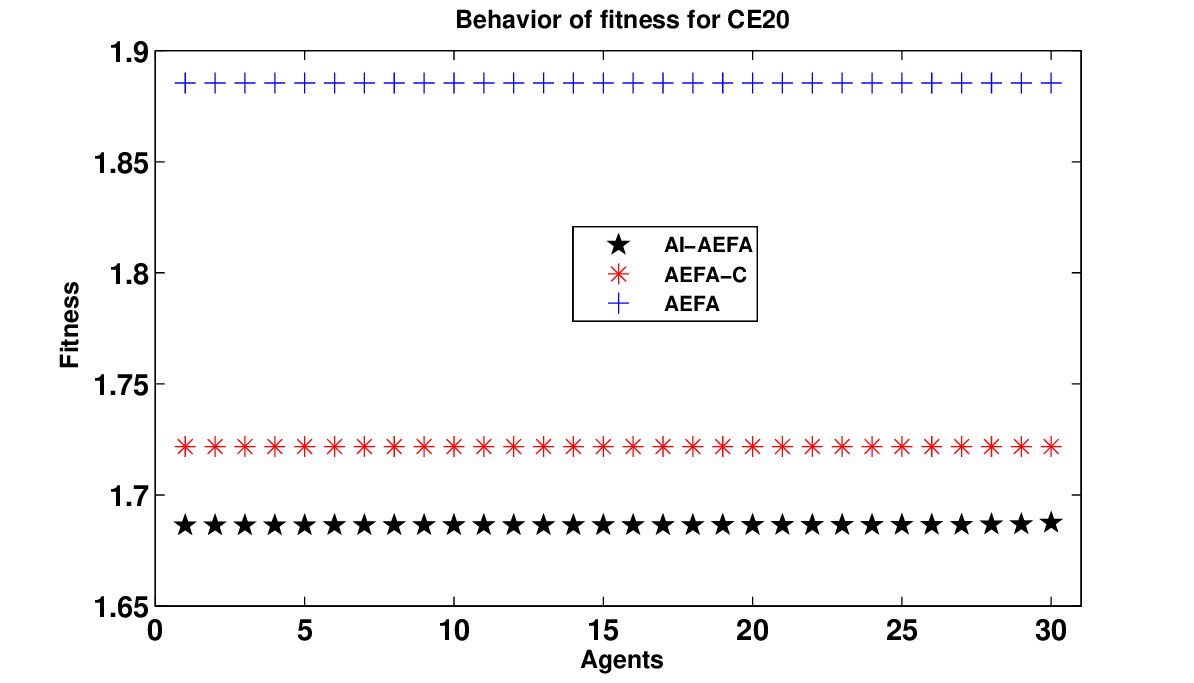}	\includegraphics[width=0.49\linewidth]{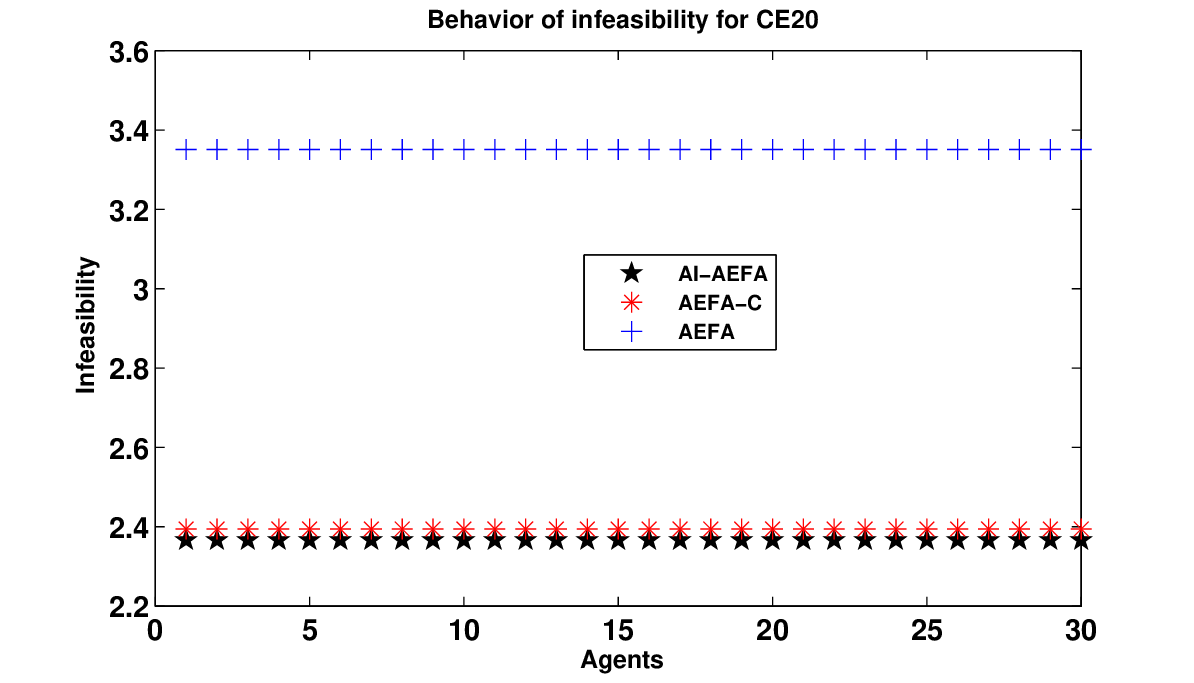}	\includegraphics[width=0.49\linewidth]{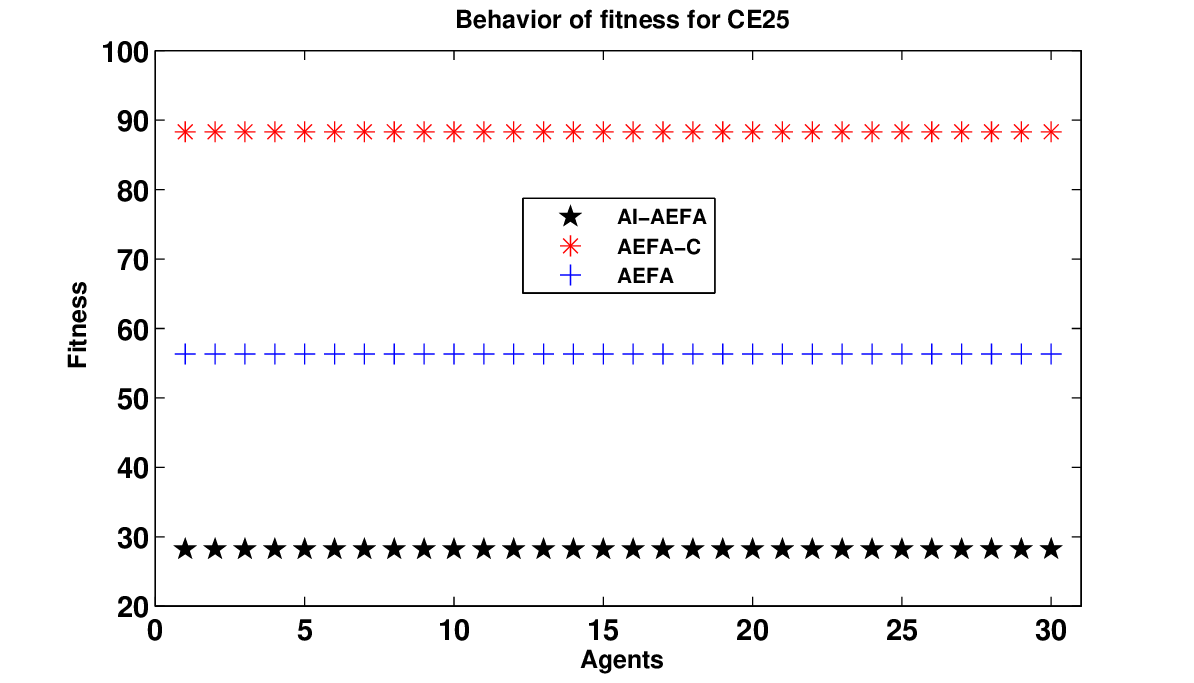}	\includegraphics[width=0.49\linewidth]{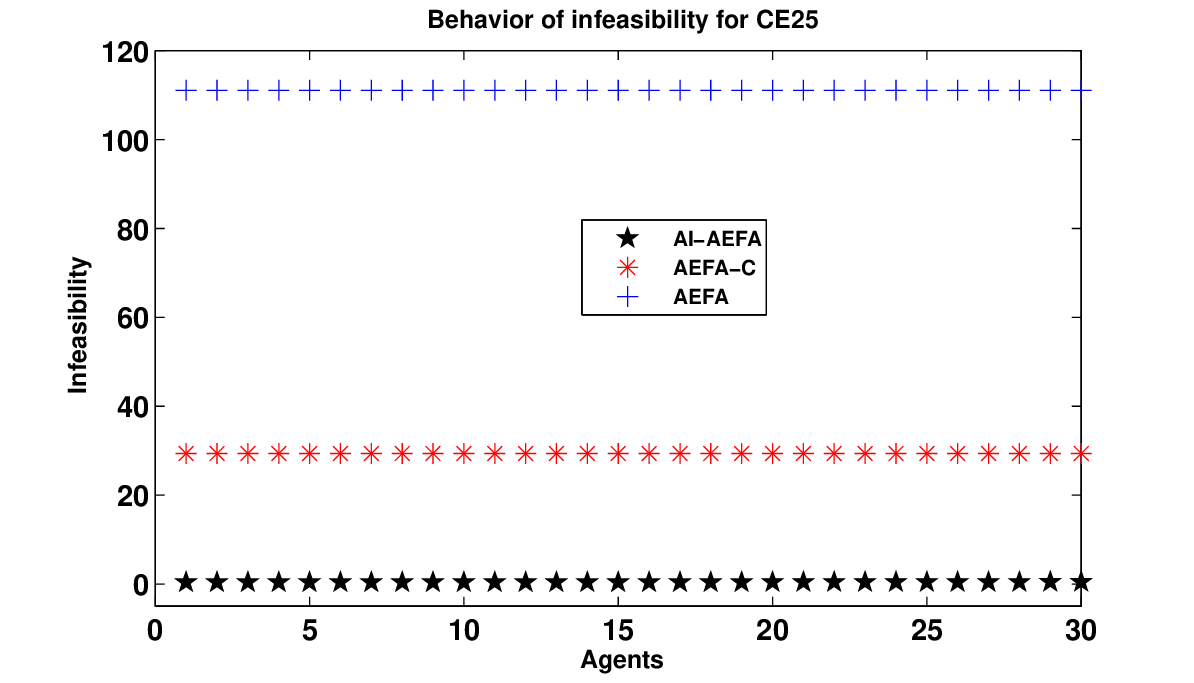}
		\end{subfigure}	
		\caption{{Behavior of fitness value and infeasibility for last iterations.}}\label{fig: fig2}
	\end{figure}

	To assess the performance of AI-AEFA, eight optimization problems (CE3, CE4, CE6, CE10, CE15, CE19, CE20, and CE25) are tackled by the proposed algorithm, alongside AEFA and AEFA-C. Experimental results are compared based on fitness values and infeasibility rates obtained in the last iterations for 30 dimensions. The infeasibility rate is calculated using Eq. (\ref{violation}). Figs. \ref{fig: fig1} and~\ref{fig: fig2} present these experimental results, revealing that AEFA and AEFA-C struggle to achieve optimal fitness values with a satisfactory feasible rate for the selected optimization problems. In contrast, the proposed algorithm efficiently solves these optimization problems with a reduced infeasibility rate.
	
	This experiment highlights that AEFA, without hierarchical Coulomb's constant and boundedness theory, may not effectively address constrained optimization problems with high feasibility rates. However, the proposed definition of Coulomb's constant significantly improves the performance of AEFA, yielding optimal results with a lower infeasibility rate compared to AEFA and AEFA-C. This outcome validates our approach, emphasizing that agents remain within their search range, providing a positive search direction. The proposed AI-AEFA demonstrates superior efficiency over both comparative schemes. In the subsequent sections, the optimization capabilities of AI-AEFA are further demonstrated on CEC 2017 constrained benchmark problems at different dimensions, real-world engineering scenarios, and RRA problems.
\section{Experimental results and discussions on CEC 2017 constrained test suite}\label{experimental}
	In this section, the performance of AI-AEFA is evaluated against other state-of-the-art algorithms using a set of benchmark problems from CEC 2017~\cite{wu2017problem}. All experiments are conducted on MATLAB (2013b) on a machine running Windows 10 with an Intel(R) Core(TM) i5 8265U CPU, featuring 8 logical processors. The experiments are repeated 20 times independently for reliability.
	
	The chosen benchmark problems in CEC 2017 encompass COPs with a combination of inequality and equality constraints. These problems are categorized based on their properties into separable, non-separable, and rotated, and they span a diverse range of search spaces. A concise overview of these optimization problems is provided in Table S1 of the supplementary file, where notations S, NS, R, and $D$ denote separable, non-separable, rotated, and the dimension of the test problems, respectively. It is worth mentioning that all figures and tables with the notation $S\#$ are included in the supplementary file, where $\#$ represents the corresponding figure or table number.
	\subsection{Parameter settings}
	In this section, twenty-eight COPs are solved at dimensions 10 and 30, respectively. The average fitness value (Mean), standard deviation (Std), and feasibility rate are obtained in 20 independent runs and compared with ten other state-of-the-art algorithms, including hybrid algorithms such as Coyote optimization algorithm (COA) \cite{pierezan2018coyote}, Crow search algorithm (CSA) \cite{askarzadeh2016novel}, Gravitational search algorithm (GSA) \cite{rashedi2009gsa}, Rat swarm optimization (RSO) \cite{dhiman2021novel}, Tunicate search algorithm (TSA) \cite{kaur2020tunicate}, Chimp optimization algorithm (ChOA) \cite{khishe2020chimp}, Black widow optimization algorithm (BWOA) \cite{hayyolalam2020black}, Constriction coefficient PSO and GSA (CPSOGSA) \cite{rather2021constriction}, AEFA~\cite{yadav2019AEFA}, and AEFA-C~\cite{yadav2020artificial}. The parameters of these optimization algorithms are kept the same as in their original work. The parameters of the proposed algorithm are presented in Table \ref{tab: tunedparameter}.
	\begin{table}[htbp]\centering
		\caption{{Parameter for the proposed algorithm.}}\label{tab: tunedparameter}
		\begin{tabular}{p{2.5cm}p{2.5cm}p{2.5cm}p{2.5cm}p{2.5cm}p{2.5cm}}\hline
			$K_{0}$&$\beta$&$\delta$&$N$&$d$&MaxFE\\\hline
			500&6&300&30&10&15,000\\\hline
			500&6&300&30&30&30,000\\\hline
		\end{tabular}
	\end{table} 
	\begin{table}
		\caption{Experimental results of all comparison algorithms along with feasibility rate, 30 $d$.}\label{tab: finaldata1-30D}
		\resizebox{1\linewidth}{!}{\begin{tabular}{llccccccccccc}\hline
				Funcs	&	Metrics	&		COA	&		CSA	&		GSA	&	RSO	&	TSA	&		ChOA	&	BWOA	&		AEFA	&		AEFA-C	&		CPSOGSA	&		AI-AEFA	\\\hline
				&	Mean	&		7.397E+03	&		3.994E+01	&		3.316E+03	&	1.058E+04	&	5.872E+03	&		8.252E+03	&	4.060E+04	&		6.166E+03	&		5.927E+03	&		4.747E+04	&	\cellcolor{gray}	1.447E+01	\\
				CE1	&	Std	&		9.707E+02	&		2.142E+01	&		7.068E+02	&	1.108E+03	&	1.821E+03	&		0.000E+00	&	1.625E+04	&		2.141E+03	&		1.314E+03	&		8.090E+03	&		1.045E+01	\\
				&	FR	&		1.000E+02	&		4.000E+00	&		1.000E+02	&	1.000E+02	&	1.000E+02	&		7.600E+01	&	1.000E+02	&		1.000E+02	&		1.000E+02	&		0.000E+00	&		1.000E+02	\\\hline
				&	Mean	&		7.327E+03	&		4.590E+01	&		3.479E+03	&	1.088E+04	&	8.585E+03	&		1.067E+04	&	8.124E+04	&		6.573E+03	&		6.828E+03	&		5.151E+04	&	\cellcolor{gray}	1.461E+01	\\
				CE2	&	Std	&		8.899E+02	&		2.381E+01	&		7.675E+02	&	1.584E+03	&	2.527E+03	&		0.000E+00	&	3.411E+04	&		2.125E+03	&		1.671E+03	&		1.220E+04	&		1.002E+01	\\
				&	FR	&		0.000E+00	&		4.000E+00	&		1.000E+02	&	1.000E+02	&	0.000E+00	&		0.000E+00	&	0.000E+00	&		5.000E+00	&		0.000E+00	&		0.000E+00	&		5.000E+00	\\\hline
				&	Mean	&		7.258E+03	&		4.876E+01	&		4.268E+03	&	8.874E+04	&	3.431E+02	&		3.003E+02	&	4.322E+02	&		3.671E+01	&	\cellcolor{gray}	3.393E+01	&		9.680E+02	&		6.804E+01	\\
				CE3	&	Std	&		6.574E+02	&		3.024E+01	&		1.179E+03	&	2.760E+04	&	3.675E+01	&		0.000E+00	&	3.898E+01	&		1.368E+01	&		8.102E+00	&		1.156E+02	&		2.065E+01	\\
				&	FR	&		0.000E+00	&		0.000E+00	&		0.000E+00	&	0.000E+00	&	1.000E+02	&		1.000E+02	&	1.000E+02	&		1.000E+02	&		1.000E+02	&		5.000E+00	&		1.000E+02	\\\hline
				&	Mean	&		7.455E+01	&	\cellcolor{gray}	2.486E+01	&		2.602E+01	&	4.253E+02	&	3.973E+07	&		3.973E+07	&	3.973E+07	&		3.973E+07	&		3.973E+07	&		3.973E+07	&		3.973E+07	\\
				CE4	&	Std	&		9.654E+00	&		9.120E+00	&		9.086E+00	&	2.803E+01	&	1.529E-08	&		0.000E+00	&	1.529E-08	&		1.529E-08	&		1.529E-08	&		1.529E-08	&		1.529E-08	\\
				&	FR	&		1.000E+02	&		4.000E+00	&		1.000E+02	&	1.000E+02	&	0.000E+00	&		0.000E+00	&	0.000E+00	&		0.000E+00	&		0.000E+00	&		0.000E+00	&		0.000E+00	\\\hline
				&	Mean	&		3.973E+07	&		1.511E+07	&		3.973E+07	&	3.973E+07	&	8.221E+02	&		7.694E+02	&	1.238E+03	&		3.552E+01	&	\cellcolor{gray}	3.035E+01	&		2.643E+03	&		4.809E+01	\\
				CE5	&	Std	&		1.529E-08	&		1.911E-09	&		1.529E-08	&	1.529E-08	&	1.228E+02	&		0.000E+00	&	1.238E+02	&		1.352E+01	&		7.996E+00	&		2.394E+02	&		1.031E+01	\\
				&	FR	&		0.000E+00	&		0.000E+00	&		0.000E+00	&	0.000E+00	&	0.000E+00	&		0.000E+00	&	0.000E+00	&		0.000E+00	&		0.000E+00	&		0.000E+00	&		0.000E+00	\\\hline
				&	Mean	&		1.109E+02	&		3.702E+01	&		2.562E+01	&	1.018E+03	&	-4.698E+02	&		-5.584E+02	&	-4.515E+02	&		-4.490E+02	&		-4.283E+02	&	\cellcolor{gray}	-6.719E+02	&		-6.031E+02	\\
				CE6	&	Std	&		1.317E+01	&		1.019E+01	&		6.107E+00	&	8.230E+01	&	3.823E+01	&		0.000E+00	&	8.725E+01	&		6.251E+01	&		5.266E+01	&		5.365E+01	&		7.118E+01	\\
				&	FR	&		0.000E+00	&		0.000E+00	&		0.000E+00	&	0.000E+00	&	0.000E+00	&		0.000E+00	&	0.000E+00	&		0.000E+00	&		0.000E+00	&		0.000E+00	&		0.000E+00	\\\hline
				&	Mean	&	\cellcolor{gray}	-1.334E+03	&		-2.910E+02	&		-1.354E+02	&	-5.959E+02	&	5.151E+00	&		-9.037E+01	&	7.982E+00	&		-1.953E+01	&		-2.056E+01	&		-3.397E+00	&		-9.037E+01	\\
				CE7	&	Std	&		2.028E+01	&		2.821E+01	&		1.347E+02	&	7.610E+01	&	1.755E+00	&		0.000E+00	&	9.660E-01	&		5.448E+00	&		5.172E+00	&		2.180E+01	&		4.374E-14	\\
				&	FR	&		0.000E+00	&		0.000E+00	&		0.000E+00	&	0.000E+00	&	0.000E+00	&		0.000E+00	&	0.000E+00	&		0.000E+00	&		0.000E+00	&		0.000E+00	&		0.000E+00	\\\hline
				&	Mean	&	\cellcolor{gray}	-9.037E+01	&		-8.304E+01	&		-5.681E+00	&	-9.037E+01	&	3.292E+00	&		-6.092E-01	&	6.889E+00	&		-4.670E-01	&		-2.618E+01	&		3.532E+00	&		-6.092E-01	\\
				CE8	&	Std	&		4.374E-14	&		4.922E+00	&		2.968E+00	&	4.374E-14	&	2.097E+00	&		0.000E+00	&	9.464E-01	&		2.226E-01	&		7.179E+00	&		1.598E+00	&		0.000E+00	\\
				&	FR	&		0.000E+00	&		0.000E+00	&		0.000E+00	&	0.000E+00	&	0.000E+00	&		0.000E+00	&	0.000E+00	&		0.000E+00	&		0.000E+00	&		0.000E+00	&		0.000E+00	\\\hline
				&	Mean	&		-6.092E-01	&		-6.053E-01	&		-5.628E-01	&	-6.092E-01	&	5.039E-01	&		-5.357E+01	&	2.984E+01	&		-2.408E+01	&	\cellcolor{gray}	-2.809E+03	&		-2.390E+01	&		-5.357E+01	\\
				CE9	&	Std	&		0.000E+00	&		4.338E-03	&		7.541E-02	&	0.000E+00	&	2.329E+01	&		0.000E+00	&	4.902E+00	&		5.353E+00	&		1.332E+02	&		2.108E+01	&		2.187E-14	\\
				&	FR	&		0.000E+00	&		0.000E+00	&		0.000E+00	&	0.000E+00	&	0.000E+00	&		0.000E+00	&	0.000E+00	&		0.000E+00	&		0.000E+00	&		0.000E+00	&		0.000E+00	\\\hline
				&	Mean	&		-5.357E+01	&		-5.962E+01	&		6.798E-01	&	-5.357E+01	&	-1.808E+03	&	\cellcolor{gray}	-2.969E+03	&	-1.617E+03	&		-2.810E+03	&		1.101E+02	&		-2.692E+03	&	\cellcolor{gray}	-2.969E+03	\\
				CE10	&	Std	&		2.187E-14	&		1.868E-01	&		4.669E+00	&	2.187E-14	&	1.386E+02	&		0.000E+00	&	2.698E+02	&		1.000E+02	&		2.147E+02	&		7.533E+01	&		9.331E-13	\\
				&	FR	&		0.000E+00	&		0.000E+00	&		0.000E+00	&	0.000E+00	&	0.000E+00	&		0.000E+00	&	0.000E+00	&		0.000E+00	&		0.000E+00	&		0.000E+00	&		0.000E+00	\\\hline
				&	Mean	&	\cellcolor{gray}	-2.969E+03	&		-8.205E+02	&		-1.689E+03	&	-2.969E+03	&	8.247E+03	&		9.600E+03	&	1.509E+04	&		7.004E+01	&		5.985E+01	&		2.167E+04	&		4.669E+01	\\
				CE11	&	Std	&		9.331E-13	&		4.689E+01	&		8.733E+01	&	9.331E-13	&	1.751E+03	&		0.000E+00	&	1.824E+03	&		5.461E+01	&		3.619E+01	&		5.809E+03	&		1.600E+01	\\
				&	FR	&		0.000E+00	&		0.000E+00	&		0.000E+00	&	0.000E+00	&	0.000E+00	&		0.000E+00	&	0.000E+00	&		0.000E+00	&		0.000E+00	&		0.000E+00	&		0.000E+00	\\\hline
				&	Mean	&		2.500E+02	&		6.703E+01	&		1.047E+02	&	1.049E+04	&	5.478E+08	&		5.417E+08	&	1.161E+09	&		1.842E+03	&	\cellcolor{gray}	1.290E+01	&		4.833E+09	&		7.946E+01	\\
				CE12	&	Std	&		1.674E+01	&		1.121E+01	&		6.760E+01	&	1.531E+03	&	2.241E+08	&		0.000E+00	&	3.336E+08	&		4.358E+03	&		7.883E+00	&		1.347E+09	&		8.916E+01	\\
				&	FR	&		0.000E+00	&		0.000E+00	&		0.000E+00	&	0.000E+00	&	0.000E+00	&		0.000E+00	&	0.000E+00	&		0.000E+00	&		0.000E+00	&		0.000E+00	&		0.000E+00	\\\hline
				&	Mean	&		2.279E+05	&		1.165E+04	&		5.288E+02	&	6.822E+08	&	2.104E+01	&		2.100E+01	&	2.100E+01	&		1.090E+01	&		1.209E+01	&		2.115E+01	&	\cellcolor{gray}	1.861E-02	\\
				CE13	&	Std	&		8.349E+04	&		2.205E+04	&		3.459E+02	&	2.169E+08	&	5.496E-02	&		0.000E+00	&	1.284E-01	&		9.307E+00	&		9.049E+00	&		7.024E-02	&		1.058E-02	\\
				&	FR	&		0.000E+00	&		0.000E+00	&		0.000E+00	&	0.000E+00	&	0.000E+00	&		0.000E+00	&	0.000E+00	&		0.000E+00	&		0.000E+00	&		0.000E+00	&		0.000E+00	\\\hline
				&	Mean	&		2.279E+05	&		1.165E+04	&		5.288E+02	&	6.822E+08	&	2.104E+01	&		2.100E+01	&	2.100E+01	&		1.090E+01	&		1.209E+01	&		2.115E+01	&	\cellcolor{gray}	1.861E-02	\\
				CE14	&	Std	&		8.349E+04	&		2.205E+04	&		3.459E+02	&	2.169E+08	&	5.496E-02	&		0.000E+00	&	1.284E-01	&		9.307E+00	&		9.049E+00	&		7.024E-02	&		1.058E-02	\\
				&	FR	&		0.000E+00	&		0.000E+00	&		0.000E+00	&	0.000E+00	&	0.000E+00	&		0.000E+00	&	0.000E+00	&		0.000E+00	&		0.000E+00	&		0.000E+00	&		0.000E+00	\\\hline
				&	Mean	&		1.711E+01	&		1.674E+00	&		1.644E+01	&	3.128E+01	&	2.995E+02	&		3.661E+02	&	5.430E+02	&		3.368E+00	&		1.895E+00	&		6.253E+02	&	\cellcolor{gray}	9.408E-02	\\
				CE15	&	Std	&		2.545E+00	&		5.996E-01	&		3.357E+00	&	1.912E-01	&	5.623E+01	&		0.000E+00	&	3.954E+01	&		7.134E+00	&		4.209E+00	&		6.761E+01	&		2.534E-02	\\
				&	FR	&		0.000E+00	&		0.000E+00	&		0.000E+00	&	0.000E+00	&	0.000E+00	&		0.000E+00	&	0.000E+00	&		6.500E+01	&		6.000E+01	&		0.000E+00	&		1.000E+01	\\\hline
				&	Mean	&		1.250E+01	&		5.954E+00	&		6.087E+00	&	4.514E+02	&	2.441E+00	&		3.149E+00	&	4.397E+00	&		1.078E-02	&		8.011E-03	&		5.809E+00	&	\cellcolor{gray}	1.217E-05	\\
				CE16	&	Std	&		1.149E+00	&		2.355E+00	&		1.412E+01	&	3.509E+01	&	3.814E-01	&		0.000E+00	&	4.822E-01	&		1.967E-02	&		1.822E-02	&		1.100E+00	&		5.458E-06	\\
				&	FR	&		0.000E+00	&		0.000E+00	&		1.000E+02	&	0.000E+00	&	0.000E+00	&		0.000E+00	&	0.000E+00	&		0.000E+00	&		0.000E+00	&		0.000E+00	&		0.000E+00	\\\hline
				&	Mean	&		7.486E-01	&		3.982E-01	&	\cellcolor{gray}	4.740E-02	&	3.515E+00	&	8.193E+03	&		8.420E+03	&	1.388E+04	&		8.426E+01	&		8.375E+01	&		2.205E+04	&		6.395E+01	\\
				CE17	&	Std	&		6.860E-02	&		1.326E-01	&		5.493E-02	&	2.505E-01	&	2.177E+03	&		0.000E+00	&	2.028E+03	&		8.636E+01	&		5.854E+01	&		4.384E+03	&		1.236E+01	\\
				&	FR	&		0.000E+00	&		0.000E+00	&		0.000E+00	&	0.000E+00	&	0.000E+00	&		0.000E+00	&	0.000E+00	&		0.000E+00	&		0.000E+00	&		0.000E+00	&		0.000E+00	\\\hline
		\end{tabular}}
	\end{table}
	
	\begin{table}
		\caption{Experimental results of all comparison algorithms along with feasibility rate, 30 $d$.}\label{tab: finaldata2-30D}
		\resizebox{1\linewidth}{!}{\begin{tabular}{llccccccccccc}\hline
				Funcs	&	Metrics	&		COA	&		CSA	&		GSA	&	RSO	&	TSA	&		ChOA	&	BWOA	&		AEFA	&		AEFA-C	&		CPSOGSA	&		AI-AEFA	\\\hline
				&	Mean	&		2.284E+02	&		5.195E+01	&		1.234E+02	&	9.545E+01	&	9.082E+01	&		7.939E+01	&	9.421E+01	&		7.604E+00	&		1.000E+01	&		1.019E+02	&	\cellcolor{gray}	-8.460E+00	\\
				CE18	&	Std	&		2.554E+01	&		1.368E+01	&		1.026E+02	&	4.254E+00	&	1.020E+01	&		0.000E+00	&	6.898E+00	&		5.483E+00	&		8.328E+00	&		7.524E+00	&		5.711E+00	\\
				&	FR	&		0.000E+00	&		0.000E+00	&		0.000E+00	&	0.000E+00	&	0.000E+00	&		0.000E+00	&	0.000E+00	&		0.000E+00	&		0.000E+00	&		0.000E+00	&		0.000E+00	\\\hline
				&	Mean	&	\cellcolor{gray}	6.362E-01	&		5.743E+00	&		4.603E+01	&	7.520E+00	&	7.726E+00	&		8.165E+00	&	8.453E+00	&		2.078E+00	&		2.228E+00	&		8.460E+00	&		1.734E+00	\\
				CE19	&	Std	&		2.381E+00	&		1.312E+00	&		5.387E+00	&	7.679E-01	&	5.473E-01	&		0.000E+00	&	6.416E-01	&		4.537E-01	&		6.217E-01	&		4.071E-01	&		3.057E-01	\\
				&	FR	&		0.000E+00	&		0.000E+00	&		0.000E+00	&	8.000E+01	&	1.000E+02	&		7.600E+01	&	6.000E+01	&		1.000E+02	&		1.000E+02	&		5.000E+00	&		1.000E+02	\\\hline
				&	Mean	&		4.806E+00	&	\cellcolor{gray}	1.292E+00	&		1.612E+00	&	3.937E+04	&	2.202E+04	&		3.016E+04	&	6.160E+04	&		2.951E+02	&		2.885E+02	&		7.567E+04	&		5.572E+01	\\
				CE20	&	Std	&		4.919E-01	&		1.544E-01	&		4.831E-01	&	8.269E+03	&	7.748E+03	&		0.000E+00	&	1.449E+04	&		1.808E+02	&		1.752E+02	&		1.522E+04	&		1.287E+01	\\
				&	FR	&		1.000E+02	&		4.000E+00	&		1.000E+02	&	0.000E+00	&	0.000E+00	&		0.000E+00	&	0.000E+00	&		0.000E+00	&		0.000E+00	&		0.000E+00	&		0.000E+00	\\\hline
				&	Mean	&		5.202E+02	&	\cellcolor{gray}	1.002E+02	&		3.259E+02	&	1.356E+10	&	4.191E+09	&		4.065E+09	&	3.546E+10	&		4.715E+05	&		1.701E+06	&		5.188E+10	&		1.842E+02	\\
				CE21	&	Std	&		3.918E+01	&		2.810E+01	&		2.360E+02	&	5.945E+09	&	2.855E+09	&		0.000E+00	&	1.580E+10	&		4.132E+05	&		4.270E+06	&		2.079E+10	&		4.936E+02	\\
				&	FR	&		0.000E+00	&		0.000E+00	&		0.000E+00	&	0.000E+00	&	0.000E+00	&		0.000E+00	&	0.000E+00	&		0.000E+00	&		0.000E+00	&		0.000E+00	&		5.000E+00	\\\hline
				&	Mean	&		1.299E+07	&		1.855E+05	&		3.277E+05	&	2.104E+01	&	2.107E+01	&		2.106E+01	&	2.105E+01	&	\cellcolor{gray}	2.000E+01	&		2.000E+01	&		2.105E+01	&		2.001E+01	\\
				CE22	&	Std	&		1.392E+07	&		2.130E+05	&		8.666E+05	&	6.177E-02	&	4.528E-02	&		0.000E+00	&	9.369E-02	&		3.041E-04	&		2.589E-04	&		4.656E-02	&		1.695E-03	\\
				&	FR	&		0.000E+00	&		0.000E+00	&		0.000E+00	&	0.000E+00	&	0.000E+00	&		0.000E+00	&	0.000E+00	&		0.000E+00	&		0.000E+00	&		0.000E+00	&		0.000E+00	\\\hline
				&	Mean	&		2.096E+01	&		2.042E+01	&		2.000E+01	&	7.139E+01	&	5.340E+01	&		6.034E+01	&	9.354E+01	&		1.482E+01	&		1.480E+01	&		1.215E+02	&	\cellcolor{gray}	8.906E-02	\\
				CE23	&	Std	&		5.847E-02	&		7.285E-02	&		2.322E-04	&	6.465E+00	&	7.496E+00	&		0.000E+00	&	1.144E+01	&		5.643E+00	&		5.483E+00	&		1.384E+01	&		2.921E-02	\\
				&	FR	&		0.000E+00	&		0.000E+00	&		0.000E+00	&	0.000E+00	&	0.000E+00	&		0.000E+00	&	0.000E+00	&		0.000E+00	&		0.000E+00	&		0.000E+00	&		5.000E+00	\\\hline
				&	Mean	&		2.898E+01	&		4.346E+00	&		2.444E+01	&	8.480E+02	&	6.098E+02	&		6.395E+02	&	1.074E+03	&		7.832E+01	&		8.711E+01	&		1.264E+03	&	\cellcolor{gray}	4.372E-01	\\
				CE24	&	Std	&		4.287E+00	&		1.598E+00	&		5.484E+00	&	8.497E+01	&	1.034E+02	&		0.000E+00	&	1.232E+02	&		3.790E+01	&		4.407E+01	&		1.384E+02	&		1.879E-01	\\
				&	FR	&		0.000E+00	&		0.000E+00	&		0.000E+00	&	0.000E+00	&	0.000E+00	&		0.000E+00	&	0.000E+00	&		3.000E+01	&		1.000E+01	&		0.000E+00	&		1.000E+01	\\\hline
				&	Mean	&		8.734E+01	&		1.752E+01	&		1.177E+02	&	1.075E+01	&	6.806E+00	&		8.303E+00	&	1.702E+01	&		4.914E-01	&		4.866E-01	&		2.069E+01	&	\cellcolor{gray}	2.032E-03	\\
				CE25	&	Std	&		1.300E+01	&		9.407E+00	&		2.943E+01	&	2.137E+00	&	1.430E+00	&		0.000E+00	&	2.885E+00	&		2.867E-01	&		2.927E-01	&		4.320E+00	&		3.889E-03	\\
				&	FR	&		0.000E+00	&		0.000E+00	&		1.000E+02	&	0.000E+00	&	0.000E+00	&		0.000E+00	&	0.000E+00	&		0.000E+00	&		0.000E+00	&		0.000E+00	&		0.000E+00	\\\hline
				&	Mean	&		1.049E+00	&		5.975E-01	&		3.544E-02	&	3.755E+04	&	2.518E+04	&		2.485E+04	&	6.581E+04	&		2.870E+02	&		3.375E+02	&		7.215E+04	&		8.200E+01	\\
				CE26	&	Std	&		1.186E-02	&		1.092E-01	&		8.936E-02	&	5.643E+03	&	4.200E+03	&		0.000E+00	&	1.064E+04	&		2.416E+02	&		2.465E+02	&		1.362E+04	&		1.554E+01	\\
				&	FR	&		0.000E+00	&		0.000E+00	&		0.000E+00	&	0.000E+00	&	0.000E+00	&		0.000E+00	&	0.000E+00	&		0.000E+00	&		0.000E+00	&		0.000E+00	&		0.000E+00	\\\hline
				&	Mean	&		4.741E+02	&		8.758E+01	&		4.776E+02	&	1.270E+02	&	1.276E+02	&		1.071E+02	&	1.355E+02	&		2.132E+01	&		2.038E+01	&		1.412E+02	&	\cellcolor{gray}	1.360E+01	\\
				CE27	&	Std	&		4.600E+01	&		2.991E+01	&		2.859E+02	&	7.538E+00	&	1.396E+01	&		0.000E+00	&	1.066E+01	&		6.796E+00	&		8.631E+00	&		1.087E+01	&		7.951E+00	\\
				&	FR	&		0.000E+00	&		0.000E+00	&		0.000E+00	&	0.000E+00	&	0.000E+00	&		0.000E+00	&	0.000E+00	&		0.000E+00	&		0.000E+00	&		0.000E+00	&		0.000E+00	\\\hline
				&	Mean	&		7.936E+01	&	\cellcolor{gray}	9.318E+00	&		5.964E+01	&	5.527E+03	&	7.193E+03	&		4.318E+04	&	6.574E+03	&		5.674E+03	&		4.830E+04	&		1.420E+01	&		1.360E+01	\\
				CE28	&	Std	&		5.519E+00	&		3.314E+00	&		8.760E+00	&	1.709E+03	&	0.000E+00	&		1.435E+04	&	1.926E+03	&		2.305E+03	&		1.168E+04	&		1.396E+01	&		7.951E+00	\\
				&	FR	&		0.000E+00	&		0.000E+00	&		0.000E+00	&	9.200E+01	&	8.000E+01	&		6.800E+01	&	1.000E+02	&		1.000E+02	&		5.000E+00	&		1.000E+02	&		0.000E+00	\\\hline
				
		\end{tabular}}
	\end{table}
	
	\subsection{Experimental discussions}
	The experimental results are presented in this section. Each experiment is conducted 20 times for dimensions 10 and 30, with a maximum of $\l_{max}\times N$ function evaluations, recording the mean, standard deviation (Std), and feasibility rate (FR). These results are compared with ten state-of-the-art algorithms, categorized into four groups. The first group includes AEFA and AEFA-C, the second group comprises GSA and its hybrid, the third group involves CSA, COA, and RSO, and the fourth group consists of ChOA, WOA, and TSA. The experimental results are detailed in Tables S2 and S3 for ten dimensions and Tables \ref{tab: finaldata1-30D} and \ref{tab: finaldata2-30D} for 30 dimensions.
	
	Upon careful examination of these tables, it is evident that no algorithm outperforms others in finding better feasible solutions for specific problems, such as CE5$-$CE11, CE14, CE18, CE19, CE23, and CE26$-$CE28. Analyzing the results in Tables S2 and S3, as well as Tables \ref{tab: finaldata1-30D} and \ref{tab: finaldata2-30D}, the comparison of algorithms in the first group (AEFA and AEFA-C) reveals that AI-AEFA achieves superior fitness values on fifteen and thirteen problems for 10 $d$ and twenty problems for 30 $d$, respectively. The proposed algorithm and AEFA/AEFA-C have identical fitness values for four problems at 10 $d$ and two problems at 30 $d$. This underscores the effectiveness of the proposed algorithm over AEFA and AEFA-C. Additionally, the proposed algorithm finds feasible solutions for ten issues, surpassing AEFA and AEFA-C, which only achieve feasibility in six problems. This underscores the potential of the new Coulomb's constant definition to prevent falling into infeasible regions and enhance optimal performance.
	
	In the second group, which includes GSA and its hybrid CPSOGSA, AI-AEFA achieves superior optimal solutions on fifteen and twenty-two problems for 10 $d$ and nineteen and twenty-five problems for 30 $d$, respectively. These results demonstrate the significant superiority of AI-AEFA over GSA and its hybrid. For the third group, encompassing COA, RSO, and CSA, AI-AEFA achieves better optimal values on twenty-two, twenty-two, and twenty-seven optimization problems for 10 $d$ and twenty, twenty-three, and twenty issues for 30 $d$, respectively.
	
	In comparing algorithms in the fourth group (TSA, ChOA, BWOA), AI-AEFA achieves better results on twenty-four, twenty-four, and twenty-two problems for 10 $d$. For 30 $d$, the proposed algorithm has minimum optimal values on twenty-seven, twenty-three, and twenty-seven, respectively. This discussion underscores the effectiveness of AI-AEFA on the third and fourth groups, which include newly designed optimization algorithms.
	
	In conclusion, the performance of AI-AEFA on the selected benchmark problems, compared to various optimization algorithms, including a variant of AEFA, a hybrid of GSA, and newly designed algorithms, is excellent. The proposed algorithm can balance searching capabilities and efficiently find optimal solutions for different constrained benchmark problems.
	
\subsection{Statistical test and time complexity}
To validate that the obtained results differ significantly from other selected algorithms, this article conducted a statistical test known as the T-test~\cite{raya2013comparison}. The T-test results, expressed as p-values, are presented in Tables S4 and S5 to compare AI-AEFA and other algorithms. In these tables, ``NA" indicates that the data is not applicable or the algorithm has the same fitness values as AI-AEFA.

From Table~\ref{tab: statistical results}, it is evident that AI-AEFA outperforms COA on nineteen problems, CSA on twenty-seven problems, GSA on twenty-four problems, RSO on twenty-two problems, TSA on twenty-five problems, ChOA on twenty-five problems, BWOA on nineteen problems, AEFA on eighteen problems, AEFA-C on twenty-two problems, and CPSOGSA on twenty problems.

Moreover, this table reveals that AI-AEFA surpasses COA on twenty-two problems, CSA on twenty-six problems, GSA on twenty-two problems, RSO on twenty-one problems, TSA on twenty-seven problems, ChOA on twenty-seven problems, BWOA on twenty-three problems, AEFA on twenty-two problems, AEFA-C on twenty-seven problems, and CPSOGSA on twenty-four problems. Consequently, the proposed algorithm demonstrates superior performance based on statistical results.
	\begin{table}
		\caption{\it Statistical test results, AI-AEFA $vs.$ }\label{tab: statistical results}
		\centering
		\begin{tabular}{lllllll}\hline
			$d$	&	Algorithms	&	COA					&	CSA					&	GSA					&	RSO					&	TSA					\\\hline
			$10$	&	$+/-/\approx$	&	19	/	7	/	2	&	27	/	1	/	0	&	24	/	4	/	0	&	22	/	4	/	2	&	25	/	3	/	0	\\\hline
			$30$	&	$+/-/\approx$	&	22	/	4	/	2	&	26	/	2	/	0	&	22	/	6	/	0	&	21	/	5	/	2	&	27	/	1	/	0	\\\hline
			$d$	&	Algorithms	&	ChOA					&	BWOA					&	AEFA					&	AEFA-C					&	CPSOGSA					\\\hline
			$10$	&	$+/-/\approx$	&	25	/	3	/	0	&	19	/	9	/	0	&	18	/	10	/	0	&	22	/	4	/	2	&	20	/	5	/	3	\\\hline
			$30$	&	$+/-/\approx$	&	27	/	1	/	0	&	23	/	5	/	0	&	21	/	7	/	0	&	27	/	1	/	0	&	24	/	2	/	2	\\\hline
		\end{tabular}
	\end{table}	
	\begin{table}
		\caption{{Time complexity reported for all algorithms according to definition given in~\cite{wu2017problem}.}}\label{tab: timecomplexity}
		\resizebox{1\linewidth}{!}{	\begin{tabular}{llllllllllll}\hline
				$d$&COA	&	CSA	&	GSA	&ChOA&	RSO	&	TSA	&	BWOA	&	AEFA	&	AEFA-C	&	CPSOGSA&AI-AEFA\\\hline	
				10&1.80E-02	&	8.64E-01	&	2.99E-01	&	8.80E-01	&	4.56E-01	&	3.35E-01	&	8.65E-01	&	4.59E-01	&	4.90E-01	&	6.00E-01	&	4.45E-01	\\\hline
				30&2.85E+00	&	5.50E+00	&	1.75E+00	&	5.71E+00	&	1.13E+01	&	3.70E+00	&	5.48E+00	&	3.43E+00	&	4.00E+00	&	3.42E+00	&	3.02E+00	\\\hline
		\end{tabular}}
	\end{table}
	
To assess the time complexity of the AI-AEFA and other selected algorithms, this paper employs the approach outlined in the literature~\cite{wu2017problem}. In this method, complexities for problems $(T_{1_i})$ and algorithms $(T_{2_i})$ are computed over 10,000 function evaluations for each problem $i$. The average time complexity for a problem is calculated as $T_1=(\sum_{i=1}^{28}T_{1_i})/28$, and the algorithm's complexity for each problem is determined as $T_2=(\sum_{i=1}^{28}T_{2_i})/28$. The final time complexity is expressed as $(T_2-T_1)/T_1$. The time complexity results for each algorithm are presented in Table~\ref{tab: timecomplexity}.

According to the outcomes reported in this table, the average time the proposed algorithm takes to solve a problem is 0.445 seconds for 10 $d$ and 3.02 seconds for 30 $d$. These values are notably lower than those for CSA, ChOA, RSO, TSA, BWOA, AEFA, CPSOGSA, and AEFA-C. Consequently, our algorithm can solve a problem in less time than other algorithms. Thus, the AI-AEFA outperforms other comparative algorithms in terms of computational efficiency.

Considering both statistical tests and computational time, the AI-AEFA exhibits significantly superior results compared to other selected competitors.
\section{Industrial optimization problems (IOPs)}\label{reallifeapplication}
In this section, the practical applicability of the proposed AI-AEFA algorithm is assessed on fifteen optimization problems from various fields, drawn from the CEC 2020 real-world optimization problems~\cite{kumar2020test}. These problems span Power Electronic Problems, Livestock Feed Ration Optimization, Industrial Chemical Processes, and Mechanical Engineering Problems. To ensure fair comparisons, each algorithm is run twenty times with the same initial values, set using the MATLAB seed function. The results are compared with those from seven state-of-the-art algorithms, including AEFA and its constrained variant AEFA-C. In the subsequent discussion, the focus will be placed on algorithms that produce results that closely match the optimal values. In the Appendix, the respective mathematical formulations with the optimal values for the selected real-world optimization problems are given. Furthermore, to assess statistical significance, a non-parametric Wilcoxon signed-rank test~\cite{derrac2011practical} is employed at a 5$\%$ significance level. The performance of the proposed AI-AEFA is evaluated based on the following criteria: if the $p-$value is less than 0.05, the performance is deemed superior $(+)$; if it equals 1, there is no discernible difference in performance $(=)$; otherwise, the performance is considered worse $(-)$.
	\subsection{Livestock feed ration optimization}
	In this section, two types of Livestock optimization problems are solved. The Beef cattle case has four problems, and the dairy cattle case has three problems. The main objective of these problems is to find the allocation of each material about its costs. The mathematical formulation of these optimization problems is explained in the supplementary file. 
	\begin{table}[h]
		\footnotesize
		\caption{{Fitness values for real-world Livestock feed ration problems along with statistical test results.}}\label{LF}
\resizebox{1\linewidth}{!}{\begin{tabular}{llllllllll}\hline
			Funcs	&Metrics	&	CSA		&	RSO	&	TSA	&ChOA&	BWOA	&	AEFA	&	AEFA-C	&	AI-AEFA	\\\hline
			&	Mean	&	6.41E+04 \textbf{+}	&	1.65E+04 \textbf{+}	&	2.33E+03 \textbf{+}	&	1.40E+03 \textbf{+}	&	1.21E+03 \textbf{+}	&	1.19E+03 \textbf{-}	&	3.42E+03 \textbf{=}	&	3.85E+03	\\
			LF1	&	Std	&	2.75E+03	&	1.31E+03	&	1.90E+03	&	1.96E+03	&	2.00E+03	&	7.68E+02	&	7.43E+02	&	1.72E+03	\\
			&	Time	&	3.77E-01	&	2.59E-01	&	3.02E+00	&	9.19E+00	&	2.68E-01	&	5.24E-01	&	5.24E-01	&	5.23E-01	\\\hline
			&	Mean	&	6.58E+04 \textbf{+}	&	3.99E+04 \textbf{+}	&	3.99E+03 \textbf{+}	&	5.00E+03 \textbf{+}	&	5.30E+03 \textbf{+}	&	3.16E+03 \textbf{+}	&	3.90E+03 \textbf{+}	&	3.24E+03	\\
			LF2	&	Std	&	2.18E+03	&	6.25E+03	&	3.10E+03	&	4.87E+03	&	3.68E+03	&	1.11E+03	&	1.51E+03	&	2.28E+03	\\
			&	Time	&	4.21E-01	&	2.93E-01	&	3.06E+00	&	9.02E+00	&	3.13E-01	&	7.41E-01	&	6.23E-01	&	6.16E-01	\\\hline
			&	Mean	&	6.24E+04 \textbf{+}	&	6.84E+03 \textbf{+}	&	3.34E+03 \textbf{+}	&	1.93E+03 \textbf{+}	&	1.57E+03 \textbf{+}	&	2.53E+03 \textbf{+}	&	4.18E+03 \textbf{-}	&	3.81E+03	\\
			LF3	&	Std	&	2.85E+03	&	2.85E+03	&	5.19E+03	&	5.95E+03	&	3.45E+03	&	8.15E+02	&	8.42E+02	&	1.77E+03	\\
			&	Time	&	4.46E-01	&	2.97E-01	&	3.11E+00	&	8.84E+00	&	2.72E-01	&	7.03E-01	&	5.53E-01	&	5.35E-01	\\\hline
			&	Mean	&	6.39E+04 \textbf{+}	&	1.85E+04 \textbf{+}	&	9.87E+05 \textbf{+}	&	1.87E+04 \textbf{+}	&	6.23E+04 \textbf{+}	&	2.51E+03 \textbf{+}	&	4.97E+03 \textbf{-}	&	4.24E+03	\\
			LF4	&	Std	&	3.07E+03	&	5.32E+03	&	3.05E+03	&	5.39E+03	&	4.27E+03	&	1.38E+03	&	2.21E+03	&	2.21E+03	\\
			&	Time	&	6.98E-01	&	6.09E-01	&	5.64E+00	&	1.67E+01	&	4.98E-01	&	6.99E-01	&	5.25E-01	&	4.97E-01	\\\hline
			&	Mean	&	7.57E+04 \textbf{+}	&	9.57E+04 \textbf{+}	&	9.43E+03 \textbf{+}	&	6.52E+04 \textbf{+}	&	5.71E+04 \textbf{+}	&	3.34E+03 \textbf{+}	&	5.31E+03 \textbf{+}	&	6.37E+03	\\
			LF5	&	Std	&	3.24E+03	&	7.49E+03	&	6.54E+03	&	8.23E+04	&	2.82E+03	&	1.40E+03	&	2.12E+03	&	2.17E+03	\\
			&	Time	&	7.95E-01	&	4.82E-01	&	9.01E+00	&	2.26E+01	&	4.22E-01	&	6.63E-01	&	4.96E-01	&	4.86E-01	\\\hline
			&	Mean	&	7.48E+04 \textbf{+}	&	2.83E+03 \textbf{+}	&	7.41E-03 \textbf{+}	&	9.88E+04 \textbf{+}	&	5.69E+03 \textbf{+}	&	2.79E+03 \textbf{+}	&	5.13E+03 \textbf{=}	&	4.80E+03	\\
			LF6	&	Std	&	3.66E+03	&	2.79E+03	&	7.65E+03	&	3.99E+03	&	5.47E+03	&	1.19E+03	&	2.02E+03	&	1.63E+03	\\
			&	Time	&	4.88E-01	&	3.28E-01	&	6.87E+00	&	1.88E+01	&	4.01E-01	&	6.43E-01	&	5.20E-01	&	5.09E-01	\\\hline
			&	Mean	&	7.52E+04 \textbf{+}	&	7.52E+03 \textbf{+}	&	1.12E-02 \textbf{+}	&	3.89E+04 \textbf{+}	&	8.52E+03 \textbf{+}	&	2.00E+03 \textbf{+}	&	5.22E+03 \textbf{+}	&	3.25E+03	\\
			LF7	&	Std	&	3.83E+03	&	7.53E+03	&	1.16E+03	&	5.46E+04	&	7.53E+02	&	8.27E+02	&	2.09E+03	&	2.44E+03	\\
			&	Time	&	5.62E-01	&	3.21E-01	&	5.73E+00	&	1.85E+01	&	3.68E-01	&	7.12E-01	&	5.86E-01	&	5.24E-01	\\\hline
			&+/-/=&7/0/0&7/0/0&7/0/0&7/0/0&7/0/0&6/1/0&3/2/2&$--$\\\hline		
		\end{tabular}}
	\end{table}
	
	The experimental results for all algorithms are presented in Table~\ref{LF}, including Mean and Std values along with computational time and statistical results. In this table, the mean values computed by the AI-AEFA are $3.85E+03$, $3.24E+03$, $3.81E+03$, and $4.24E+03$ for beef cattle case problems LF1, LF2, LF3, and LF4. Notably, the optimal values of LF1, LF2, and LF4 achieved by the AI-AEFA are closer to the known fitness values than those obtained by CSA, RSO, TSA, ChOA, BWOA, AEFA, and AEFA-C. While AEFA-C performs better than the proposed algorithm for LF3, the overall effectiveness of the AI-AEFA surpasses other optimization algorithms, including AEFA. Additionally, the proposed algorithm solves these problems in less computation time than AEFA, AEFA-C, TSA, and ChOA for LF1, LF2, and LF3. For LF4, the computation time taken by the AI-AEFA is better than other selected algorithms, including AEFA and its variants.
	
	For the dairy cattle problems, the fitness values of LF5, LF6, and LF7 evaluated by the AI-AEFA are $6.37E+3$, $4.80E+03$, and $3.25E+03$. The performance of the AI-AEFA is superior to other algorithms on two problems, LF5 and LF7. While CSA achieves a better fitness value for LF6 than the AI-AEFA, the optimal value obtained by the AI-AEFA is better than RSO, TSA, ChOA, BWOA, AEFA, and AEFA-C.
	
	Regarding the Wilcoxon signed-rank test results, the proposed AI-AEFA outperforms CSA, RSO, TSA, ChOA, BWOA, AEFA, and AEFA-C on seven, seven, seven, seven, seven, six, and three problems of the Livestock feed ration optimization set, respectively. This non-parametric test provides statistical justification for the observed results.
	\begin{table}[h]
		\footnotesize
		\caption{{Fitness values for real-world mechanical engineering problem along with statistical test results.}}\label{ME}
	\resizebox{1\linewidth}{!}{	\begin{tabular}{llllllllll}\hline
			Funcs	&Metrics	&	CSA		&	RSO	&	TSA	&ChOA&	BWOA	&	AEFA	&	AEFA-C	&	AI-AEFA	\\\hline
			&	Mean	&	5.05E+00 \textbf{+}	&	9.90E+00 \textbf{+}	&	3.57E+00 \textbf{+}	&	3.79E+00 \textbf{+}	&	2.65E+00 \textbf{+}	&	2.88E+00 \textbf{+}	&	3.09E+00 \textbf{-}	&	2.64E+00	\\
			ME1	&	Std	&	3.34E-01	&	2.29E+00	&	4.19E-01	&	7.15E-01	&	4.68E-16	&	1.27E-01	&	2.07E-01	&	1.19E-02	\\
			&	Time	&	7.36E+00	&	7.32E+00	&	8.42E+00	&	9.71E+00	&	7.34E+00	&	3.12E+00	&	2.87E+00	&	2.65E+00	\\\hline
			&+/-/=&1/0/0&1/0/0&1/0/0&1/0/0&1/0/0&1/0/0&0/1/0&$--$\\\hline
		\end{tabular}}
	\end{table}
	\subsection{Mechanical design problem}
	In this section, we address a mechanical design problem known as Topology optimization. The main objective of this problem~\cite{sigmund200199} is to optimize the material layout within a specified search range. The mathematical formulation of this problem is provided in the supplementary file. The known objective value for this problem is $2.6393464970E+00$.
	
	The results obtained by each algorithm are presented in Table~\ref{ME}, including Mean and standard deviation (Std) values along with computational time. It is evident from this table that the mean value computed by the AI-AEFA for problem ME1 is lower than that of other competitive algorithms, including AEFA and AEFA-C. Additionally, the proposed algorithm solves this problem in less computation time than others. In terms of statistical results, the performance of the proposed AI-AEFA surpasses that of CSA, RSO, TSA, ChOA, BWOA, and AEFA. This non-parametric test validates the observed results.
	\begin{table}[htbp]
		\footnotesize
		\caption{{Fitness values for real-world power electronic problems along with statistical test results.}}\label{PE}
\resizebox{1\linewidth}{!}{		\begin{tabular}{llllllllll}\hline
			Funcs	&Metrics	&	CSA		&	RSO	&	TSA	&ChOA&	BWOA	&	AEFA	&	AEFA-C	&	AI-AEFA	\\\hline
			&	Mean	&	2.93E-01 \textbf{+}	&	3.48E-01 \textbf{+}	&	2.35E-01 \textbf{+}	&	2.73E-01 \textbf{+}	&	2.02E-01 \textbf{+}	&	1.29E-01 \textbf{+}	&	1.21E-01 \textbf{+}	&	6.40E-02	\\
			PE1	&	Std	&	4.55E-02	&	1.41E-01	&	2.12E-02	&	4.59E-02	&	1.16E-01	&	2.45E-02	&	1.59E-02	&	3.01E-02	\\
			&	Time	&	1.01E+00	&	5.09E-01	&	2.25E+00	&	4.31E+00	&	5.97E-01	&	1.15E-01	&	1.40E-01	&	7.86E-02	\\\hline
			&	Mean	&	1.47E-01 \textbf{+}	&	1.73E-01 \textbf{+}	&	1.30E-01 \textbf{+}	&	1.46E-01 \textbf{+}	&	9.17E-02 \textbf{+}	&	8.17E-02 \textbf{+}	&	7.89E-02 \textbf{+}	&	2.92E-02	\\
			PE2	&	Std	&	1.99E-02	&	2.37E-02	&	1.76E-02	&	1.61E-02	&	7.32E-02	&	2.62E-02	&	2.63E-02	&	1.79E-02	\\
			&	Time	&	9.78E-01	&	4.82E-01	&	2.32E+00	&	4.19E+00	&	8.81E-01	&	1.28E-01	&	1.42E-01	&	4.31E-02	\\\hline
			&	Mean	&	9.48E-02 \textbf{+}	&	1.38E-01 \textbf{+}	&	8.37E-02 \textbf{+}	&	6.57E-02 \textbf{+}	&	6.76E-02 \textbf{-}	&	4.58E-02 \textbf{+}	&	5.23E-02 \textbf{+}	&	3.27E-02	\\
			PE3	&	Std	&	1.72E-02	&	5.23E-02	&	7.52E-03	&	1.70E-02	&	5.24E-02	&	8.46E-03	&	1.62E-02	&	5.16E-03	\\
			&	Time	&	9.74E-01	&	5.64E-01	&	2.03E+00	&	4.20E+00	&	8.50E-01	&	5.68E-02	&	7.67E-02	&	3.68E-02	\\\hline
			&	Mean	&	7.34E-02 \textbf{+}	&	1.44E-01 \textbf{+}	&	6.58E-02 \textbf{+}	&	5.04E-02 \textbf{+}	&	3.16E-02 \textbf{-}	&	3.58E-02 \textbf{+}	&	4.18E-02 \textbf{+}	&	2.29E-02\\
			PE4	&	Std	&	9.09E-03	&	5.13E-02	&	1.14E-02	&	3.44E-03	&	4.31E-02	&	9.19E-03	&	1.62E-02	&	2.63E-03	\\
			&	Time	&	1.15E+00	&	5.90E-01	&	2.49E+00	&	4.96E+00	&	9.51E-01	&	2.41E-02	&	2.85E-02	&	2.22E-02	\\\hline
			&	Mean	&	6.41E-02 \textbf{+}	&	7.34E-02 \textbf{+}	&	5.27E-02 \textbf{+}	&	5.40E-02 \textbf{+}	&	3.03E-02 \textbf{-}	&	3.35E-02 \textbf{+}	&	3.43E-02 \textbf{+}	&	1.89E-02	\\
			PE5	&	Std	&	9.21E-03	&	1.26E-02	&	7.88E-03	&	7.41E-04	&	4.32E-02	&	1.05E-02	&	6.23E-03	&	3.85E-03	\\
			&	Time	&	1.14E+00	&	5.50E-01	&	2.43E+00	&	4.95E+00	&	9.78E-01	&	3.28E-02	&	3.25E-02	&	1.76E-02	\\\hline
			&+/-/=&5/0/0&5/0/0&5/0/0&5/0/0&2/3/0&5/0/0&5/0/0&$--$\\\hline
		\end{tabular}}
	\end{table}  
	\subsection{Power electronic problems: synchronous optimal pulse-width modulation (SOPM)}
The SOPM is a widely used technique for controlling medium-voltage (MV) drives. It effectively reduces the switching frequency without increasing distortion, resulting in decreased switching losses and improved inverter performance. Switching angles are determined over a single fundamental period while concurrently reducing current distortion. This scalable, constrained optimization problem can be strategically designed.

The statistical results of each algorithm are summarized in Table~\ref{PE}, providing Mean and Std values along with computational time. It is evident from this table that the mean values calculated by the AI-AEFA for problems PE1, PE2, PE3, PE4, and PE5 (0.0640, 0.0292, 0.0372, 0.0229, and 0.0189, respectively) are lower than those obtained by other competitive algorithms, including AEFA and AEFA-C. Furthermore, the proposed algorithm demonstrates superior efficiency in solving these problems within a shorter computation time than other algorithms. In terms of Wilcoxon Signed-rank test results, the performance of the proposed AI-AEFA outperforms CSA, RSO, TSA, ChOA, BWOA, AEFA, and AEFA-C for all five problems in the SOPM set. This non-parametric test provides additional support for the observed results.

	\begin{table}[h]
		\footnotesize
		\caption{{Fitness values for real-world industrial chemical problems along with statistical test results.}}\label{IC}
	\resizebox{1\linewidth}{!}{\begin{tabular}{llllllllll}\hline
			Funcs	&Metrics	&	CSA		&	RSO	&	TSA	&ChOA&	BWOA	&	AEFA	&	AEFA-C	&	AI-AEFA	\\\hline
			&	Mean	&	2.31E+01 \textbf{+}	&	2.45E+01 \textbf{+}	&	2.85E+01 \textbf{+}	&	3.65E+01 \textbf{+}	&	5.05E+01 \textbf{+}	&	7.00E+02 \textbf{+}	&	5.66E+02 \textbf{+}	&	1.68E+02	\\
			IC1	&	Std	&	9.49E+00	&	1.31E+02	&	6.51E+02	&	1.31E+02	&	2.61E+02	&	2.23E+02	&	1.61E+02	&	6.62E+01	\\
			&	Time	&	1.38E-01	&	1.00E-01	&	5.20E-01	&	1.45E+00	&	1.34E-01	&	7.12E-01	&	5.86E-01	&	5.24E-01	\\\hline
			&	Mean	&	8.57E+02 \textbf{+}	&	1.01E+03 \textbf{+}	&	3.58E+02 \textbf{+}	&	3.58E+02 \textbf{+}	&	8.37E+02 \textbf{+}	&	8.73E+03 \textbf{+}	&	1.18E+04 \textbf{+}	&	6.91E+03	\\
			IC2	&	Std	&	1.53E+02	&	3.03E+02	&	5.99E-14	&	5.99E-14	&	3.34E+02	&	3.45E+03	&	1.61E+04	&	9.16E+02	\\
			&	Time	&	2.81E-01	&	1.84E-01	&	1.44E+00	&	3.18E+00	&	2.46E-01	&	3.59E-01	&	1.76E-01	&	1.75E-01	\\\hline
			&+/-/=&2/0/0&2/0/0&2/0/0&2/0/0&2/0/0&2/0/0&2/0/0&$--$\\\hline		
		\end{tabular}}
	\end{table}
	\subsection{Industrial chemical processes}
	In this section, two optimization problems related to industrial chemical processes are addressed, specifically focusing on heat exchanger network design for Case 1 and Case 2. These optimization problems are inherently nonlinear and involve nonlinear constraints. The mathematical models for these optimization problems are detailed in the supplementary file.
	
	The results for these two optimization problems obtained by each algorithm are presented in Table~\ref{IC}, showcasing mean and standard deviation values alongside computational time. A closer inspection of the table reveals that the mean values achieved by AI-AEFA, namely $1.68E+02$ for IC1 and $6.91E+03$ for IC2, are in proximity to the objective values compared to other competitive algorithms, including AEFA and AEFA-C. Regarding computational time, AI-AEFA demonstrated efficiency by solving IC1 in 0.524 seconds, which is less time-consuming than AEFA and AEFA-C. For IC2, the algorithm completed the optimization in 0.175 seconds, outperforming CSA, RSO, ChOA, BWOA, AEFA, and AEFA-C in terms of time efficiency. The statistical comparison further supports the superior performance of the proposed AI-AEFA over other comparative algorithms.
	\section{Reliability-redundancy allocation (RRA) problems}~\label{reliability}
	 In this section, seven RRA problems (defined in the supplementary file) are examined, and the results obtained are compared with those obtained using several state-of-the-art algorithms. The performance of AI-AEFA is thoroughly explored in the context of RRA problems, and specific adaptations in the AI-AEFA approach to address integer values in RRA problems are outlined in Algorithm~\ref{pseudorrap}.
	 
	 Moreover, the maximum possible improvement index (MPII) is introduced as a metric to assess the relative improvement and gauge the effectiveness of the proposed technique. MPII is defined as:
	\begin{equation}
	MPII=abs\bigg(\frac{Mean(\text{AI-AEFA})-Mean(\text{Other algorithm})}{1-Mean(\text{Other algorithm})}\bigg).
	\end{equation}
	In this equation, $Mean(\text{AI-AEFA})$ and $Mean(\text{Other algorithms})$ represent the mean fitness values of the proposed AI-AEFA and other selected competitive algorithms for the RRA problem, respectively. A larger value for MPII indicates greater improvements. Initially, the experimental results are compared with variants of AEFA, followed by a comparison with results from various optimization algorithms reported in relevant studies, such as SCA~\cite{gen2006soft}, SAA~\cite{kim2006reliability}, GA~\cite{yokota1996genetic}, IA~\cite{hsieh2011effective}, TLNNABC~\cite{kundu2022hybrid}, PSO~\cite{huang2015particle}, INGHS~\cite{ouyang2015improved}, NMDE~\cite{zou2011novel}, EBBO~\cite{garg2015efficient}, IABC, GHS~\cite{omran2008global}, HDE~\cite{liao2010two}, NGHS~\cite{zou2010novel}, IPSO~\cite{wu2011improved}, and CS1~\cite{valian2013cuckoo}. The results obtained by AI-AEFA are averaged over 10 independent runs and presented in Tables~\ref{tab:reli1} and ~\ref{tab:reli2}.
	\begin{algorithm}
		\caption{: {Pseudo code for AI-AEFA to deal with RRA problems}}\label{pseudorrap}
		\small
		\begin{algorithmic}[1]
			\STATE Initialize $K_{0}$, $a$, $b$, upper bound, lower bound, $\beta$, $N$, $D$, $\l_{max}$, $\delta$, set $sign=0$ for mixed integer programming problem and $sign=1$ for integer programming problem, $D_c$ for continuous, and $D_{i}$ for integer, $D_c+D_i=D$,
			\STATE Generate initial position within the search bounds $[lb,ub]$ for each agent,
			\STATE  Set initial velocity to zero,
			\FOR {$\l=1:\l_{max}$}
			\STATE Execute lines 4 to 19 of Algorithm~\ref{pseudoAEFAHK}
			\STATE	Update position as:
			\FOR {$i=1:N$}
			\IF {$sign==0$}
			\FOR {$j=1:D_{c}$}
			\STATE $\mathbf{x}(i,j)=\mathbf{x}(i,j)+\mathbf{v}(i,j),$
			\ENDFOR
			\FOR {$j=D_{i}:D$}
			\STATE $\mathbf{x}(i,j)=round(\mathbf{x}(i,j)+\mathbf{v}(i,j)),$ 
			\ENDFOR
			\ELSE
			\FOR {$j=1:D$}
			\STATE $\mathbf{x}(i,j)=round(\mathbf{x}(i,j)+\mathbf{v}(i,j)),$
			\ENDFOR
			\ENDIF
			\ENDFOR
			\STATE	Execute lines 21 to 27 of Algorithm~\ref{pseudoAEFAHK}
			\ENDFOR
			\RETURN the best search agent and the best solution.
		\end{algorithmic}
	\end{algorithm} 

	\begin{table}[h]
	\caption{{Comparison with AEFA variants on RRA problems.}}
	\label{tab:reli1}
	\centering\small
\resizebox{1\linewidth}{!}{	\begin{tabular}{p{3cm}p{1cm}p{3cm}p{3cm}p{3cm}p{3cm}}\hline
		Algorithms&Func&Mean&Std&MPII&Time\\\hline
		AEFA	&	R1	&	9.73530E-01	&	2.18991E-02	&	1.58092E+00	&	1.00000E+00	\\
		AEFA-C	&	R1	&	9.70255E-01	&	1.71968E-02	&	1.29675E+00	&	1.00000E+00	\\
		AI-AEFA	&	R1	&	9.31682E-01	&	1.28682E-02	&	$--$	&	9.99877E-01	\\\hline
		AEFA	&	R2	&	9.99999E-01	&	2.49519E-06	&	5.86957E-01	&	1.00000E+00	\\
		AEFA-C	&	R2	&	9.99998E-01	&	3.07922E-06	&	6.77966E-01	&	1.00000E+00	\\
		AI-AEFA	&	R2	&	9.99999E-01	&	1.20107E-06	&	$--$	&	1.00000E+00	\\\hline
		AEFA	&	R3	&	1.98798E+00	&	1.08263E-02	&	1.00010E+00	&	2.00000E+00	\\
		AEFA-C	&	R3	&	1.99023E+00	&	8.35820E-03	&	1.00010E+00	&	2.00000E+00	\\
		AI-AEFA	&	R3	&	9.99899E-01	&	1.56949E-02	&	$--$	&	9.80603E-01	\\\hline
		AEFA	&	R4	&	9.81621E-01	&	1.18943E-02	&	9.98332E-01	&	1.00000E+00	\\
		AEFA-C	&	R4	&	9.85445E-01	&	1.45196E-02	&	9.97894E-01	&	1.00000E+00	\\
		AI-AEFA	&	R4	&	9.99969E-01	&	8.32841E-03	&	$--$	&	9.83908E-01	\\\hline
		AEFA	&	R5	&	8.35042E-01	&	6.39555E-02	&	1.82699E-01	&	9.99188E-01	\\
		AEFA-C	&	R5	&	8.34117E-01	&	7.65469E-02	&	1.76103E-01	&	9.99587E-01	\\
		AI-AEFA	&	R5	&	8.04904E-01	&	5.61179E-02	&	$--$	&	9.98772E-01	\\\hline
		AEFA	&	R6	&	5.66853E-01	&	1.19434E-01	&	8.74451E-01	&	9.97141E-01	\\
		AEFA-C	&	R6	&	5.96518E-01	&	7.24238E-02	&	8.65221E-01	&	9.64262E-01	\\
		AI-AEFA	&	R6	&	9.45619E-01	&	1.38081E-01	&	$--$	&	8.34767E-01	\\\hline
		AEFA	&	R7	&	9.82329E-01	&	7.12388E-03	&	2.64737E+01	&	1.00000E+00	\\
		AEFA-C	&	R7	&	9.83598E-01	&	1.44553E-02	&	2.85989E+01	&	1.00000E+00	\\
		AI-AEFA	&	R7	&	5.19975E-01	&	1.21076E-02	&	$--$	&	1.00000E+00	\\\hline
		
	\end{tabular}}
	
\end{table}
\begin{table}[h]
	\caption{{The mean and MPII values of RRA problems are compared with other algorithms.}}
	\label{tab:reli2}
	\centering
	\scriptsize
\resizebox{1\linewidth}{!}{	\begin{tabular}{ccccccccc}\hline
		Func	&	SCA	&	SAA	&	GA	&	IA	&	TLNNABC	&	PSO	&	INGHS	&	AI-AEFA	\\\hline
		R1	&	9.31680E-01	&	9.31363E-01	&	9.31460E-01	&	9.31682E-01	&	9.31682E-01	&	8.88504E-01	&	9.31682E-01	&	9.31682E-01	\\\hline
		MPII	&	3.49458E-05	&	4.65328E-03	&	3.24464E-03	&	5.67200E-06	&	1.28078E-06	&	3.87266E-01	&	1.28078E-06	&	$--$	\\\hline
		Func	&	SCA	&	SAA	&	GA	&	IA	&	TLNNABC	&	PSO	&	INGHS	&	AI-AEFA	\\\hline
		R2	&	9.99974E-01	&	9.99976E-01	&	9.99968E-01	&	9.99976E-01	&	9.99986E-01	&	7.51587E-01	&	9.99977E-01	&	9.99999E-01	\\\hline
		MPII	&	9.78077E-01	&	9.76250E-01	&	9.82188E-01	&	9.76250E-01	&	9.59286E-01	&	9.99998E-01	&	9.75599E-01	&	$--$	\\\hline
		Func	&	SCA~	&	SAA	&	GA	&	IA	&	TLNNABC	&	PSO	&	INGHS	&	AI-AEFA	\\\hline
		R3	&	9.99789E-01	&	9.99888E-01	&	9.99879E-01	&	9.99890E-01	&	9.99890E-01	&	9.99671E-01	&	9.99890E-01	&	9.99899E-01	\\\hline
		MPII	&	5.20893E-01	&	1.02313E-01	&	1.65012E-01	&	8.60507E-02	&	8.58023E-02	&	6.92940E-01	&	8.57501E-02	&$--$		\\\hline
		Func&	NMDE	&	SAA	&	EBBO	&	IA	&	TLNNABC	&	PSO	&	INGHS	&	AI-AEFA	\\\hline
		R4	&	9.99955E-01	&	9.99945E-01	&	9.99955E-01	&	9.99955E-01	&	9.99955E-01	&	8.99872E-01	&	9.99955E-01	&	9.99969E-01	\\\hline
		MPII	&	3.23627E-01	&	4.42545E-01	&	3.23567E-01	&	3.23627E-01	&	3.23558E-01	&	9.99694E-01	&	3.23563E-01	&	$--$	\\\hline
		Func&	IABC	&	GHS	&	GA	&	HDE	&	TLNNABC	&	NGHS	&	INGHS	&	AI-AEFA	\\\hline
		R5	&	8.08844E-01	&	8.08844E-01	&	8.08844E-01	&	8.08844E-01	&	8.08844E-01	&	8.08844E-01	&	8.08844E-01	&	8.04904E-01	\\\hline
		MPII	&	2.06101E-02	&	2.06101E-02	&	2.06101E-02	&	2.06101E-02	&	2.06101E-02	&	2.06101E-02	&	2.06101E-02	&$--$		\\\hline
		Func&	HDE	&	IABC	&	GA	&	GHS	&	TLNNABC	&	NGHS	&	INGHS	&	AI-AEFA	\\\hline
		R6	&	9.45613E-01	&	9.45613E-01	&	9.45613E-01	&	9.45613E-01	&	9.45613E-01	&	9.45613E-01	&	9.45613E-01	&	9.45619E-01	\\\hline
		MPII	&	1.09217E-04	&	1.09217E-04	&	1.09217E-04	&	1.09217E-04	&	1.09217E-04	&	1.09217E-04	&	1.09217E-04	&$--$		\\\hline
		Func&	SCA&	IPSO	&	CS1	&	NGHS	&	TLNNABC	&	IABC	&	INGHS	&	AI-AEFA	\\\hline
		R7	&	5.19976E-01	&	5.19976E-01	&	5.19975E-01	&	5.19976E-01	&	5.19975E-01	&	5.19975E-01	&	5.19975E-01	&	5.19975E-01	\\\hline
		MPII	&	1.13753E-02	&	1.13753E-02	&	1.13732E-02	&	1.13753E-02	&	1.13732E-02	&	1.13732E-02	&	1.13732E-02	&$--$		\\\hline
	\end{tabular}}
\end{table}	

Table~\ref{tab:reli1} shows that AI-AEFA outperforms AEFA and AEFA-C in fitness values for R1. The MPII values with AI-AEFA are approximately 1.58092E+00 and 1.29675E+00 higher than AEFA and AEFA-C, indicating better performance. Comparing AI-AEFA with other algorithms in Table~\ref{tab:reli2}, the mean fitness value is 9.31682E-01. For this problem, AI-AEFA achieves improved MPII values ranging from 3.49458E-05 to 3.87266E-01~\cite{gen2006soft}$-$\cite{ouyang2015improved}, showcasing substantial improvements.

Similarly, in R2, AI-AEFA's fitness values surpass those of AEFA-C and closely match AEFA (Table~\ref{tab:reli1}). The MPII values for R2 with AI-AEFA are around 5.86957E-01 and 6.77966E-01 higher than AEFA and AEFA-C, demonstrating superior performance. The mean fitness value of AI-AEFA is 9.99999E-01. For R2, AI-AEFA yields enhanced MPII values ranging from 9.59286E-01 to 9.99998E-01~\cite{gen2006soft}$-$\cite{ouyang2015improved}, indicating a significant improvement in results. In summary, AI-AEFA consistently outperforms AEFA and AEFA-C across both R1 and R2, showcasing better fitness and MPII values, leading to substantial improvements in reliability-redundancy allocation problem solutions. 

Table~\ref{tab:reli1} reveals that AI-AEFA outperforms AEFA and AEFA-C in fitness values for R3. MPII values with AI-AEFA demonstrate an improvement of approximately 1.00010E+00 over AEFA and AEFA-C, indicating superior performance. Comparative analysis with other algorithms (Table~\ref{tab:reli2}) shows a mean fitness value of 9.99899E-01 for AI-AEFA. For R3, MPII values with AI-AEFA exhibit enhancements ranging from 3.23558E-01 to 9.99694E-01~\cite{gen2006soft}$-$\cite{ouyang2015improved}, reflecting a substantial overall improvement.

Similarly, for R4, Table~\ref{tab:reli1} illustrates that AI-AEFA's fitness values surpass those of AEFA and AEFA-C. MPII values for this problem with AI-AEFA exhibit improvements of approximately 9.98332E-01 and 9.97894E-01 compared to AEFA and AEFA-C. Comparative results (Table~\ref{tab:reli2}) show a mean fitness value of 9.99969E-01 for AI-AEFA. The MPII values with AI-AEFA for R4 demonstrate enhancements ranging from 3.23558E-01 to 9.99694E-01~\cite{kim2006reliability, hsieh2011effective, kundu2022hybrid, huang2015particle, ouyang2015improved,garg2015efficient, zou2011novel}, indicating a substantial overall improvement. In conclusion, AI-AEFA consistently exhibits superior fitness and MPII values for both R3 and R4, showcasing substantial improvements in solving RRA problems.

In Table~\ref{tab:reli1}, AI-AEFA demonstrates superior fitness values compared to AEFA and AEFA-C for R5. MPII values for AI-AEFA exhibit an improvement of approximately 1.82699E-01 and 1.76103E-01 over AEFA and AEFA-C, showcasing its enhanced performance. Comparative results with other algorithms (Table~\ref{tab:reli2}) reveal a mean fitness value of 8.04904E-01 for AI-AEFA. For R5, MPII values with AI-AEFA show improvements ranging from 2.06101E-02 to 2.06101E-02~\cite{ghambari2018improved,omran2008global, yokota1996genetic, liao2010two,kundu2022hybrid, zou2010novel, ouyang2015improved}, indicating a substantial overall improvement.

Similarly, for R6, Table~\ref{tab:reli1} illustrates that the fitness values obtained by AI-AEFA surpass those of AEFA and AEFA-C. MPII values for this problem with AI-AEFA show improvements of approximately 8.74451E-01 and 8.65221E-01 compared to AEFA and AEFA-C, reflecting its superior performance. Comparative results (Table~\ref{tab:reli2}) reveal a mean fitness value of 9.45619E-01 for AI-AEFA. For R6, MPII values with AI-AEFA exhibit improvements ranging from 1.09217E-04 to 1.09217E-04~\cite{liao2010two, ghambari2018improved, yokota1996genetic, omran2008global, kundu2022hybrid, zou2010novel, ouyang2015improved}, indicating a substantial overall improvement.

Furthermore, for R7, Table~\ref{tab:reli1} indicates that the fitness values obtained by AI-AEFA outperform those of AEFA and AEFA-C. MPII values for this problem with AI-AEFA show improvements of approximately 2.64737E+01 and 2.85989E+01 compared to AEFA and AEFA-C, emphasizing its superior performance. Comparative results (Table~\ref{tab:reli2}) reveal a mean fitness value of 5.14516E-01 for AI-AEFA. For R7, MPII values with AI-AEFA exhibit improvements ranging from 1.13732E-02 to 1.13753E-02~\cite{gen2006soft, wu2011improved, valian2013cuckoo, zou2010novel, kundu2022hybrid, ghambari2018improved, ouyang2015improved}, indicating a substantial overall improvement. In summary, AI-AEFA consistently demonstrates superior fitness and MPII values for R5, R6, and R7, indicating significant advancements in solving RRA problems.

\section{Explainability of AI-AEFA}\label{explainable}
In this section we have made a novel use of SHAP (Shapley Additive Explanations)~\cite{scott2017unified} that makes AI-AEFA more transparent and interpretable, helping to understand how its key components: Coulombs $(K)$, charge $(Q)$, acceleration $(A)$, and electrostatic force $(E)$, contribute to its performance. SHAP assigns importance scores to each feature based on cooperative game theory, providing a clear breakdown of how different parameters influence the optimization process. This global and local interpretability ensures that AI-AEFA's decision-making is not a black box but rather a well-understood system where feature contributions are quantified. This not only improves trust in the algorithm but also ensures that AI-AEFA maintains an optimal balance between exploration and exploitation, ultimately leading to more efficient and interpretable optimization solutions. Before the presentation of the experimental results, it is important to record some basics of SHAP. 

In this approach, the inputs act like players, and the predictions are seen as the rewards. SHAP determines how much each player contributes to the outcome of the game. It uses a straightforward explanation along with Shapley values (Eq. \ref{shap1}) to create the first prediction model.
\begin{equation} \label{shap1}
  h(z^{'})   =\phi_{0}  + \sum_{i=1}^{N} \phi_{i} z_{i}^{'}   
\end{equation}

In Eq. \ref{shap1},  $h$  stands for the model that explains things, and $z^{'}$  shows the fundamental features. $N$ is the largest size of the coalition, and $\phi$ represents the attribution of the features. According to Lundberg and Lee \cite{scott2017unified}, you can use Eq. \ref{shap2} and Eq. \ref{shap3} to figure out the attribution for each feature.
\begin{equation} \label{shap2}
 \phi_i = \sum_{k \subseteq M(i)} \frac{|k|!(N-|k|-1)!}{N!}[g_{x}(k \cup\{i\})-g_{x}(k)] 
 \end{equation} 

\begin{equation}\label{shap3}
g_{x}(k) = \mathbf{E'}[g(x)|x_k]
\end{equation} 

The term $k$ refers to a subset of the features (inputs), while $M$ represents the complete set of inputs. $ \mathbf{E'}[g(x)|x_k]$ denotes the expected value of the function for the subset $k$.\\

\begin{figure*}[]
    \centering
    \includegraphics[width=1\textwidth]{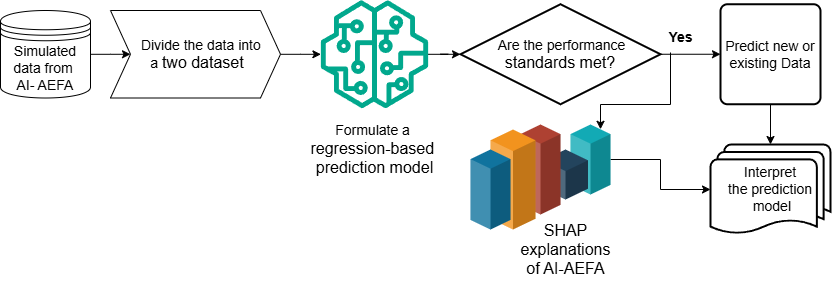} 
     \caption{Workflow using SHAP for an Explainability AI-AEFA Model.}
     \label{fig(12)}
\end{figure*}
\subsection{Experiment and discussion}
\subsubsection{Experimental setup and data}

In this experimental setup, we first stored the values of all parameters of the AI-AEFA model in each iteration process. 
We have considered two datasets. The first data set consists of 3 features, namely, Coulombs $(K)$, charge $(Q)$, and f best for all CE problems in 10 $d$ and 30 $d$. Here, we assume Coulombs $(K)$ and charge $(Q)$ are input features, and $f$ is the best target vector for the predictive model (SHAP value). Similarly, the 2nd dataset also consists of 3 features: acceleration $(A)$ (denoted by $ac$ in the previous sections), electrostatic force $(E)$ (denoted by $F$ in the previous sections), and $X$ vector for all problems in 10 $d$ and 30 $d$. Here, we assume acceleration $(A)$ and electrostatic force $(E)$ are input features, and the X vector is the target vector for the predictive model (SHAP value).

\subsection{Discussion of the results}
 The discussions about the experimental results are explained in Fig. \ref{fig(12)}. In this subsection, we start our analysis with the simulated data of CE1, CE2,..., CE28 for 10 $d$ and 30 $d$. In these figures, we see a cell for each CE problem, within which the contribution to the predictive regression model (SHAP-value) of each characteristic is indicated.

To begin, we plot the correlation heatmap between both data sets for each CE function in 10 $d$ and 30 $d$. Upon analyzing the results (fig \ref{fig(13)}), we observe the following trends: In 10 $d$, for the problems CE4 and CE6, Coulombs $(K)$ shows a stronger correlation with $f$ best compared to charge $(Q)$. However, for the CE9, CE10, and CE23 problems, the charge $(Q)$ is more strongly correlated with f best than Coulombs $(K)$.

In contrast, in 30 $d$, the correlations reverse for certain problems. Specifically, in problem CE4, charge $(Q)$ is more strongly correlated with f best compared to Coulombs $(K)$, a relationship opposite to what was observed in 10 $d$. Similarly, in the problem  CE10,  Coulombs $(K)$ exhibits a stronger correlation with $f$ best than charge $(Q)$, as seen in 10 $d$. On the other hand, in the second dataset, acceleration $(A)$ shows a stronger correlation with the $X$ vector compared to electrostatic force $(E)$.

\begin{figure}[h]
    \centering
    \begin{minipage}{0.49\textwidth}
        \includegraphics[width=\linewidth]{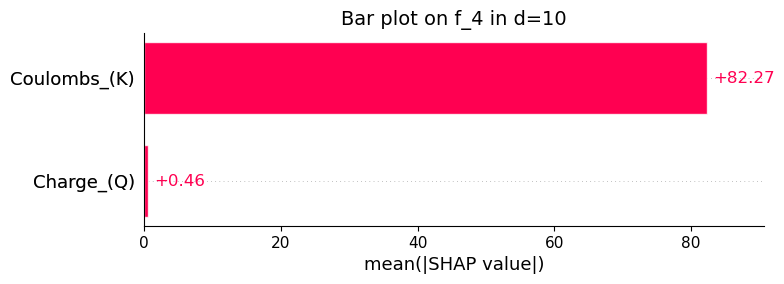}
    \end{minipage}
    \hfill 
    \begin{minipage}{0.49\textwidth}
        \includegraphics[width=\linewidth]{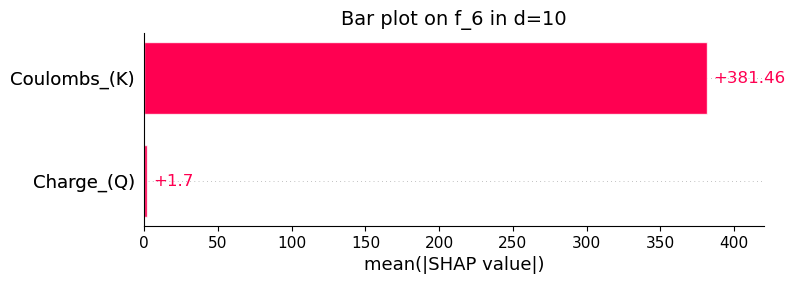}
    \end{minipage}
    \hfill 
     \begin{minipage}{0.49\textwidth}
        \includegraphics[width=\linewidth]{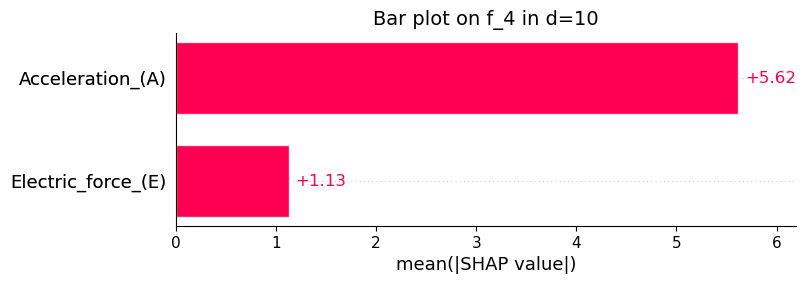}
    \end{minipage}
    \hfill 
    \begin{minipage}{0.49\textwidth}
        \includegraphics[width=\linewidth]{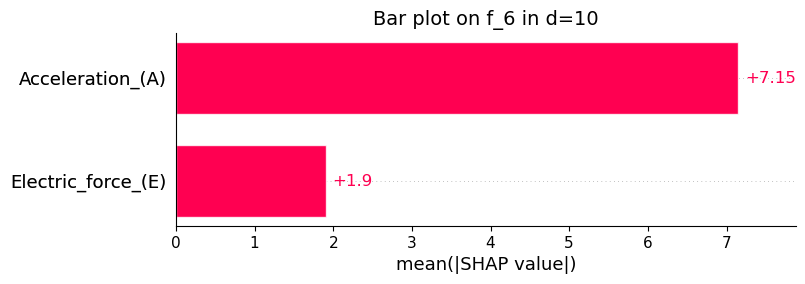}
    \end{minipage}
     \caption{Bar plots on 1st and 2nd datasets for f4 and f6  with d = 10.}  
    
    \label{fig(14)}
\end{figure}

\begin{figure}[]
    \centering
    \begin{minipage}{0.70\textwidth}
        \includegraphics[width=\linewidth]{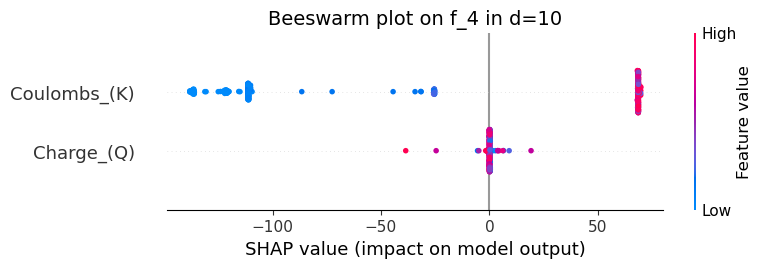}
    \end{minipage}
      \hfill 
    \begin{minipage}{0.70\textwidth}
        \includegraphics[width=\linewidth]{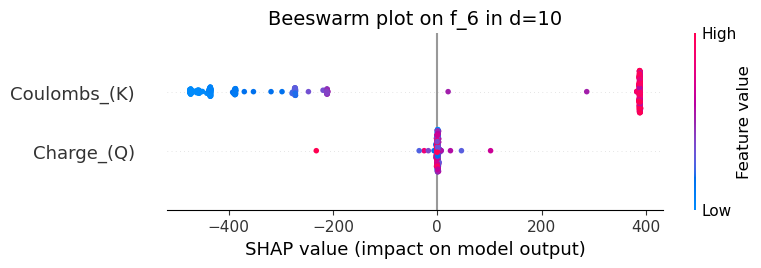 }
       
    \end{minipage}
     \caption{Beeswarm plots on 1st dataset for function f4 and f6 with d=10.}
      \label{fig(15)}
    \end{figure}

Figures \ref{fig(14)}, \ref{fig(15)}, and \ref{fig(16)} show two types of SHAP (Shapley Additive Explanations) value plots: the Bar plot and Beeswarm plot. A SHAP bar plot shows each feature's contribution to the model's prediction, with the bar length indicating the magnitude of the SHAP value. A longer bar means a greater impact on the prediction. For example, in Fig. \ref{fig(14)}, Coulomb's $(K)$ has a longer bar than charge $(Q)$, indicating it contributes more to the prediction. Acceleration $(A)$ generally has larger mean SHAP values than electrostatic force $(E)$, except in CE23 and CE28 for both 10 $d$ and 30 $d$. In 10 $d$ (CE10, CE11) and 30 $d$ (CE14), electrostatic force $(E)$ has higher mean SHAP values.

\begin{figure}[H]
    \centering
    \begin{minipage}{0.3\textwidth}
        \includegraphics[width=\linewidth]{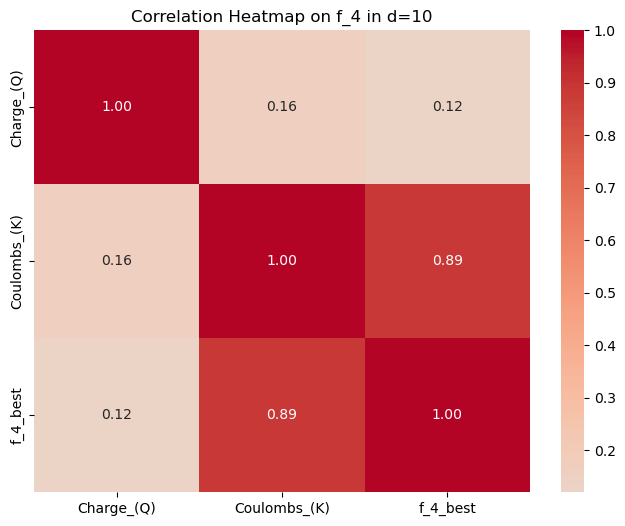 }
    \end{minipage}
    \hfill 
    \begin{minipage}{0.3\textwidth}
        \includegraphics[width=\linewidth]{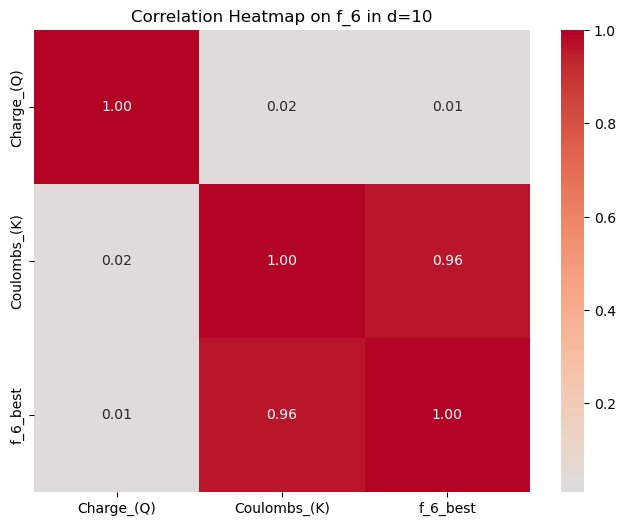}
    \end{minipage}
     \hfill 
    \begin{minipage}{0.3\textwidth}
        \includegraphics[width=\linewidth]{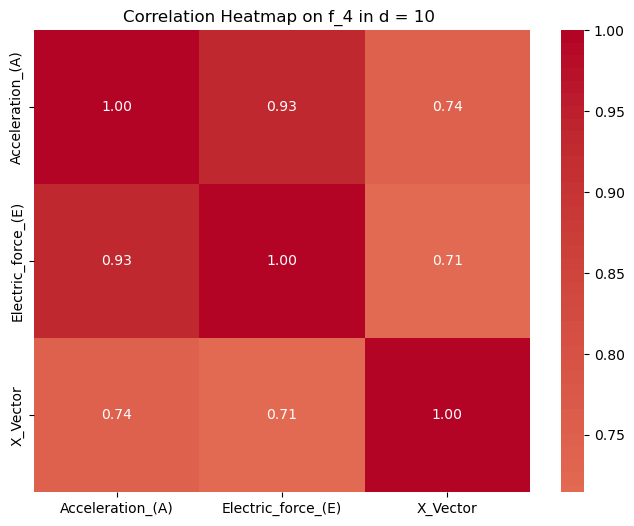}
    \end{minipage}
      \hfill 
    \begin{minipage}{0.3\textwidth}
        \includegraphics[width=\linewidth]{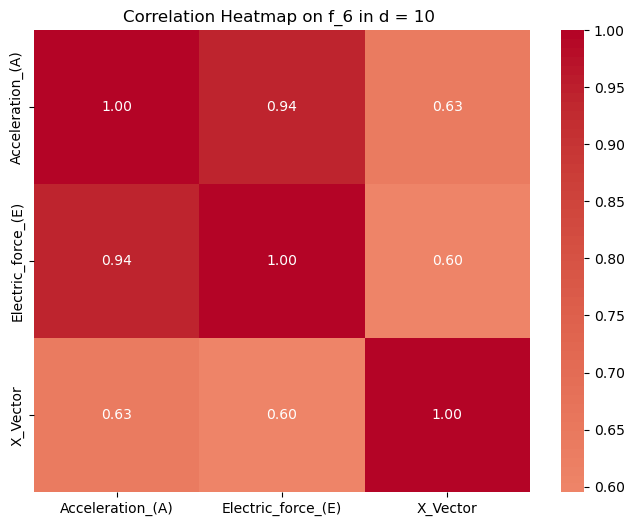 }
    \end{minipage}
        \hfill 
    \begin{minipage}{0.3\textwidth}
        \includegraphics[width=\linewidth]{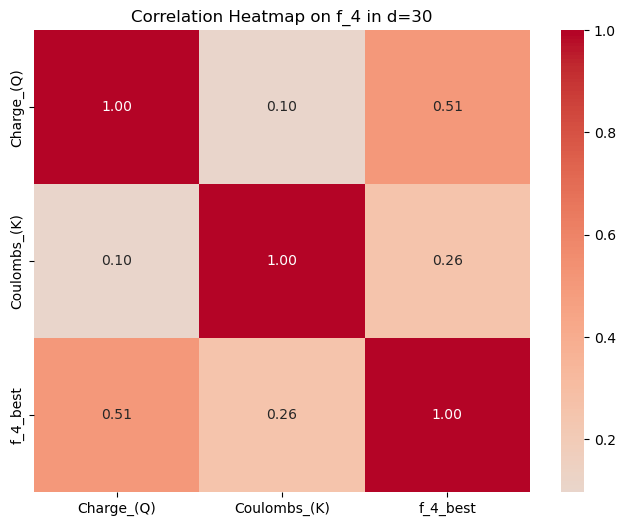}
    \end{minipage}
     \hfill 
    \begin{minipage}{0.3\textwidth}
        \includegraphics[width=\linewidth]{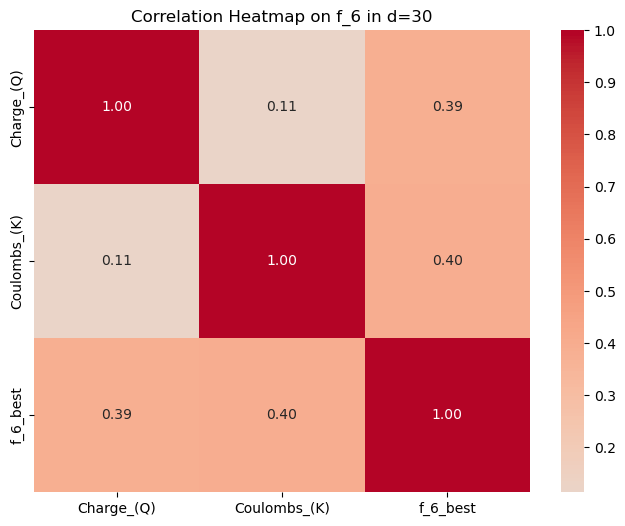}
    \end{minipage}
         \label{fig(13)}
    \begin{minipage}{0.3\textwidth}
        \includegraphics[width=\linewidth]{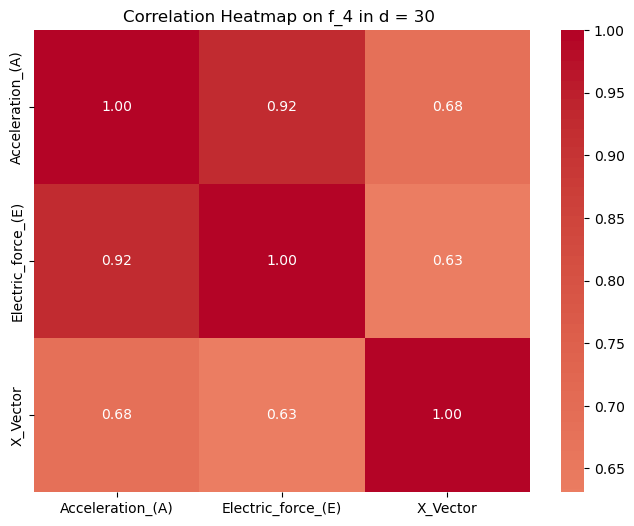 }
    \end{minipage}
    \begin{minipage}{0.3\textwidth}
        \includegraphics[width=\linewidth]{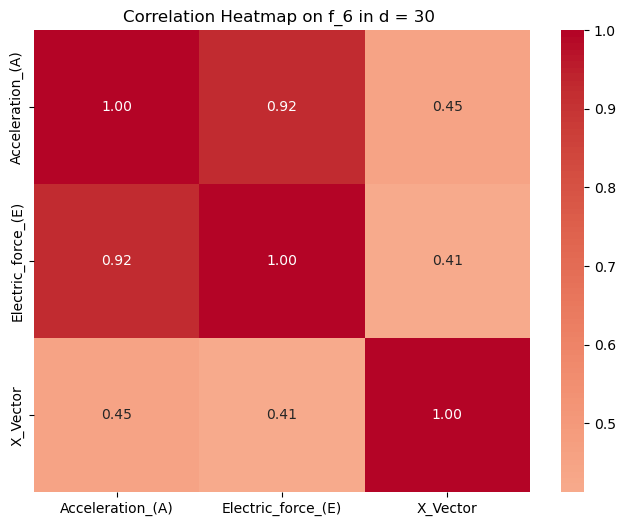 }
    \end{minipage}
    \caption{Correlation heatmap on 1st and 2nd datasets for f4 and f6  with d = 10 and 30.} 
     \label{fig(13)}
\end{figure}

    \begin{figure}[H]
    \centering
    \begin{minipage}{0.7\textwidth}
        \includegraphics[width=\linewidth]{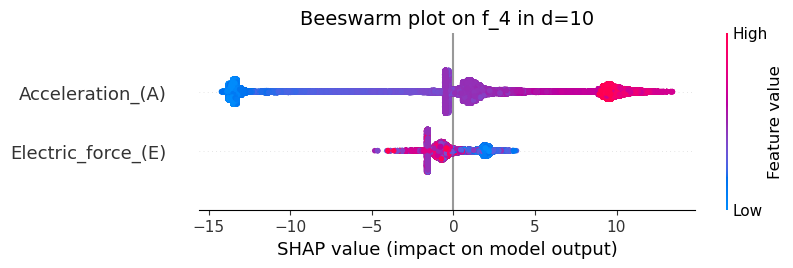}
    \end{minipage}
      \hfill 
    \begin{minipage}{0.7\textwidth}
        \includegraphics[width=\linewidth]{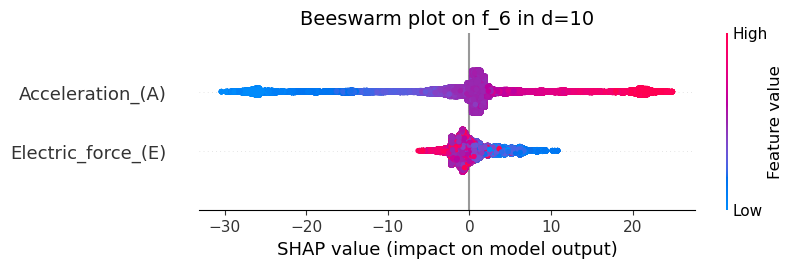 }
    \end{minipage}
   \caption{Beeswarm plots on 2nd dataset for function f4 and f6 with d=10. }
    \label{fig(16)}
\end{figure}

For several features, we see a consistent separation between the two colors, where blue dots consistently have negative SHAP values and red dots are positive. The Beeswarm plots shown in figures \ref{fig(15)} and \ref{fig(16)} sorted by SHAP values, provide a clear visualization of each feature’s impact on the model's predictions. Coulomb's $(K)$ exhibits a strong influence, with both positive and negative SHAP values depending on its range. High values of  Coulomb's $(K)$ lead to positive SHAP values, while low values correspond to negative SHAP values. In contrast, charge $(Q)$ has a minimal impact, as its SHAP values cluster around zero, indicating a neutral effect on the predictions. This sorting in the Beeswarm plot highlights the relative importance of each feature in the model's decision-making process. Similarly, acceleration $(A)$ mostly shows a strong positive influence on the predictions, as indicated by red dots on the positive side. However, for CE23 and CE28, in both 10 $d$ and 30 $d$, this positive influence is weaker, suggesting a lower probability of acceleration $(A)$ driving a positive prediction in these cases. The remaining graphs (Correlation heatmap, Bar, and Beeswarms plots) of all CE functions are placed in the supplementary file.

\subsection{Observations and Recommendations}
The performance of AI-AEFA in solving constrained optimization problems is driven by several key factors, as highlighted by the SHAP-based feature importance analysis, correlation heatmaps, and comparative performance evaluations. The most influential factor is Coulomb's $(K)$, which plays a crucial role in guiding the movement of agents through electrostatic interactions, significantly improving convergence speed and preventing premature stagnation. Acceleration $(A)$ further enhances this process by dynamically adjusting movement intensity, allowing AI-AEFA to balance exploration in early iterations and exploitation in later stages. Electrostatic force $(E)$, influenced by Coulomb's $(K)$ and charge $(Q)$, determines the search direction, with the correlation heatmaps indicating that its role varies across problem dimensions. While charge $(Q)$ contributes minimally, it still interacts with other parameters, suggesting that further refinements could optimize its impact. AI-AEFA also benefits from a robust constraint-handling mechanism, ensuring that solutions remain feasible while optimizing performance, which is evident in its superior feasibility rates compared to AEFA and AEFA-C.
Additionally, its parameter reconfiguration mechanism allows adaptability across different problem scales, maintaining strong performance across varying complexity levels. Maintaining an effective exploration-exploitation balance is another defining strength, where Coulomb's $(K)$ and acceleration $(A)$ drive initial exploration, while a controlled electrostatic force mechanism refines solutions in later stages. Collectively, these factors enable AI-AEFA to achieve faster convergence, higher feasibility rates, and improved optimization efficiency, making it highly effective for industrial and reliability-redundancy allocation problems.

\section{Conclusion}\label{conclusion}
This article proposes a parameter reconfiguration optimization algorithm (AI-AEFA) based on an intelligent parameter and adaptive bounds approach for industrial and reliability optimization problems. The experimental results analysis shows that the effectiveness of AI-AEFA is superior to other comparative state-of-the-art methods, including AEFA and its variant, over twenty-eight CEC 2017 constrained problems, fifteen large-scale industrial optimization problems, and seven reliability-redundancy allocation problems. The statistical test suggests the superiority of the results. From the computation time results, AI-AEFA has solved most optimization problems in less computation time. Furthermore, the use of SHAP has made AI-AEFA more interpretable, allowing for a deeper understanding of the role of key parameters such as Coulombs $(K)$, charge $(Q)$, acceleration $(A)$, and electrostatic force $(E)$. The analysis revealed that Coulomb's $(K)$ is the most influential parameter, playing a crucial role in guiding the search process, while charge $(Q)$ had a minimal impact. The correlation heatmap demonstrated how parameter relationships shift with increasing problem complexity, emphasizing the need for dimension-aware tuning strategies. Additionally, SHAP confirmed that AI-AEFA maintains an optimal balance between exploration and exploitation, contributing to its high feasibility rates and superior optimization performance. The findings of the experimental results suggest that the proposed scheme is highly suitable for solving large-scale, real-world constrained optimization problems with enhanced explainability and efficiency. In the future, this work may be extended to solve multi/many-objectives real-world optimization problems. Additionally, AI-AEFA could be applied to train various neural network models for image segmentation problems, especially those involving large-scale decision variables where interpretability and parameter optimization play a crucial role.

	\section*{Data Availability Statement}
	This work does not include any generated data.
	\section*{Acknowledgment}
	\noindent The authors are thankful to Dr. B. R. Ambedkar National Institute of Technology Jalandhar for the necessary support to this research. The first author is thankful to the Ministry of Education, Government of India, for providing financial support to carry out this work.
	\section*{Compliance with ethical standards}
	\noindent \textbf{Conflict of interest:} All the authors declare that they have no conflict of interest.
	
	\noindent \textbf{Ethical approval:} This article does not contain any studies with human participants or animals performed by any of the authors.
	
	\section*{Supplementary File}
	The supplemental file has been uploaded to GitHub. Anyone can access it using this link \url{https://github.com/ChauhanDikshit/Supplementary-File-AI-AEFA}.
\clearpage
\bibliographystyle{unsrt}
\small
\bibliography{AEFA_CHK}

\begin{thebibliography}{10}

\bibitem{navarro2020survey}
Jorge Navarro-Ortiz, Pablo Romero-Diaz, Sandra Sendra, Pablo Ameigeiras, Juan~J Ramos-Munoz, and Juan~M Lopez-Soler.
\newblock A survey on 5g usage scenarios and traffic models.
\newblock {\em IEEE Communications Surveys \& Tutorials}, 22(2):905--929, 2020.

\bibitem{trivedi2017reliability}
Kishor~S Trivedi and Andrea Bobbio.
\newblock {\em Reliability and availability engineering: modeling, analysis, and applications}.
\newblock Cambridge University Press, 2017.

\bibitem{garg2015efficient}
Harish Garg.
\newblock An efficient biogeography based optimization algorithm for solving reliability optimization problems.
\newblock {\em Swarm and Evolutionary Computation}, 24:1--10, 2015.

\bibitem{ardakan2018multi}
Mostafa~Abouei Ardakan and Mohammad~Taghi Rezvan.
\newblock Multi-objective optimization of reliability--redundancy allocation problem with cold-standby strategy using nsga-ii.
\newblock {\em Reliability Engineering \& System Safety}, 172:225--238, 2018.

\bibitem{yeh2017optimal}
Cheng-Ta Yeh and Lance Fiondella.
\newblock Optimal redundancy allocation to maximize multi-state computer network reliability subject to correlated failures.
\newblock {\em Reliability Engineering \& System Safety}, 166:138--150, 2017.

\bibitem{peiravi2019reliability}
Abdossaber Peiravi, Mahdi Karbasian, Mostafa~Abouei Ardakan, and David~W Coit.
\newblock Reliability optimization of series-parallel systems with k-mixed redundancy strategy.
\newblock {\em Reliability Engineering \& System Safety}, 183:17--28, 2019.

\bibitem{li2022methods}
Yan-Fu Li and Hanxiao Zhang.
\newblock The methods for exactly solving redundancy allocation optimization for multi-state series--parallel systems.
\newblock {\em Reliability Engineering \& System Safety}, 221:108340, 2022.

\bibitem{nath2022evolutionary}
Rahul Nath and Pranab~K Muhuri.
\newblock Evolutionary optimization based solution approaches for many objective reliability-redundancy allocation problem.
\newblock {\em Reliability Engineering \& System Safety}, 220:108190, 2022.

\bibitem{sun2015reliable}
Li~Sun and Chang Liu.
\newblock Reliable and flexible steam and power system design.
\newblock {\em Applied Thermal Engineering}, 79:184--191, 2015.

\bibitem{yeh2019solving}
Wei-Chang Yeh.
\newblock Solving cold-standby reliability redundancy allocation problems using a new swarm intelligence algorithm.
\newblock {\em Applied Soft Computing}, 83:105582, 2019.

\bibitem{hsieh2021component}
Tsung-Jung Hsieh.
\newblock Component mixing with a cold standby strategy for the redundancy allocation problem.
\newblock {\em Reliability Engineering \& System Safety}, 206:107290, 2021.

\bibitem{gholinezhad2017new}
Hadi Gholinezhad and Ali~Zeinal Hamadani.
\newblock A new model for the redundancy allocation problem with component mixing and mixed redundancy strategy.
\newblock {\em Reliability Engineering \& System Safety}, 164:66--73, 2017.

\bibitem{ashraf2015pso}
Zubair Ashraf, Pranab~K. Muhuri, and Q.~M. Danish~Lohani.
\newblock Particle swam optimization based reliability-redundancy allocation in a type-2 fuzzy environment.
\newblock In {\em 2015 IEEE Congress on Evolutionary Computation (CEC)}, pages 1212--1219, 2015.

\bibitem{ouyang2019improved}
Zhiyuan Ouyang, Yu~Liu, Sheng-Jia Ruan, and Tao Jiang.
\newblock An improved particle swarm optimization algorithm for reliability-redundancy allocation problem with mixed redundancy strategy and heterogeneous components.
\newblock {\em Reliability Engineering \& System Safety}, 181:62--74, 2019.

\bibitem{baladeh2022reliability}
Aliakbar~Eslami Baladeh and Sharareh Taghipour.
\newblock Reliability optimization of dynamic k-out-of-n systems with competing failure modes.
\newblock {\em Reliability Engineering \& System Safety}, 227:108734, 2022.

\bibitem{yi2019trade}
Kunxiang Yi, Hui Xiao, Gang Kou, and Rui Peng.
\newblock Trade-off between maintenance and protection for multi-state performance sharing systems with transmission loss.
\newblock {\em Computers \& Industrial Engineering}, 136:305--315, 2019.

\bibitem{meng2021comparative}
Zeng Meng, Gang Li, Xuan Wang, Sadiq~M Sait, and Ali~R{\i}za Y{\i}ld{\i}z.
\newblock A comparative study of metaheuristic algorithms for reliability-based design optimization problems.
\newblock {\em Archives of Computational Methods in Engineering}, 28:1853--1869, 2021.

\bibitem{ha2006reliability}
Chunghun Ha and Way Kuo.
\newblock Reliability redundancy allocation: An improved realization for nonconvex nonlinear programming problems.
\newblock {\em European Journal of Operational Research}, 171(1):24--38, 2006.

\bibitem{monalisa2023multi}
Panda Monalisa, Dehuri Satchidananda, and Jagadev~Alok Kumar.
\newblock Multi-objective artificial bee colony algorithm in redundancy allocation problem.
\newblock {\em International Journal of Advanced Intelligence Paradigms}, 25(1-2):24--50, 2023.

\bibitem{zou2011effective}
Dexuan Zou, Liqun Gao, Steven Li, and Jianhua Wu.
\newblock An effective global harmony search algorithm for reliability problems.
\newblock {\em Expert Systems with Applications}, 38(4):4642--4648, 2011.

\bibitem{afonso2013modified}
Leonardo~Dallegrave Afonso, Viviana~Cocco Mariani, and Leandro dos Santos~Coelho.
\newblock Modified imperialist competitive algorithm based on attraction and repulsion concepts for reliability-redundancy optimization.
\newblock {\em Expert Systems with Applications}, 40(9):3794--3802, 2013.

\bibitem{yeh2021simplified}
Wei-Chang Yeh, Yi-Zhu Su, Xiao-Zhi Gao, Cheng-Feng Hu, Jing Wang, and Chia-Ling Huang.
\newblock Simplified swarm optimization for bi-objection active reliability redundancy allocation problems.
\newblock {\em Applied Soft Computing}, 106:107321, 2021.

\bibitem{he2015novel}
Qiang He, Xiangtao Hu, Hong Ren, and Hongqi Zhang.
\newblock A novel artificial fish swarm algorithm for solving large-scale reliability--redundancy application problem.
\newblock {\em ISA transactions}, 59:105--113, 2015.

\bibitem{yadav2019AEFA}
Anita and Anupam Yadav.
\newblock Aefa: Artificial electric field algorithm for global optimization.
\newblock {\em Swarm and Evolutionary Computation}, 48:93--108, 2019.

\bibitem{chauhan2023archive}
Dikshit Chauhan and Anupam Yadav.
\newblock An archive-based self-adaptive artificial electric field algorithm with orthogonal initialization for real-parameter optimization problems.
\newblock {\em Applied Soft Computing}, page 111109, 2023.

\bibitem{li2011novel}
Yantao Li, Shaojiang Deng, and Di~Xiao.
\newblock A novel hash algorithm construction based on chaotic neural network.
\newblock {\em Neural Computing and Applications}, 20(1):133--141, 2011.

\bibitem{mirjalili2017chaotic}
Seyedali Mirjalili and Amir~H Gandomi.
\newblock Chaotic gravitational constants for the gravitational search algorithm.
\newblock {\em Applied soft computing}, 53:407--419, 2017.

\bibitem{wu2017problem}
Guohua Wu, Rammohan Mallipeddi, and Ponnuthurai~Nagaratnam Suganthan.
\newblock Problem definitions and evaluation criteria for the cec 2017 competition on constrained real-parameter optimization.
\newblock {\em National University of Defense Technology, Changsha, Hunan, PR China and Kyungpook National University, Daegu, South Korea and Nanyang Technological University, Singapore, Technical Report}, 2017.

\bibitem{pierezan2018coyote}
Juliano Pierezan and Leandro Dos~Santos Coelho.
\newblock Coyote optimization algorithm: a new metaheuristic for global optimization problems.
\newblock In {\em 2018 IEEE congress on evolutionary computation (CEC)}, pages 1--8. IEEE, 2018.

\bibitem{askarzadeh2016novel}
Alireza Askarzadeh.
\newblock A novel metaheuristic method for solving constrained engineering optimization problems: crow search algorithm.
\newblock {\em Computers \& Structures}, 169:1--12, 2016.

\bibitem{rashedi2009gsa}
Esmat Rashedi, Hossein Nezamabadi-Pour, and Saeid Saryazdi.
\newblock Gsa: a gravitational search algorithm.
\newblock {\em Information sciences}, 179(13):2232--2248, 2009.

\bibitem{dhiman2021novel}
Gaurav Dhiman, Meenakshi Garg, Atulya Nagar, Vijay Kumar, and Mohammad Dehghani.
\newblock A novel algorithm for global optimization: rat swarm optimizer.
\newblock {\em Journal of Ambient Intelligence and Humanized Computing}, 12(8):8457--8482, 2021.

\bibitem{kaur2020tunicate}
Satnam Kaur, Lalit~K Awasthi, AL~Sangal, and Gaurav Dhiman.
\newblock Tunicate swarm algorithm: A new bio-inspired based metaheuristic paradigm for global optimization.
\newblock {\em Engineering Applications of Artificial Intelligence}, 90:103541, 2020.

\bibitem{khishe2020chimp}
M~Khishe and Mohammad~Reza Mosavi.
\newblock Chimp optimization algorithm.
\newblock {\em Expert systems with applications}, 149:113338, 2020.

\bibitem{hayyolalam2020black}
Vahideh Hayyolalam and Ali Asghar~Pourhaji Kazem.
\newblock Black widow optimization algorithm: a novel meta-heuristic approach for solving engineering optimization problems.
\newblock {\em Engineering Applications of Artificial Intelligence}, 87:103249, 2020.

\bibitem{rather2021constriction}
Sajad~Ahmad Rather and P~Shanthi Bala.
\newblock Constriction coefficient based particle swarm optimization and gravitational search algorithm for multilevel image thresholding.
\newblock {\em Expert Systems}, 38(7):e12717, 2021.

\bibitem{yadav2020artificial}
Anita, Anupam Yadav, and Nitin Kumar.
\newblock Artificial electric field algorithm for engineering optimization problems.
\newblock {\em Expert Systems with Applications}, 149:113308, 2020.

\bibitem{raya2013comparison}
Michele~A Raya, Robert~S Gailey, Ignacio~A Gaunaurd, Daniel~M Jayne, Stuart~M Campbell, Erica Gagne, Patrick~G Manrique, Daniel~G Muller, and Christen Tucker.
\newblock Comparison of three agility tests with male servicemembers: Edgren side step test, t-test, and illinois agility test.
\newblock {\em Journal of Rehabilitation Research \& Development}, 50(7), 2013.

\bibitem{kumar2020test}
Abhishek Kumar, Guohua Wu, Mostafa~Z Ali, Rammohan Mallipeddi, Ponnuthurai~Nagaratnam Suganthan, and Swagatam Das.
\newblock A test-suite of non-convex constrained optimization problems from the real-world and some baseline results.
\newblock {\em Swarm and Evolutionary Computation}, 56:100693, 2020.

\bibitem{derrac2011practical}
Joaqu{\'\i}n Derrac, Salvador Garc{\'\i}a, Daniel Molina, and Francisco Herrera.
\newblock A practical tutorial on the use of nonparametric statistical tests as a methodology for comparing evolutionary and swarm intelligence algorithms.
\newblock {\em Swarm and Evolutionary Computation}, 1(1):3--18, 2011.

\bibitem{sigmund200199}
Ole Sigmund.
\newblock A 99 line topology optimization code written in matlab.
\newblock {\em Structural and multidisciplinary optimization}, 21(2):120--127, 2001.

\bibitem{gen2006soft}
Mitsuo Gen and YoungSu Yun.
\newblock Soft computing approach for reliability optimization: State-of-the-art survey.
\newblock {\em Reliability Engineering \& System Safety}, 91(9):1008--1026, 2006.

\bibitem{kim2006reliability}
Ho-Gyun Kim, Chang-Ok Bae, and Dong-Jun Park.
\newblock Reliability-redundancy optimization using simulated annealing algorithms.
\newblock {\em Journal of Quality in Maintenance Engineering}, 2006.

\bibitem{yokota1996genetic}
Takao Yokota, Mitsuo Gen, and Yin-Xiu Li.
\newblock Genetic algorithm for non-linear mixed integer programming problems and its applications.
\newblock {\em Computers \& industrial engineering}, 30(4):905--917, 1996.

\bibitem{hsieh2011effective}
Y-C Hsieh and P-S You.
\newblock An effective immune based two-phase approach for the optimal reliability--redundancy allocation problem.
\newblock {\em Applied Mathematics and Computation}, 218(4):1297--1307, 2011.

\bibitem{kundu2022hybrid}
Tanmay Kundu and Harish Garg.
\newblock A hybrid tlnnabc algorithm for reliability optimization and engineering design problems.
\newblock {\em Engineering with Computers}, pages 1--45, 2022.

\bibitem{huang2015particle}
Chia-Ling Huang.
\newblock A particle-based simplified swarm optimization algorithm for reliability redundancy allocation problems.
\newblock {\em Reliability Engineering \& System Safety}, 142:221--230, 2015.

\bibitem{ouyang2015improved}
Hai-bin Ouyang, Li-qun Gao, Steven Li, and Xiang-yong Kong.
\newblock Improved novel global harmony search with a new relaxation method for reliability optimization problems.
\newblock {\em Information Sciences}, 305:14--55, 2015.

\bibitem{zou2011novel}
Dexuan Zou, Haikuan Liu, Liqun Gao, and Steven Li.
\newblock A novel modified differential evolution algorithm for constrained optimization problems.
\newblock {\em Computers \& mathematics with applications}, 61(6):1608--1623, 2011.

\bibitem{omran2008global}
Mahamed~GH Omran and Mehrdad Mahdavi.
\newblock Global-best harmony search.
\newblock {\em Applied mathematics and computation}, 198(2):643--656, 2008.

\bibitem{liao2010two}
T~Warren Liao.
\newblock Two hybrid differential evolution algorithms for engineering design optimization.
\newblock {\em Applied Soft Computing}, 10(4):1188--1199, 2010.

\bibitem{zou2010novel}
Dexuan Zou, Liqun Gao, Jianhua Wu, Steven Li, and Yang Li.
\newblock A novel global harmony search algorithm for reliability problems.
\newblock {\em Computers \& Industrial Engineering}, 58(2):307--316, 2010.

\bibitem{wu2011improved}
Peifeng Wu, Liqun Gao, Dexuan Zou, and Steven Li.
\newblock An improved particle swarm optimization algorithm for reliability problems.
\newblock {\em ISA transactions}, 50(1):71--81, 2011.

\bibitem{valian2013cuckoo}
Ehsan Valian and Elham Valian.
\newblock A cuckoo search algorithm by l{\'e}vy flights for solving reliability redundancy allocation problems.
\newblock {\em Engineering Optimization}, 45(11):1273--1286, 2013.

\bibitem{ghambari2018improved}
Soheila Ghambari and Amin Rahati.
\newblock An improved artificial bee colony algorithm and its application to reliability optimization problems.
\newblock {\em Applied Soft Computing}, 62:736--767, 2018.

\bibitem{scott2017unified}
M~Scott, Lee Su-In, et~al.
\newblock A unified approach to interpreting model predictions.
\newblock {\em Advances in neural information processing systems}, 30:4765--4774, 2017.

\end{thebibliography}
\end{document}